\documentclass{article} 
\usepackage{iclr2017_conference,times}
\usepackage{hyperref}
\usepackage{url}
\usepackage{amsfonts}
\usepackage{amssymb}
\usepackage{amsmath}

\usepackage{graphicx}
\usepackage{caption}
\usepackage{subfig}

\usepackage{tablefootnote}
\usepackage{footmisc}
\usepackage{placeins}
\usepackage{multirow}
\usepackage[normalem]{ulem}

\usepackage{booktabs,colortbl}
\usepackage{arydshln}
\title{8-bit Numerical Formats for Deep Neural \\ Networks}

\author{Badreddine Noune, Phil Jones, Daniel Justus, Dominic Masters, and Carlo Luschi \\
Graphcore Research\\
Bristol, UK \\
\texttt{\{badreddine,philj,danielj,dominicm,carlo\}@graphcore.ai} \\
}

\newcommand{\dx}{\, dx}

\iclrfinalcopy 

\begin{document}
\setlength{\abovedisplayskip}{12pt plus 2pt minus 2pt}
\setlength{\belowdisplayskip}{12pt plus 2pt minus 2pt}
\setlength{\abovedisplayshortskip}{8pt plus 2pt minus 2pt}
\setlength{\belowdisplayshortskip}{8pt plus 2pt minus 2pt}
\addtolength{\jot}{6pt}
\maketitle

\begin{abstract}
Given the current trend of increasing size and complexity of machine learning architectures, it has become of critical importance to identify new approaches to improve the computational efficiency of model training.
In this context, we address the advantages of floating-point over fixed-point representation, and present an in-depth study on the use of 8-bit floating-point number formats for activations, weights, and gradients for both training and inference. We explore the effect of different bit-widths for exponents and significands and different exponent biases. The experimental results demonstrate that a suitable choice of these low-precision formats enables faster training and reduced power consumption without any degradation in accuracy for a range of deep learning models for image classification and language processing.
\end{abstract}

\section{Introduction}
The recent advances in AI research and the ground-breaking results achieved in a wide range of practical applications~\citep{Goodfellow16} rely on the possibility of training increasingly large over-parametrized models, based on larger and larger datasets or using unsupervised/self-supervised learning. These breakthroughs have been enabled by the availability of increasing computational resources. However, current technology is no longer able to double the processing speed with every new silicon generation at a constant power, and it has become essential to make more efficient use of the available power.

With the continuing increase of complexity of deep learning applications, the scalability of machine learning systems has also become indispensable. Training of large distributed models creates a number of challenges, relying on the effective use of the available compute, memory, and networking resources shared among the different nodes, limited by the available power budget.
In this context, the use of efficient numerical formats is of critical importance, since it allows increased power efficiency due to both improved computational efficiency and communication efficiency in the exchange of data among processing units.

This paper reviews the recent advances in the use of low-precision numerical formats to enable reliable training of modern deep architectures with power efficient computations, and provides a comprehensive study of the effect of the use of low-precision formats on the training and generalization performance of different model architectures. We also address the trade-off of floating-point formats versus integer formats, and the effect of low-precision arithmetic on non-convex optimization and generalization.

The paper is organized as follows. Section~\ref{sec:Background} reviews recent work on the use of reduced precision numerical formats, and on low-precision training of modern deep neural networks. Section~\ref{sec:8bitNumericalFormats} discusses the performance of model architectures using 8-bit number formats, presenting a range of experimental results of different 8-bit floating-point formats and a systematic investigation of the effect of different exponent biases, providing guidelines for an effective mixed-precision design. 
Section~\ref{sec:AutomaticLossScaling} reports experimental results on the use of adaptive loss scaling algorithms to automate the scaling of gradients and further confirms the performance of the 8-bit formats presented in Section~\ref{sec:8bitNumericalFormats}.
Finally, conclusions are drawn in Section~\ref{sec:Conclusions}.

\section{Background}\label{sec:Background}

\subsection{Low-precision training of deep neural networks}
\citet{Courbariaux14} studied the viability of low precision number formats for training deep neural networks, comparing the accuracy achieved with reduced precision floating-point, fixed-point, and dynamically scaled fixed-point numbers on different image classification benchmarks.
\citet{Gupta15} demonstrated that it is possible to train Fully Connected and Convolutional Neural Networks (CNNs) for image classification using only 16-bit fixed-point arithmetic with {\em stochastic rounding}. \citet{Das18} successfully trained CNNs with 16-bit integer activations, weights and gradients on the ImageNet dataset \citep{Russakovsky15}, using per-tensor scaling factors.

The use of a numeric precision as low as 1 bit for weights and activations of models trained on image classification tasks has been explored by \cite{Courbariaux15}, \cite{Courbariaux16}, and \cite{Hubara16}.
However, these approaches generally induced performance degradation when used on more complex tasks and datasets.  \cite{Zhou16} presented CNNs with numerical precision as low as 1 bit for activations and weights and 2 bits for gradients, but also found a considerable degradation in accuracy on different image classification tasks. \citet{Banner18} demonstrated a successful 8-bit fixed-point quantization scheme for image classification models. This study achieved an accuracy close to baseline on the ImageNet benchmark by adaptively scaling tensors to the dynamic range of the 8-bit fixed-point format, but kept critical values and operations such as gradients with respect to activations and weight updates in higher precision.

\subsection{Scaled integers vs floating-point representation}
While a fixed-point representation covers its dynamic range uniformly, floating-point numbers provide a non-uniform coverage of the quantization range. 
Since activations, weights, and gradients of deep neural networks are typically distributed non-uniformly with mode close to zero~\citep{Miyashita16,Banner18}, floating-point numbers are well-suited for their representation.
Moreover, 8-bit floating-point formats have a much wider dynamic range compared to 8-bit fixed precision representations, and therefore do not necessarily require the implementation of schemes for adaptive scaling per layer or per tensor~\citep{Sun19}.

An 8-bit fixed-point format gives a dynamic range of $42.1 \; \mathrm{dB}$ and a maximum Signal-to-Noise Ratio (SNR) for a standard normal distributed input signal equal to $40.5 \; \mathrm{dB}$, which is obtained for a quantization step $q\approx0.031$ or about $2^{-5}$ (see Appendix~\ref{sec:AppendixSNR_FixedPoint}). However, for quantization steps $q<2^{-5}$ the noise from clipping at overflow rapidly increases, while for $q>2^{-5}$ the noise due to quantization within the dynamic range dominates (Appendix~\ref{sec:AppendixSNR_FixedPoint}, Figure~\ref{fig:FixedPointSNR}(a)). This leaves only a very narrow region where the fixed-point quantization yields a good SNR: for $q=2^{-4}$ the SNR is $34.9 \; \mathrm{dB}$, while for $q=2^{-6}$ we have an SNR of $19.2 \; \mathrm{dB}$ (Appendix~\ref{sec:AppendixSNR_FixedPoint}, Figure~\ref{fig:FixedPointSNR}(b)). Therefore, in the case of sub-optimal scaling or for heavy tailed signal distributions, the SNR of a fixed-point quantization can quickly become much smaller than that of a floating-point quantization \citep{Widrow08}. In contrast, the SNR of a floating-point number is approximately constant within its dynamic range -- see Table~\ref{tab:NumericalFormatsRangeSNR} for a comparison of the dynamic range and SNR of different floating-point formats.

In this paper, different 8-bit floating-point formats are identified with the notation {\em{1.$E$.$p$}}, which specifies the use of 1 sign bit, $E$ exponent bits and $p$ significand bits.
For each floating-point format, we consider a {\em{bias}} value which offsets the range of values covered by the exponent field. This is equivalent to applying a fixed scaling factor.

A particular case of this format is 1.0.7. This format has no exponent bits and a 7 bit significand field, which corresponds to an 8 bit signed integer. Here we still have the freedom to apply an overall scaling factor, which controls the range of numbers represented by the format.
The 1.0.7 format is therefore equivalent to a {\em{scaled integer}}.  

\begin{table}[htb]
\caption{Explicitly stored number of bits for exponent ($E$) and significand ($p$), dynamic range ($D$) and SNR of different floating-point formats.}
\begin{center}
\renewcommand*{\arraystretch}{1.15}
\begin{tabular}{ l l l l l } 
\toprule
\small \sc Format &\small \sc $E$ & \small \sc $p$ & \small \sc $D$ & \small \sc SNR  \\
\midrule
\small IEEE float-32 & \small 8 bits & \small 23 bits & \small $1667.7 \; \mathrm{dB}$ & \small $151.9 \; \mathrm{dB}$ \\
\small IEEE float-16 & \small 5 bits & \small 10  bits & \small $240.8 \; \mathrm{dB}$ & \small $73.7 \; \mathrm{dB}$ \\
\small BFloat-16 & \small 8 bits & \small 7 bits & \small $1571.3 \; \mathrm{dB}$ & \small $55.6 \; \mathrm{dB}$ \\
\small DLFloat\footref{footnote:NoSubNorm}\textsuperscript{, }\footref{footnote:MaxExponent} & \small 6 bits & \small 9 bits & \small $385.3 \; \mathrm{dB}$ & \small $67.6 \; \mathrm{dB}$ \\
\small 1.5.2\footref{footnote:MaxExponent} & \small 5 bits & \small 2 bits & \small $197.5 \; \mathrm{dB}$ & \small $25.5 \; \mathrm{dB}$ \\
\small 1.4.3\footref{footnote:MaxExponent} & \small 4 bits & \small 3 bits & \small $107.8 \; \mathrm{dB}$ & \small $31.5 \; \mathrm{dB}$ \\
\small 1.3.4\footref{footnote:MaxExponent} & \small 3 bits & \small 4 bits & \small $66.0\; \mathrm{dB}$ & \small $37.5 \; \mathrm{dB}$ \\
\bottomrule
\end{tabular}
\end{center}
\label{tab:NumericalFormatsRangeSNR}%
\end{table}

\subsection{Alternative number formats} \label{sec:AlternativeNumberFormats}
 Alternative numerical representations include tapered formats like the {\em posit} format~\citep{Gustafson17}, which aim to improve the trade-off between dynamic range and precision of floating-point formats with the same number of bits. However, this comes at the cost of losing the property of a constant relative quantization error.

Logarithmic floating-point formats such as the {\em deepfloat} format proposed by~\citet{Johnson18} can reduce the hardware cost of multiplications, but increase the implementation complexity of additions. Moreover, deepfloat computation requires new hardware based on Kulisch accumulation~\citep{Kulisch13} and LUT log-to-linear and linear-to-log conversions. On the other hand, the 8-bit floating-point formats considered in this paper have the advantage that they can reuse the same hardware resources as IEEE float-16 and float-32 formats, with the benefit of increased arithmetic efficiency by up to $4\times$ with respect to float-16.

\stepcounter{footnote}\footnotetext{\label{footnote:NoSubNorm}~No sub-normal values.}
\stepcounter{footnote}\footnotetext{\label{footnote:MaxExponent}~Extended exponent range, only one codeword reserved for \texttt{Inf} and \texttt{NaN}.}

\subsection{Mixed-precision floating-point design} \label{sec:MixedPrecisionFloatingPointDesign}
Recently, {\em mixed precision} floating-point approaches have been developed to reduce the memory requirements and speed up computation for training a range of different deep learning models \citep{Micikevicius17}. By using different floating-point representations for activations, weights, gradients and accumulation of partial sums, state-of-the-art results can be obtained while improving memory and computational efficiency. \citet{Micikevicius17} have used IEEE 16-bit floating-point numbers with 5 bits of exponent and 10 bits of significand for representing weights, activations and gradients, while performing accumulation in convolutions and matrix multiplications in IEEE float-32 precision. A float-32 master copy of the weights is kept and used for the weight update to prevent underflow. To enable the use of the IEEE float-16 format, which has a much smaller dynamic range than float-32 (Table~\ref{tab:NumericalFormatsRangeSNR}), \citet{Micikevicius17} have introduced {\em loss scaling}, which is shown to improve the representation of small magnitude gradients for the backward-pass, as will be discussed in detail in Section~\ref{sec:LossScaling}.

To increase the numerical range of the IEEE float-16 format, the {\em DLFloat} format with 6 exponent bits and 9 significand bits \citep{Agrawal19} and the {\em BFloat-16} format with 8 exponent bits and 7 significand bits \citep{Kalamkar19} have been proposed for training deep neural networks. As shown in Table~\ref{tab:NumericalFormatsRangeSNR}, these number formats considerably extend the dynamic range at the cost of a reduced signal-to-noise ratio.

A more aggressive reduction of numeric precision during training promises further improvements in memory and computational efficiency. \citet{Wang18} studied the performance of training with 8-bit floating-point representation of weights, activations and gradients, using 5 exponent bits and 2 significand bits. This work also considered the use of half-precision partial accumulation for matrix multiplications and convolutions and for weight update accumulation, relying on chunk-based computation and floating-point {\em stochastic rounding}. \citet{Mellempudi19} have extended the use of 8-bit floating-point to additional types of models and datasets.

\citet{Sun19} have recently proposed the use of different 8-bit floating-point formats during the forward and the backward pass. This study concluded that models using gradients with floating-point format 1.5.2 and weights and activations with floating-point format 1.4.3 can approach their float-32 counterparts across different applications in image classification and language processing.
However, the study did not consider different values of exponent bias, and did not systematically explore 8-bit quantization of the first layer and of the last fully connected layer.

\subsection{Loss Scaling}\label{sec:LossScaling}
The magnitude of gradients is generally considerably smaller than that of weights and activations~\citep[see also Appendix~\ref{sec:AppendixHistograms}, Figures~\ref{fig:ResNet-32_CIFAR-100_ActivationsHistogram}--\ref{fig:Transformer_HistogramGradWeights}]{Glorot10}. The limited numerical precision of float-16 and float-8 number formats therefore makes low precision models prone to numerical underflow and vanishing of gradients. As a result, training of these models typically relies on one of two alternative methods to improve the representation of small gradients in low precision numerical formats: the use of {\em large exponent biases}, or the use of {\em loss scaling}.

Whilst a large exponent bias moves the range of representable values to smaller magnitudes, {\em loss scaling} scales up the value of the loss by a factor $\alpha$ at the end of the forward pass, before starting the backward pass, and is recovered during weight update~\citep{Micikevicius17,Kuchaiev18}.
For instance, in the case of Stochastic Gradient Descent (SGD) optimization, the loss scaling is simply recovered by scaling down by $\alpha$ the learning rate for the update of the model parameters.

The need for a model-dependent selection of the appropriate value of the loss scaling hyperparameter can be avoided by implementing {\em adaptive loss scaling}~\citep{Kuchaiev18,Zhao19}.
The performance of automatic loss scaling for 8-bit floating-point formats will be considered in Section~\ref{sec:AutomaticLossScaling}.

\section{8-bit numerical formats for mixed precision training}\label{sec:8bitNumericalFormats}
This section discusses the design and performance of 8-bit floating-point formats for quantization of the activations, weights and gradients for training and inference of different deep learning models, for different applications. Experimental results are presented for image classification on CIFAR-100~\citep{Krizhevsky09} and ImageNet~\citep{Russakovsky15} using ResNet models~\citep{He15} and EfficientNet models~\citep{Tan19}, and for language processing, including WMT14 English-German translation~\citep{Bojar16} using the Transformer model~\citep{Vaswani17}, and language understanding using BERT~\citep{BERT18}.

\subsection{Floating-point design}
The IEEE-754 standard, the most commonly used numerical representation, defines a formulaic representation of real numbers as an approximation in order to support a trade-off between range and precision. 

To support the detection of disallowed numerical operations that produce undefined or unrepresentable values, along with the representation of non-finite quantities such as infinities, often encountered in floating-point arithmetic, the IEEE-754 standard reserves the use of the largest unbiased exponent field to signal special values. More specifically, if the biased-exponent field of a floating-point number is filled with all 1-bits, this indicates either an infinity (\texttt{Inf}) or an invalid result of a computation (\texttt{NaN}). However, despite the convenience in signalling \texttt{Inf}/\texttt{NaN} values, this implies that  the number system reduces the range of representable values by a single exponent. Without doubt, as the floating-point format size reduces, the designer’s foremost objective is to maximise the offered range and precision. To this end, the number of codes reserved for non-numerical quantities must be kept at a minimum.

The IEEE-754 floating-point arithmetic also considers the existence of two zeros, namely a negative and a positive zero, despite their equal mathematical significance. In the context of machine intelligence workloads, which favour low-precision floating point arithmetic, it is unnecessary for the number system to comprise of two separate codewords that redundantly represent the same value.

As a nonconformity with the IEEE-754 standard, and in order to efficiently use the (256) available codewords, instead of reserving the all-one biased exponent code to represent special non-numerical values, this biased exponent can extend the range of normalised values by an extra exponent.

\begin{figure}[htb]
\centering
\subfloat[]{\includegraphics[width=0.425\textwidth]{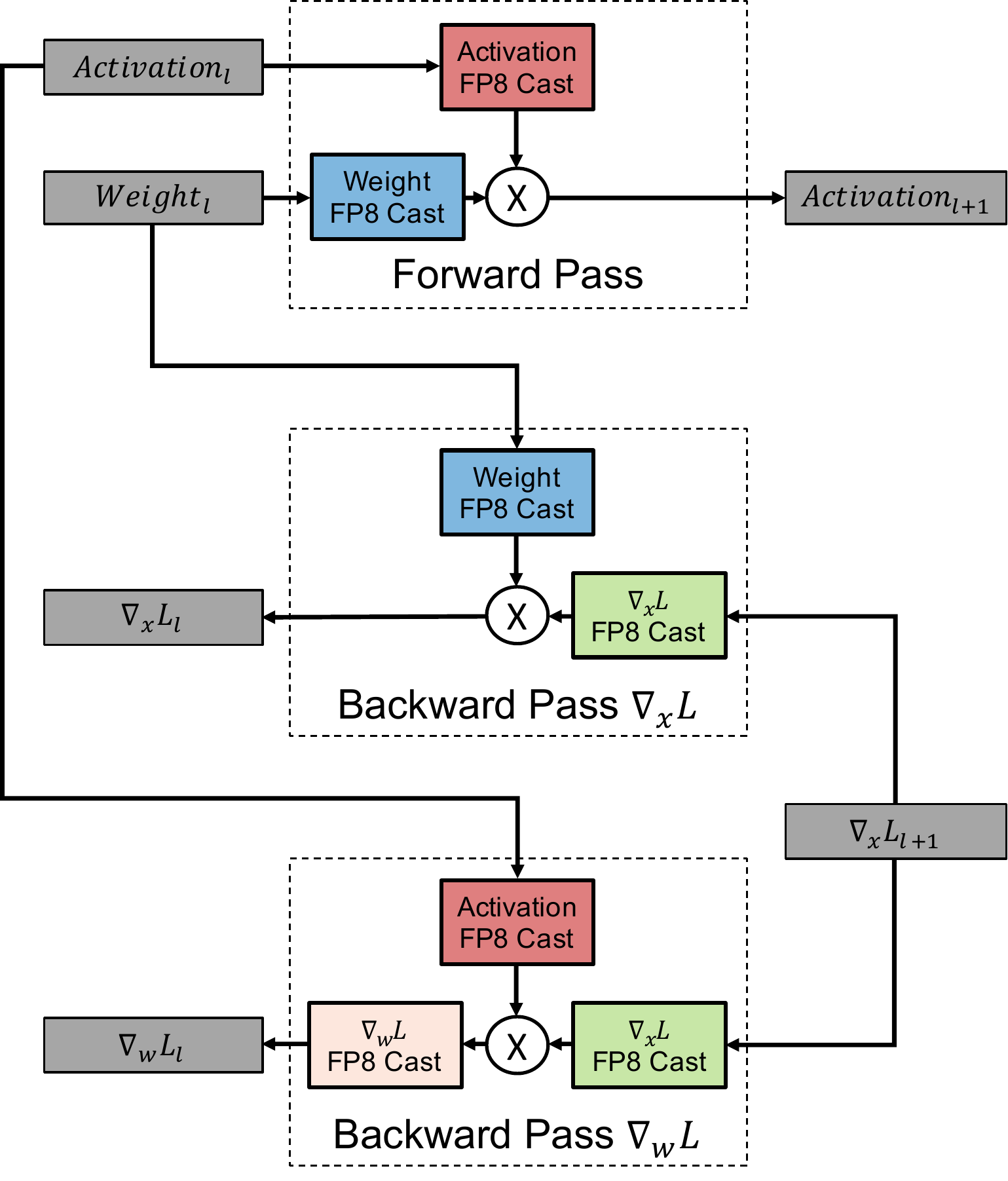}}%
\quad
\subfloat[]{\includegraphics[width=0.425\textwidth]{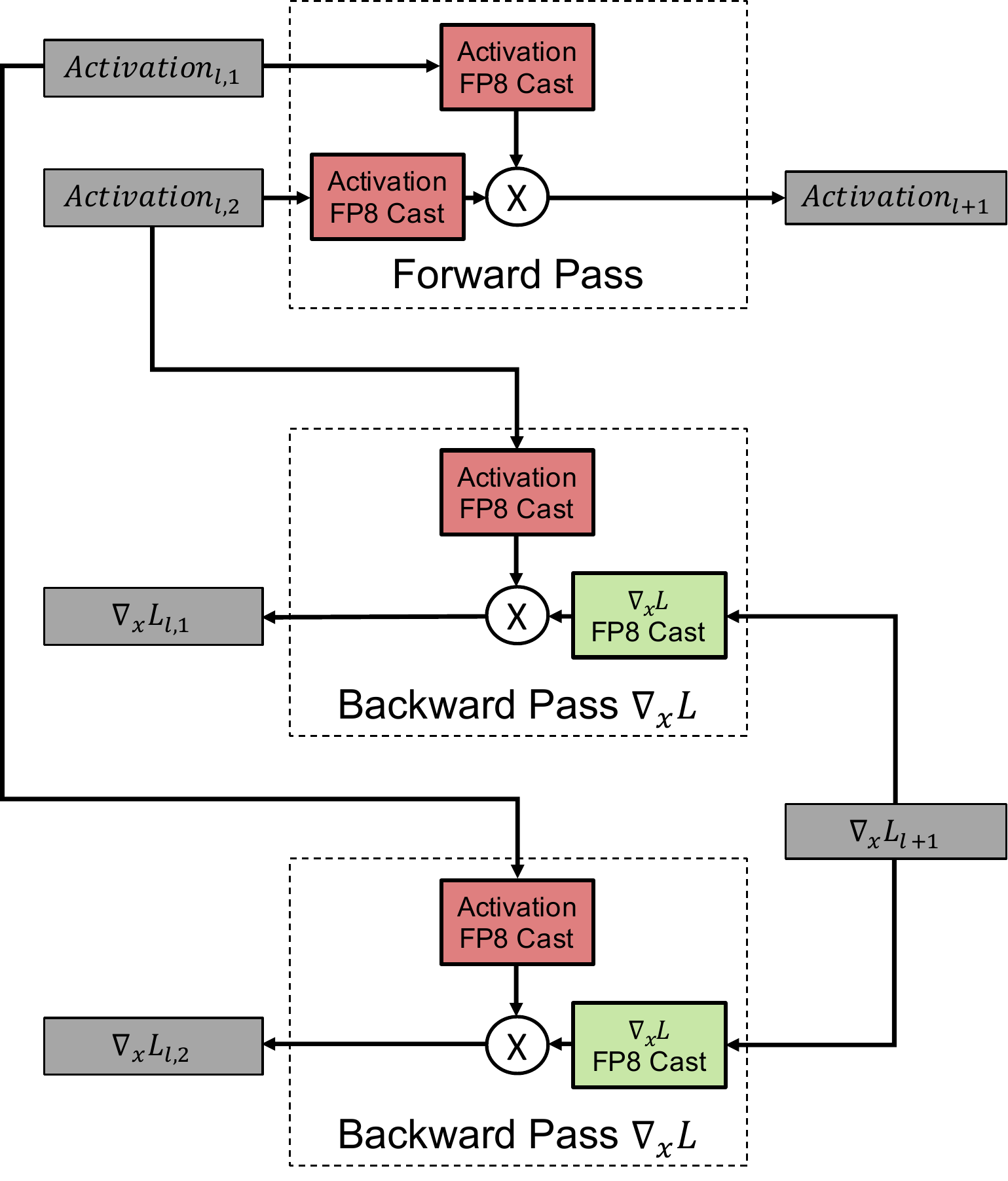}}%
\caption{Illustration of the quantization procedure (a) for fully connected and convolutional layers and (b) for matrix multiplications in attention layers.}%
\label{fig:QuantizationDiagram}%
\end{figure}

\subsection{8-bit floating-point formats for activations, weights and gradients}
Our investigation has been carried out on the range of floating-point representation parameters (exponent field size, mantissa field size and exponent bias range) capable of maintaining the reference baseline performance for different 8-bit floating-point formats.

The best results have been obtained with the 1.5.2 and 1.4.3 formats, which combine a large dynamic range with a sufficiently high SNR. We also report the results obtained with the scaled integer format. Additional results are presented in Appendix~\ref{sec:AppendixOtherNumberFormats} for the float-8 formats 1.6.1, 1.3.4, and 1.2.5.

In all cases, instead of reserving one exponent field to represent \texttt{Inf} and \texttt{NaN}, we reserve only a single codeword (corresponding to negative zero), which allows extension of the range by one extra exponent. If not mentioned otherwise, values in all the experiments are clipped to the maximal representable number instead of returning \texttt{NaN} if an overflow occurs during quantization.

Matrix multiplications and convolutions are by far the most computationally expensive operations for commonly employed deep neural networks. Therefore, the highest gains in terms of performance and energy consumption can be obtained by quantizing the inputs to these operations, in both the forward and backward pass of the training process. Additionally, quantization of the gradients with respect to weights $\nabla_{\bf w}L$ is investigated, as summarized in Figure~\ref{fig:QuantizationDiagram}.

\subsection{Results for Image Processing}\label{sec:Results_ImageClassification}

\begin{figure}[!ht]
\centering
\subfloat[]{\includegraphics[width=0.32\columnwidth]{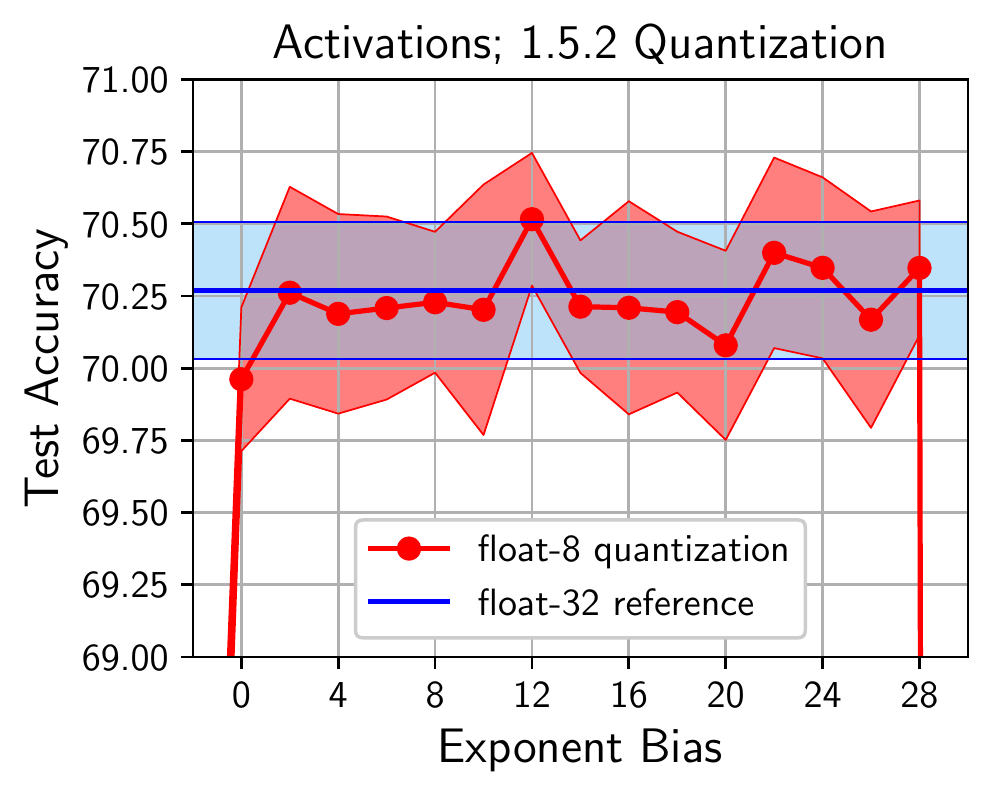}}
\subfloat[]{\includegraphics[width=0.32\columnwidth]{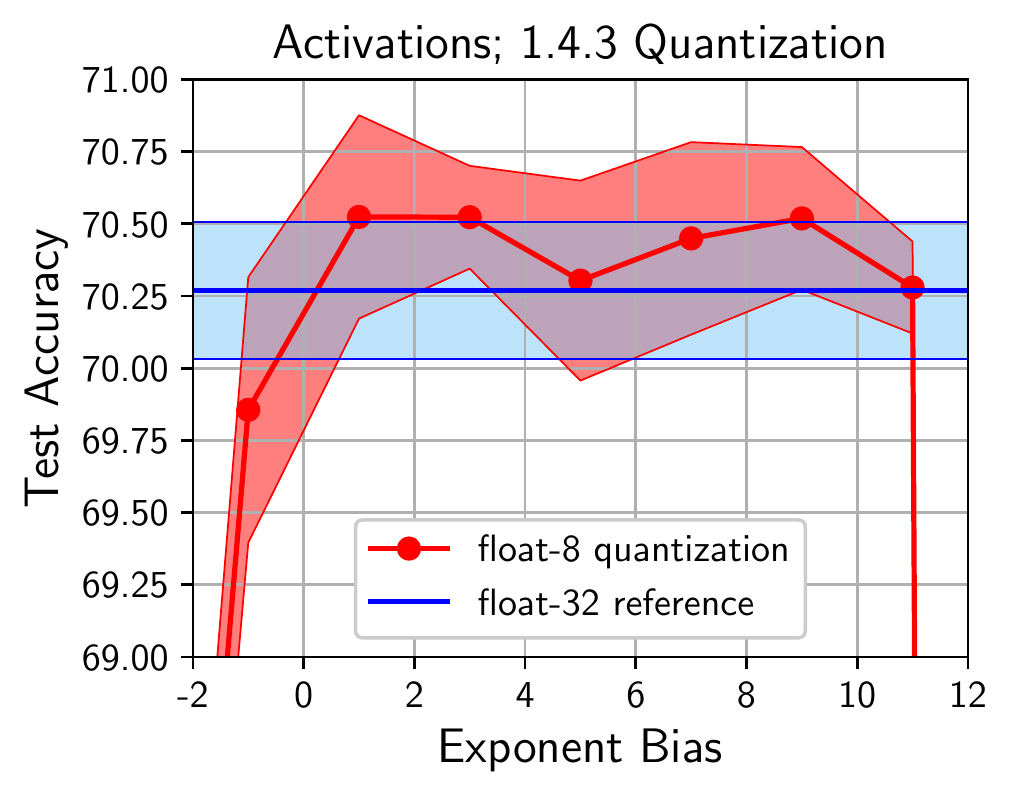}}
\subfloat[]{\includegraphics[width=0.32\columnwidth]{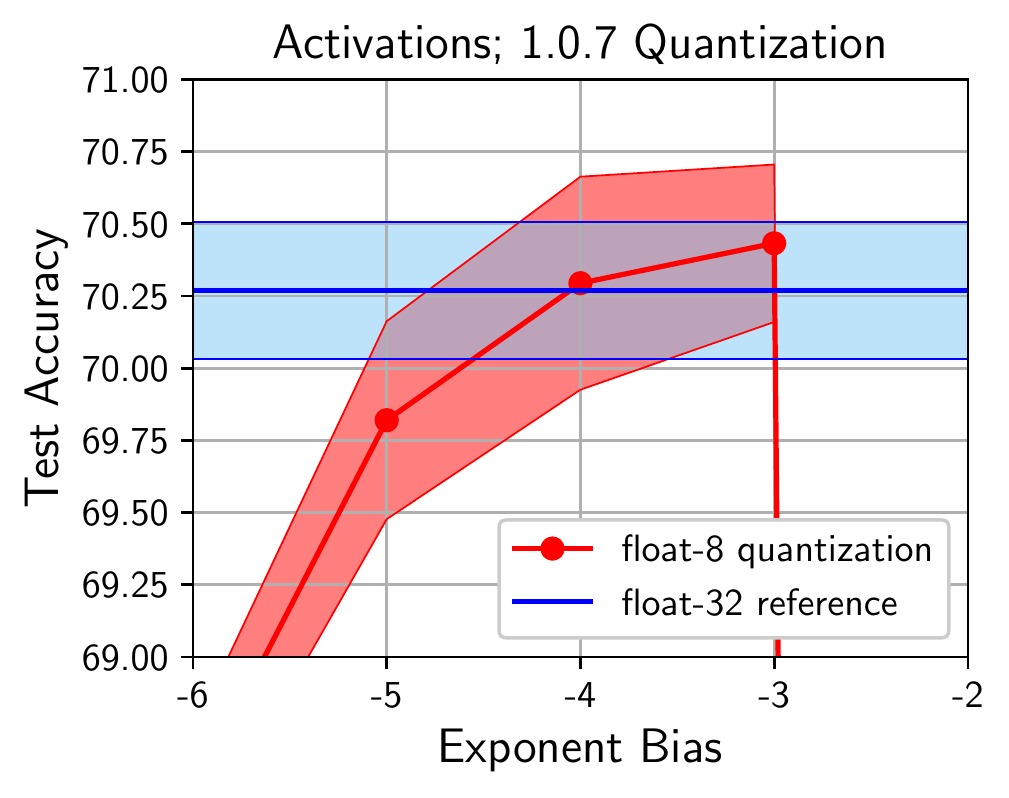}}\\
\subfloat[]{\includegraphics[width=0.32\columnwidth]{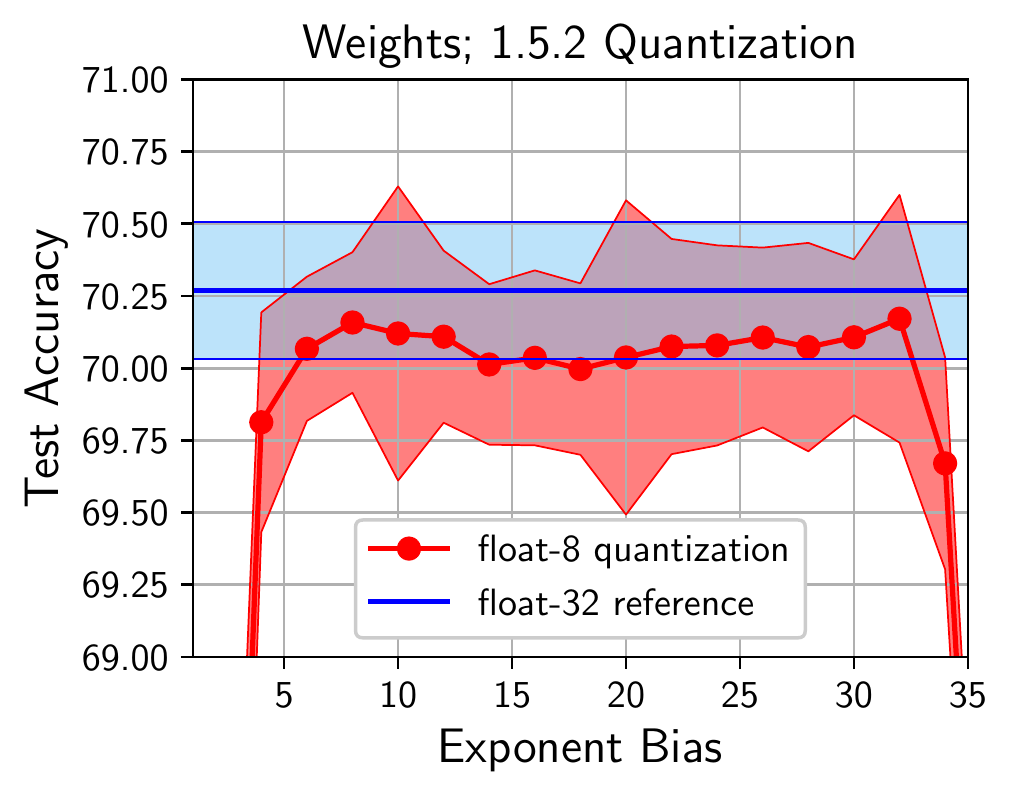}}
\subfloat[]{\includegraphics[width=0.32\columnwidth]{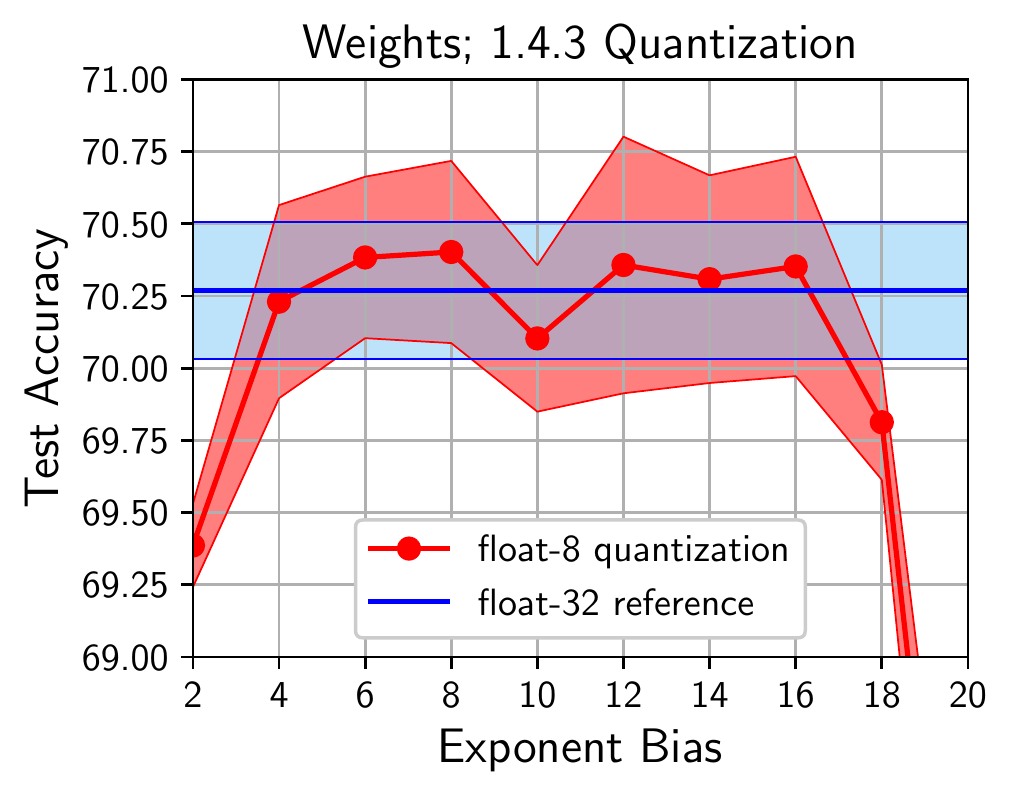}}%
\subfloat[]{\includegraphics[width=0.32\columnwidth]{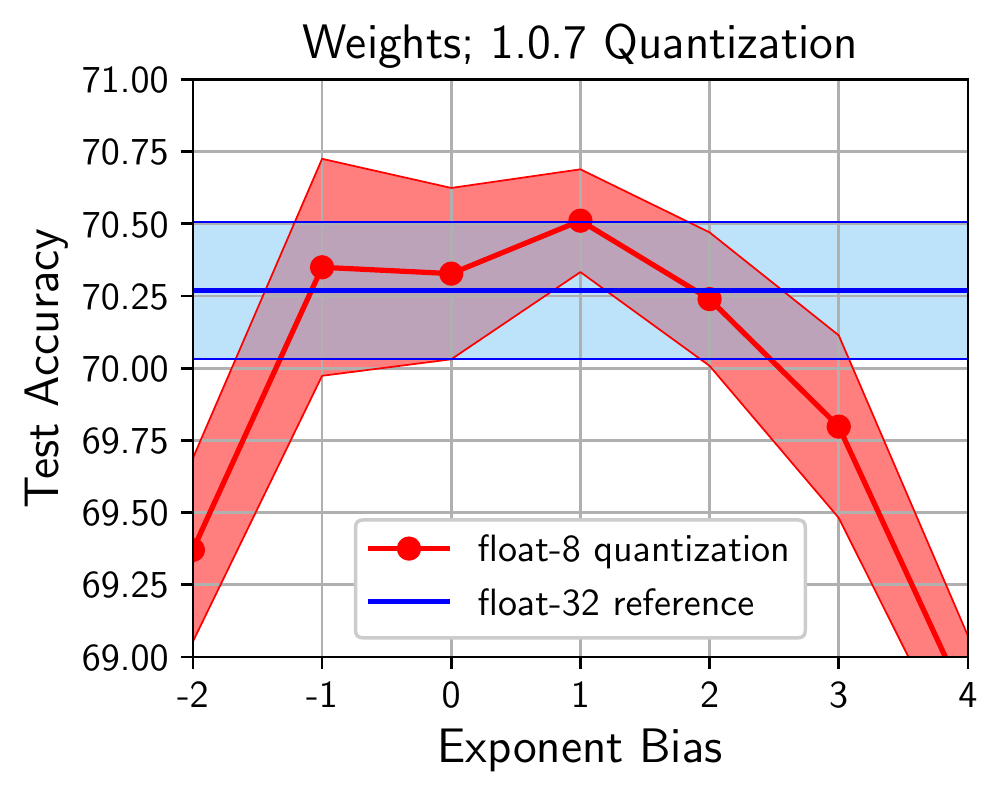}}
\caption{ResNet-32 CIFAR-100 test performance of two different 8-bit floating-point formats compared to the scaled integer format (1.0.7) for representation of the activations without quantization of inputs to the first layer (a, b, c), and of the weights (d, e, f). Test accuracy mean $\pm$ standard deviation over ten independent runs.}%
\label{fig:ResNet-32_CIFAR-100_ActivationsWeightsAccuracy}%
\end{figure}

\begin{figure}[!ht]
\centering
\subfloat[]{\includegraphics[width=0.32\columnwidth]{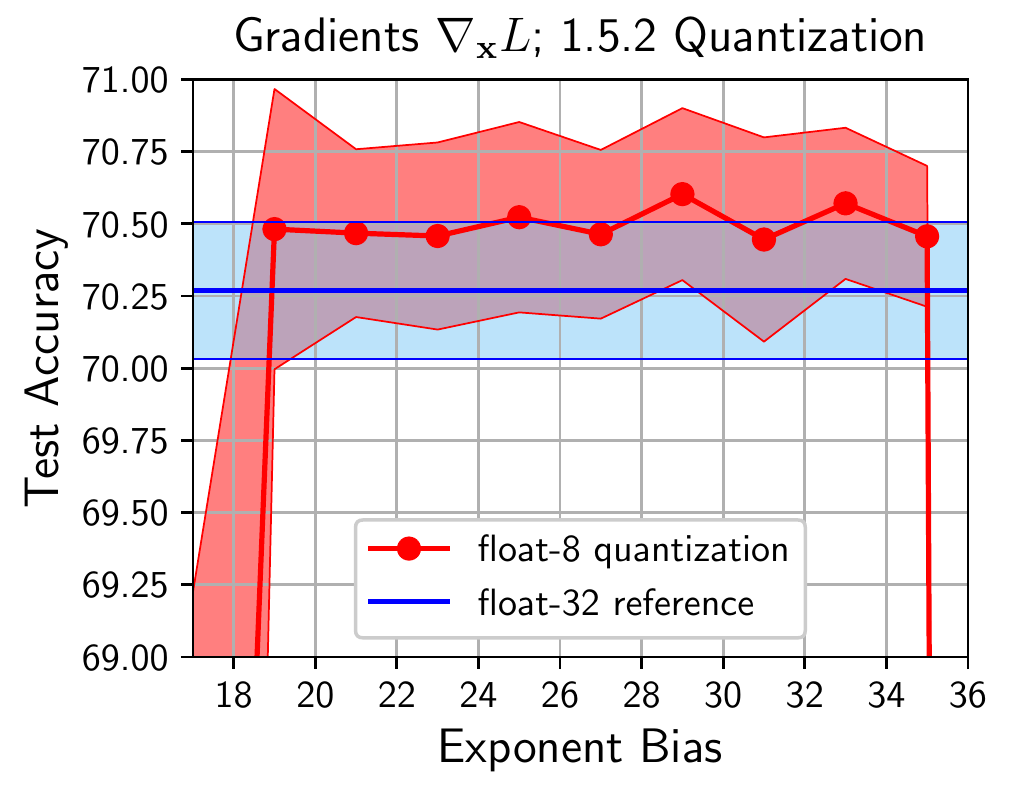}}
\subfloat[]{\includegraphics[width=0.32\columnwidth]{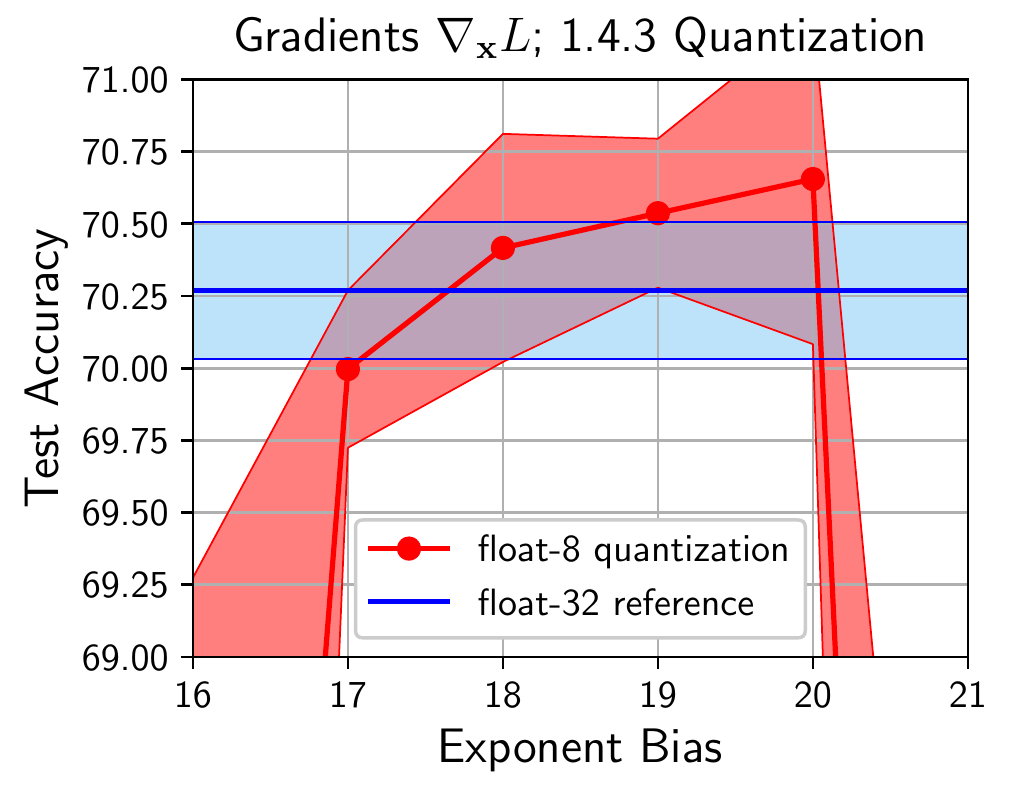}}
\subfloat[]{\includegraphics[width=0.32\columnwidth]{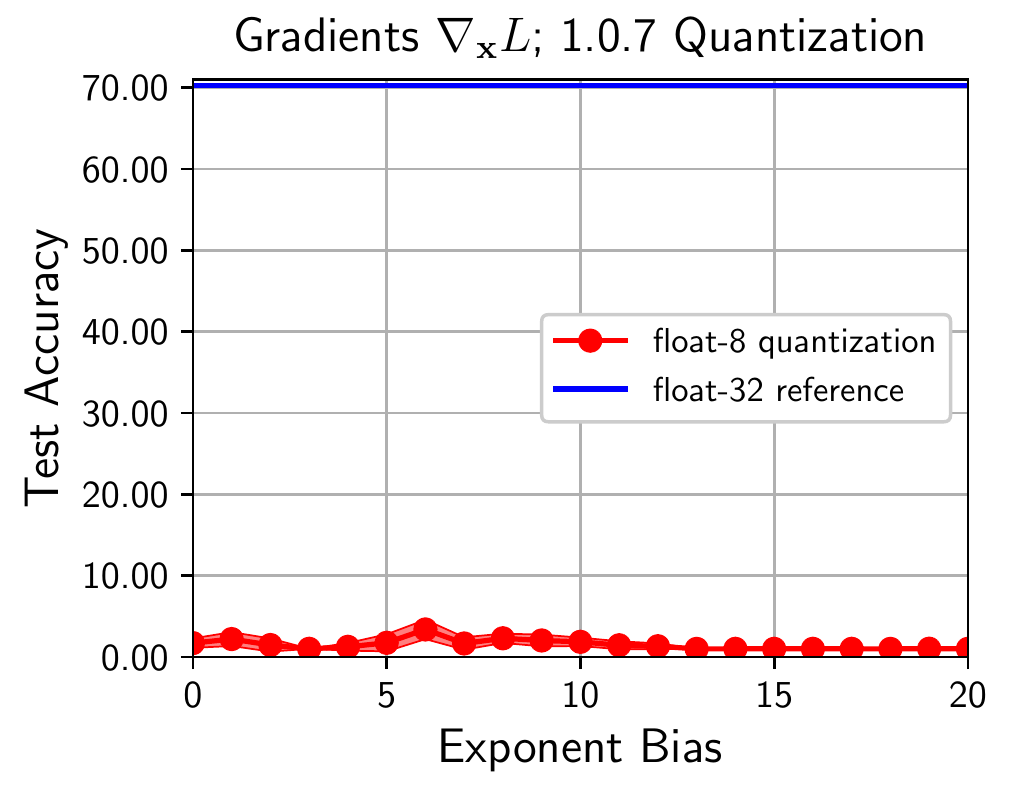}}\\
\subfloat[]{\includegraphics[width=0.32\columnwidth]{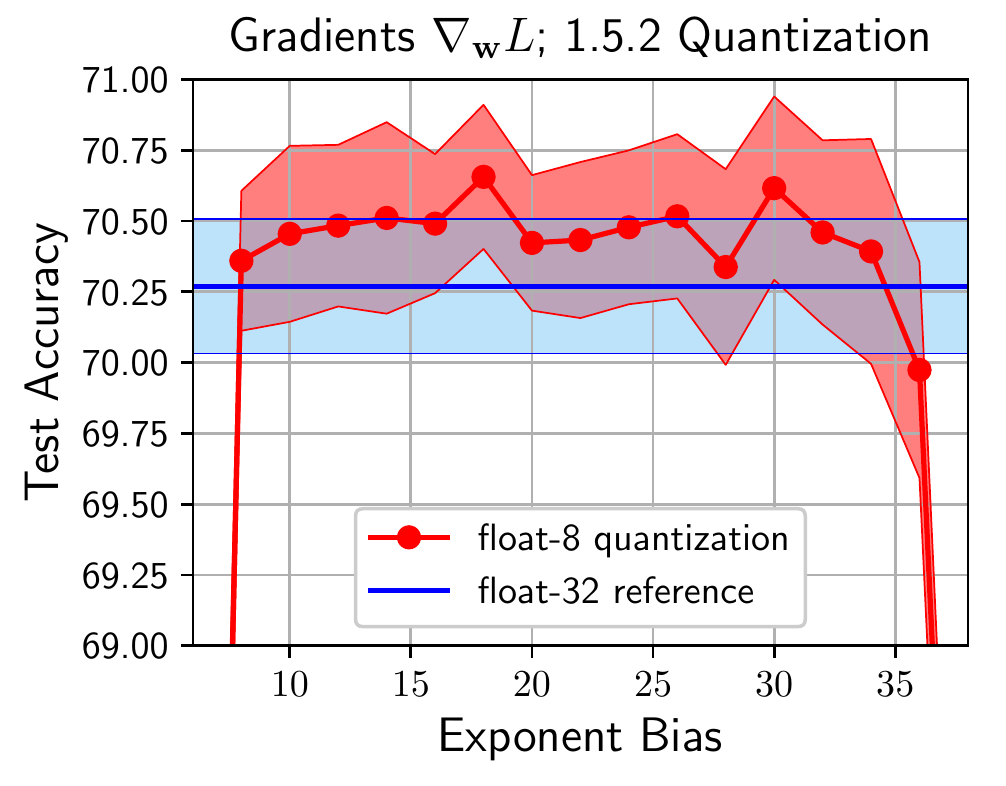}}
\subfloat[]{\includegraphics[width=0.32\columnwidth]{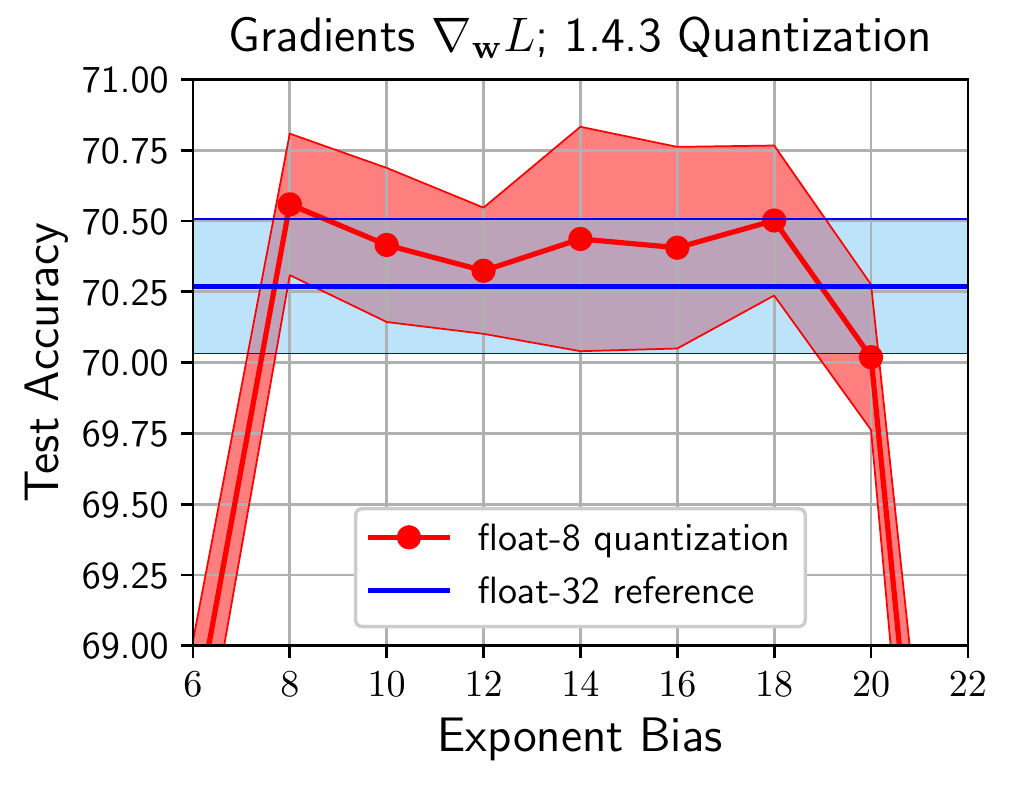}}
\subfloat[]{\includegraphics[width=0.31\columnwidth]{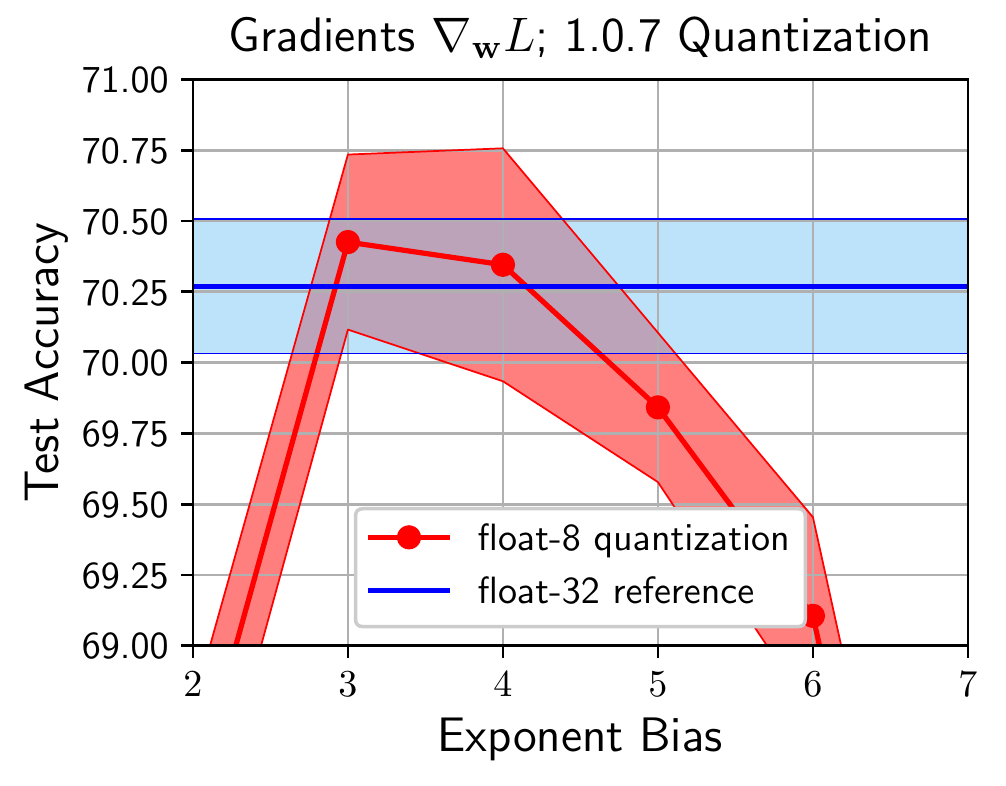}}%
\caption{ResNet-32 CIFAR-100 test performance of two different 8-bit floating-point formats compared to the scaled integer format (1.0.7) for representation of the gradients with respect to activations (a, b, c), and of the gradients with respect to weights (d, e, f). Test accuracy mean $\pm$ standard deviation over ten independent runs.}%
\label{fig:ResNet-32_CIFAR-100_GradientsAccuracy}%
\end{figure}

Figures~\ref{fig:ResNet-32_CIFAR-100_ActivationsWeightsAccuracy} and \ref{fig:ResNet-32_CIFAR-100_GradientsAccuracy} report the CIFAR-100 test performance of the floating-point formats 1.5.2, 1.4.3 and 1.0.7 (scaled integer) with a range of exponent biases for separate quantization of the activations, weights and gradients for ResNet-32 training. The test performance corresponding to the use of other 8-bit floating-point formats is reported in Appendix~\ref{sec:AppendixOtherNumberFormats}.
For each case, the figures give the performance of one of the above 8-bit formats for different values of the exponent bias.
The results have been obtained with Stochastic Gradient Descent (SGD) optimization with momentum~\citep{Poliak64} with batch size $m=32$, base learning rate $\tilde{\eta}=2^{-9}$ (learning rate $\eta=m \, \tilde{\eta}=2^{-4}$), momentum coefficient $\alpha=0.9$, and weight decay parameter $\lambda=2 \cdot 10^{-4}$. The CIFAR-100 training experiments have been run for a total of 200 epochs, with a learning rate schedule based on a reduction by a factor of 10 at 50\% and at 75\% of the total number of iterations.

The results of Figure~\ref{fig:ResNet-32_CIFAR-100_ActivationsWeightsAccuracy} show that separate quantization of weights and activations of all ResNet-32 layers, with the exception of the first layer inputs, achieves good test performance for the float-8 formats 1.5.2 and 1.4.3 for a wide range of biases.
In contrast, the usable range of bias values is greatly reduced with the 1.0.7 format for quantization of weights and activations. If we attempt to quantize {\em{all}} layer inputs including the input of the first layer, we obtain the test performances shown in Appendix~\ref{sec:AppendixOtherNumberFormats}, Figure~\ref{fig:ResNet-32_CIFAR-100_ActivationsAccuracy_Input}. From these results it is clear that 3 bits of significand are required for quantization of the first layer input.

Similarly, Figure~\ref{fig:ResNet-32_CIFAR-100_GradientsAccuracy} presents evidence that 1.5.2 and 1.4.3 quantization preserves test accuracy for the loss gradients (both gradients with respect to activations and gradients with respect to weights).  Note that for the gradients there is the option of using loss scaling to achieve the equivalent of an exponent bias shift. For gradients with respect to activations, at least 4 exponent bits are required (see also Appendix~\ref{sec:AppendixOtherNumberFormats}, Figure~\ref{fig:ResNet-32_CIFAR-100_GradientsActivationsAccuracy_Other}).

The exponent bias sweeps of Figures ~\ref{fig:ResNet-32_CIFAR-100_ActivationsWeightsAccuracy} and \ref{fig:ResNet-32_CIFAR-100_GradientsAccuracy} were then used, together with the histograms of the different quantities reported in Appendix~\ref{sec:AppendixHistograms}, to select suitable 8-bit floating-point formats for simultaneous quantization of activations, weights, and gradients with respect to activations and weights.
Table~\ref{tab:ResNet-32_CIFAR-100_WeightsActivationsGradientsAccuracy} reports the CIFAR-100 test performance for quantization of activations, weights and gradients for ResNet-32 training.
In this case, in addition to leaving the input to the first layer unquantized, it was found to be beneficial not to quantize the gradient with respect to the activations at the output of the first layer.

Table~\ref{tab:ResNet-18_ImageNet_WeightsActivationsGradientsAccuracy} reports the ImageNet test performance with 8-bit floating-point quantization of the activations, weights and gradients for ResNet-18 training. The ImageNet results have been obtained with SGD optimization with momentum with batch size $m=32$,
base learning rate $\tilde{\eta}=2^{-11}$ (learning rate $\eta=m \, \tilde{\eta}=2^{-6}$), momentum coefficient $\alpha=0.9$, and weight decay parameter $\lambda=10^{-4}$. The ImageNet training experiments have been run for a total of 100 epochs, with learning rate decay schedule based on a learning rate reduction by a factor of 10 at 30\%, 60\%, 80\% and 90\% of the total number of iterations. 
Note that using the 1.4.3 format for quantization of activations and weights and the 1.5.2 format for quantization of gradients allows us to match the reference accuracy, whereas using either the 1.4.3 format or the 1.5.2 format for all quantizations does not fully recover the baseline accuracy.

\begin{table}[htb]
\caption{ResNet-32 CIFAR-100 performance of different 8-bit floating-point formats (float-8 format, bias) for activations, weights and gradients. For the first layer, both activations and gradients with respect to activations use float-32 format. Test accuracy mean $\pm$ standard deviation over ten independent runs. The accuracy of all quantized models is statistically indistinguishable from the baseline accuracy based on a one-sided Mann-Whitney U-test~\citep{Mann47} at 5\% level of significance.}
\begin{center}
\renewcommand*{\arraystretch}{1.15}
\begin{tabular}{ l l l l l l } 
\toprule
 & \small \sc Activations & \small \sc Weights & \small \sc $\nabla_{\bf x}L$ & \small \sc $\nabla_{\bf w}L$ & \small \sc Accuracy (\%) \\
\midrule
\small \small Baseline & \small float-32 & \small float-32 & \small float-32 & \small float-32 & \small 70.26 $\pm$ 0.24 \\
\hdashline
\small 1.4.3 formats & \small 1.4.3, 10 & \small 1.4.3, 14 & \small 1.4.3, 20 & \small 1.4.3, 16 & \small 70.30 $\pm$ 0.37 \\
\small 1.5.2 formats & \small 1.5.2, 24 & \small 1.5.2, 28 & \small 1.5.2, 32 & \small 1.5.2, 33 & \small 70.02 $\pm$ 0.40 \\
\small 1.4.3 / 1.5.2 formats & \small 1.4.3, 10 & \small 1.4.3, 14 & \small 1.5.2, 33 & \small 1.5.2, 31 & \small 70.42 $\pm$ 0.45 \\
\bottomrule
\end{tabular}
\end{center}
\label{tab:ResNet-32_CIFAR-100_WeightsActivationsGradientsAccuracy}%
\end{table}

\begin{table*}[htb]
\caption{ResNet-18 ImageNet performance of different 8-bit floating-point formats (float-8 format, bias) for activations, weights and gradients. For the first layer, both activations and gradients with respect to activations use float-32 format. Validation accuracy mean $\pm$ standard deviation over five independent runs.
Asterisks indicate a difference with respect to the baseline accuracy based on a one-sided Mann-Whitney U-test~\citep{Mann47} at 5\% level of significance.}
\begin{center}
\renewcommand*{\arraystretch}{1.15}
\begin{tabular}{ l l l l l l } 
\toprule
 & \small \sc Activations & \small \sc Weights & \small \sc $\nabla_{\bf x}L$ & \small \sc $\nabla_{\bf w}L$ & \small \sc Accuracy (\%) \\
\midrule
\small \small Baseline & \small float-32 & \small float-32 & \small float-32 & \small float-32 & \small 70.35 $\pm$ 0.06 \\
\hdashline
\small 1.5.2 formats & \small 1.5.2, 28 & \small 1.5.2, 24 & \small 1.5.2, 34 & \small 1.5.2, 31 & \small 69.95 $\pm$ 0.10 * \\
\small 1.4.3 / 1.5.2 formats & \small 1.4.3, 10 & \small 1.4.3, 14 & \small 1.5.2, 34 & \small 1.5.2, 31 & \small 70.29 $\pm$ 0.03 \\
\bottomrule
\end{tabular}
\end{center}
\label{tab:ResNet-18_ImageNet_WeightsActivationsGradientsAccuracy}%
\end{table*}

\begin{table*}[htb]
\begin{minipage}[h]{\textwidth}
\caption{ResNet-50 ImageNet performance of different 8-bit floating-point formats (float-8 format, bias) for activations, weights and gradients. For the first layer, both activations and gradients with respect to activations use float-32 format. Validation accuracy mean $\pm$ standard deviation over five independent runs. Asterisks indicate a difference with respect to baseline accuracy based on a one-sided Mann-Whitney U-test~\citep{Mann47} at 5\% level of significance.}
\begin{center}
\renewcommand*{\arraystretch}{1.15}
\begin{tabular}{ l l l l l l } 
\toprule
 & \small \sc Activations & \small \sc Weights & \small \sc $\nabla_{\bf x}L$ & \small \sc $\nabla_{\bf w}L$ & \small \sc Accuracy (\%) \\
\midrule
\small Baseline & \small float-32 & \small float-32 & \small float-32 & \small float-32 & \small 76.57 $\pm$ 0.09 \\
\hdashline
\small 1.5.2 formats & \small 1.5.2, 24 & \small 1.5.2, 28 & \small 1.5.2, 32 & \small 1.5.2, 32 & \small 76.43 $\pm$ 0.09 * \\
\small 1.4.3 / 1.5.2 formats & \small 1.4.3, 10 & \small 1.4.3, 14 & \small 1.5.2, 32 & \small 1.5.2, 32 & \small 76.61 $\pm$ 0.10 \\
\bottomrule
\end{tabular}
\end{center}
\label{tab:ResNet-50_ImageNet_WeightsActivationsGradientsAccuracy}%
\end{minipage}
\end{table*}

Table~\ref{tab:ResNet-50_ImageNet_WeightsActivationsGradientsAccuracy} shows the ImageNet test performance with 8-bit floating-point quantization of the activations, weights and gradients for ResNet-50 training, for SGD with momentum with batch size $m=32$, with bias values consistent with the histograms reported in Appendix~\ref{sec:AppendixHistograms}.

Figures~\ref{fig:ResNet-32_CIFAR-100_GradientsAccuracy}(a) and~\ref{fig:ResNet-32_CIFAR-100_GradientsAccuracy}(d) show the accuracy changes with 1.5.2 format, for gradients with respect to activations and weights respectively, as the exponent bias is swept over a range of values.  In each of these plots, only a single quantity (gradients with respect to activations or with respect to weights) is quantized with 8-bit float precision, while all other quantities are kept in float-32 precision.

\begin{figure}[!bp]
\centering
{\includegraphics[width=0.7\columnwidth]{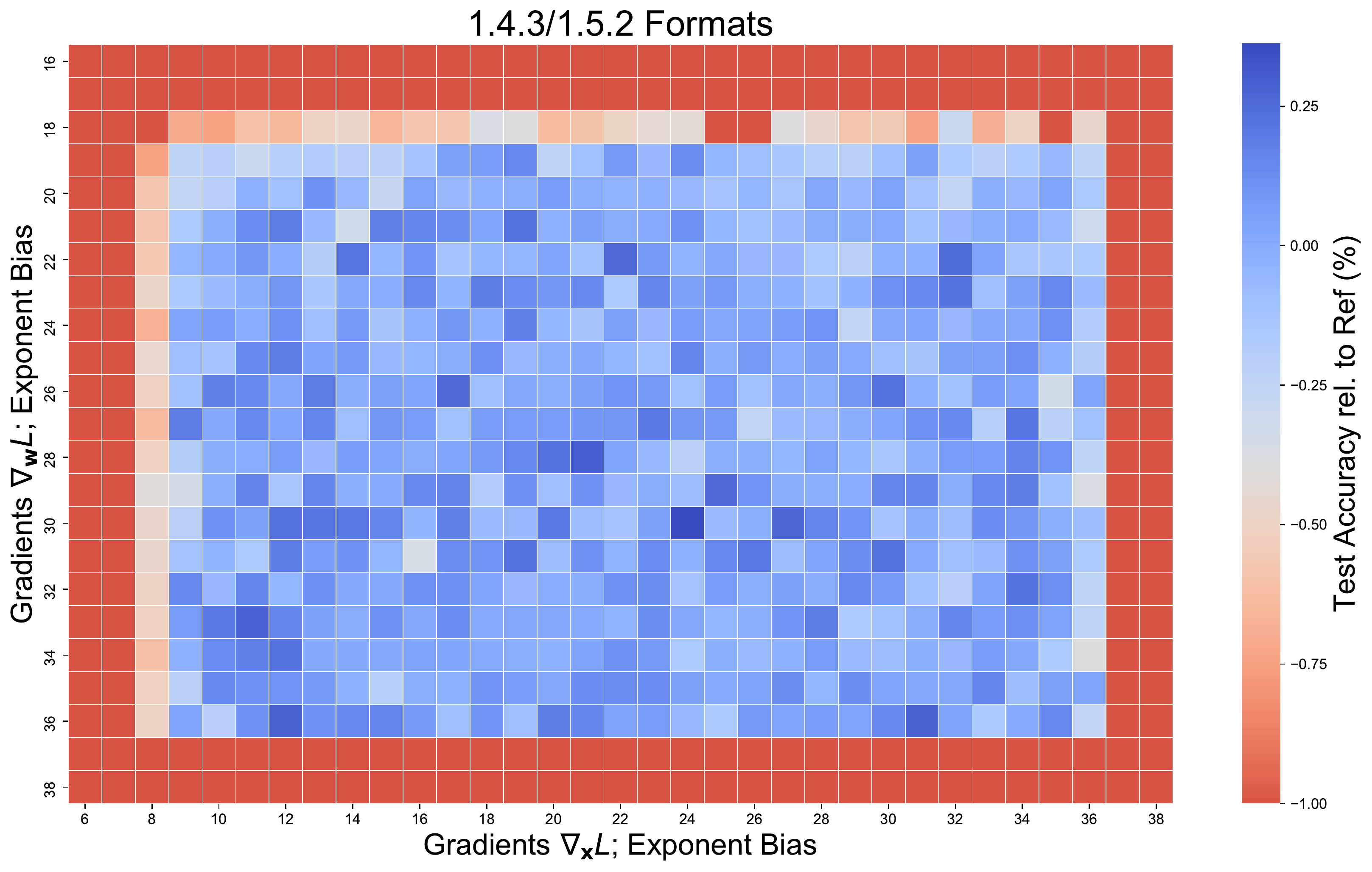}}
\caption{ResNet-32 CIFAR-100 test performance relative to reference float-32 performance.  Weights/activations quantization format 1.4.3 with biases 10, 14 respectively.  Loss gradients with respect to weights/activations quantization format 1.5.2 with biases indicated on the x and y axes.}
\label{fig:ResNet-32_CIFAR-100_GradientHeatmap}%
\end{figure}

In contrast, in Tables~\ref{tab:ResNet-32_CIFAR-100_WeightsActivationsGradientsAccuracy}--\ref{tab:ResNet-50_ImageNet_WeightsActivationsGradientsAccuracy} {\em{all}}
quantities are quantized using 8-bit floating-point formats.  It is important to confirm that the workable bias ranges of Figure~\ref{fig:ResNet-32_CIFAR-100_GradientsAccuracy} are still valid in this case.
For this purpose, Figure~\ref{fig:ResNet-32_CIFAR-100_GradientHeatmap} reports the test accuracy results of a range of bias {\em combinations} for the exponent biases applied to gradients with respect to weights and activations.  For these tests the mixed format combination 1.4.3/1.5.2 was chosen and the exponent biases for weights and activations were kept at 14 and 10 as in Table~\ref{tab:ResNet-32_CIFAR-100_WeightsActivationsGradientsAccuracy}.

As mentioned in Section~\ref{sec:Background}, Figures~\ref{fig:ResNet-32_CIFAR-100_ActivationsWeightsAccuracy}(c) and (f) and~\ref{fig:ResNet-32_CIFAR-100_GradientsAccuracy}(c) and (f) show the test performance for the floating-point number format 1.0.7.  This format is equivalent to using a {\em scaled integer}, which consists of an 8-bit signed integer together with a scaling value that specifies the range to be represented by the integer values.  This scaling factor can be represented by a suitable value of exponent bias applied to the 1.0.7 format. 
The test performance results demonstrate that it is not possible to represent the gradients with respect to activations in this format, and the other quantities have extremely narrow bias ranges over which satisfactory performance is achieved.  This is also illustrated in the histogram data of Figure{~\ref{fig:ResNet-32_CIFAR-100_Format107_AllLayer_Histogram}}, which reports the scaled integer coverage ranges for some typical bias values from Figures~\ref{fig:ResNet-32_CIFAR-100_ActivationsWeightsAccuracy}(c) and (f) and ~\ref{fig:ResNet-32_CIFAR-100_GradientsAccuracy}(c) and (f). From these results, it is evident that the scaled integer format does not provide sufficient range to represent an adequate fraction of the bins in the histogram.

It could perhaps be argued that the performance of the scaled integer could be improved by allowing the freedom of a scaling value chosen per layer. However, the considered 8-bit floating-point formats are able to match the reference float-32 performance with a {\em single} exponent bias value common to all layers.

\begin{table*}[tb]
\begin{minipage}[h]{\textwidth}
\caption{EfficientNet-B0 ImageNet performance of different 8-bit floating-point formats (float-8 format, bias) for activations, weights and gradients. For the first layer, both activations and gradients with respect to activations use float-16 format. Test accuracy mean $\pm$ standard deviation over five independent runs. Asterisks indicate a difference with respect to baseline accuracy based on a one-sided Mann-Whitney U-test~\citep{Mann47} at 5\% level of significance.}
\begin{center}
\renewcommand*{\arraystretch}{1.15}
\begin{tabular}{ l l l l l } 
\toprule
 & \small \sc Activations & \small \sc Weights & \small \sc $\nabla_{\bf x}L$ & \small \sc Accuracy (\%) \\
\midrule
\small Baseline & \small float-16 & \small float-16 & \small float-16 & \small 76.34 $\pm$ 0.18 \\
\hdashline
\small 1.4.3 / 1.5.2 formats & \small 1.4.3, 7 & \small 1.4.3, 12 & \small 1.5.2, 16 & \small 76.33 $\pm$ 0.18 \\
\small 1.4.3 / 1.5.2 formats & \small 1.4.3, 7 & \small 1.4.3, 12 & \small 1.5.2, 17 & \small 76.23 $\pm$ 0.08 \\
\small 1.4.3 / 1.5.2 formats & \small 1.4.3, 7 & \small 1.4.3, 12 & \small 1.5.2, 18 & \small 76.36 $\pm$ 0.17 \\
\bottomrule
\end{tabular}
\end{center}
\label{tab:EfficientNet-B0_ImageNet_WeightsActivationsGradientsAccuracy}%
\end{minipage}
\end{table*}

\begin{table*}[tb]
\begin{minipage}[h]{\textwidth}
\caption{EfficientNet-B2 ImageNet performance of different 8-bit floating-point formats (float-8 format, bias) for activations, weights and gradients. For the first layer, both activations and gradients with respect to activations use float-16 format. Test accuracy mean $\pm$ standard deviation over five independent runs. Asterisks indicate a difference with respect to baseline accuracy based on a one-sided Mann-Whitney U-test~\citep{Mann47} at 5\% level of significance.}
\begin{center}
\renewcommand*{\arraystretch}{1.15}
\begin{tabular}{ l l l l l } 
\toprule
 & \small \sc Activations & \small \sc Weights & \small \sc $\nabla_{\bf x}L$ & \small \sc Accuracy (\%) \\
\midrule
\small Baseline & \small float-16 & \small float-16 & \small float-16 & \small 79.42 $\pm$ 0.07 \\
\hdashline
\small 1.4.3 / 1.5.2 formats & \small 1.4.3, 7 & \small 1.4.3, 12 & \small 1.5.2, 16 & \small 79.43 $\pm$ 0.13 \\
\bottomrule
\end{tabular}
\end{center}
\label{tab:EfficientNet-B2_ImageNet_WeightsActivationsGradientsAccuracy}%
\end{minipage}
\end{table*}

\begin{table*}[tb]
\begin{minipage}[h]{\textwidth}
\caption{EfficientNet-B4 ImageNet performance of different 8-bit floating-point formats (float-8 format, bias) for activations, weights and gradients. For the first layer, both activations and gradients with respect to activations use float-16 format. Test accuracy mean $\pm$ standard deviation over five independent runs. Asterisks indicate a difference with respect to baseline accuracy based on a one-sided Mann-Whitney U-test~\citep{Mann47} at 5\% level of significance.}
\begin{center}
\renewcommand*{\arraystretch}{1.15}
\begin{tabular}{ l l l l l } 
\toprule
 & \small \sc Activations & \small \sc Weights & \small \sc $\nabla_{\bf x}L$ & \small \sc Accuracy (\%) \\
\midrule
\small Baseline & \small float-16 & \small float-16 & \small float-16 & \small 82.42 $\pm$ 0.10 \\
\hdashline
\small 1.4.3 / 1.5.2 formats & \small 1.4.3, 7 & \small 1.4.3, 12 & \small 1.5.2, 16 & \small 82.34 $\pm$ 0.10 \\
\bottomrule
\end{tabular}
\end{center}
\label{tab:EfficientNet-B4_ImageNet_WeightsActivationsGradientsAccuracy}%
\end{minipage}
\end{table*}

Tables~\ref{tab:EfficientNet-B0_ImageNet_WeightsActivationsGradientsAccuracy}-\ref{tab:EfficientNet-B4_ImageNet_WeightsActivationsGradientsAccuracy} show the ImageNet test performance with 8-bit floating-point quantization of the activations, weights and gradients for different EfficientNet models~\citep{Tan19}. Following~\citet{Tan19}, we train on ImageNet for 350 epochs with RMSProp optimization~\citep{Rmsprop12} and decay the learning rate exponentially by a factor 0.97 every 2.4 epochs. We use a weight decay parameter $\lambda=10^{-5}$ on the convolutional weights, and a label smoothing factor of 0.1. We use a slightly smaller global batch size $m=768$ across all training cases and scale the original learning rate and RMSProp decay factor. For the RMSprop optimizer we use learning rate $m \cdot 2^{-14}$,  momentum coefficient $\alpha=0.9$ and learning rate decay $1.0-m \cdot 2^{-14}$. Our final weights are obtained by using an exponentially weighted average over checkpoints from each training epoch, with decay factor $0.97$. All the accuracy results derive from an averaging over five independent runs. In addition, as presented by~\citet{Masters21}, the augmentation strategy uses a combination of Mixup~\citep{Mixup17} and CutMix~\citep{CutMix19}.
Also in this case, the accuracy obtained using the 1.4.3 format for quantization of activations and weights and the 1.5.2 format for quantization of gradients fully matches the baseline accuracy.

\subsection{Results for Language Processing }\label{sec:Results_LanguageTranslation}

The performance of 8-bit floating-point formats has also been assessed for Natural Language Processing (NLP) applications. The more recent advanced NLP models are based on the Transformer \citep{Vaswani17} and in particular its attention concept \citep{Bahdanau14}. 
Based on the model published by \citet{Kuchaiev18}, a version of the Transformer base model has been implemented with quantized activations, weights, and gradients in all fully connected layers and matrix multiplications. 
The inputs to the layer normalization and softmax layers are left unquantized. The model performance for language translation has been evaluated on an English-German translation task using the WMT14 dataset. For each experiment, the model has been trained for 400,000 iterations with a batch size of 128 sentence pairs, using Adam optimization \citep{Kingma14} with a base learning rate $\tilde{\eta}=2/\sqrt{512}$, optimizer parameters $\beta_1=0.9$, $\beta_2=0.997$ and $\epsilon=10^{-9}$, and a learning rate schedule with linear increase of the learning rate during 8,000 warm-up steps and learning rate decay proportional to $1/\sqrt{step+1}$ afterwards \citep{Kuchaiev18}. 
To reproduce the performance of Transformer models reported in the literature \citep{Vaswani17}, in Section \ref{sec:AutomaticLossScaling} we also show the performance results of both the float-32 and float-8 models trained for 100,000 iterations with a batch size of 1024 sentence pairs. Similar to \citet{Vaswani17} for these runs we used checkpoint-averaging to obtain the final weights used for testing. 

\begin{figure}[!tb]
\centering
\subfloat[]{\includegraphics[width=0.34\columnwidth]{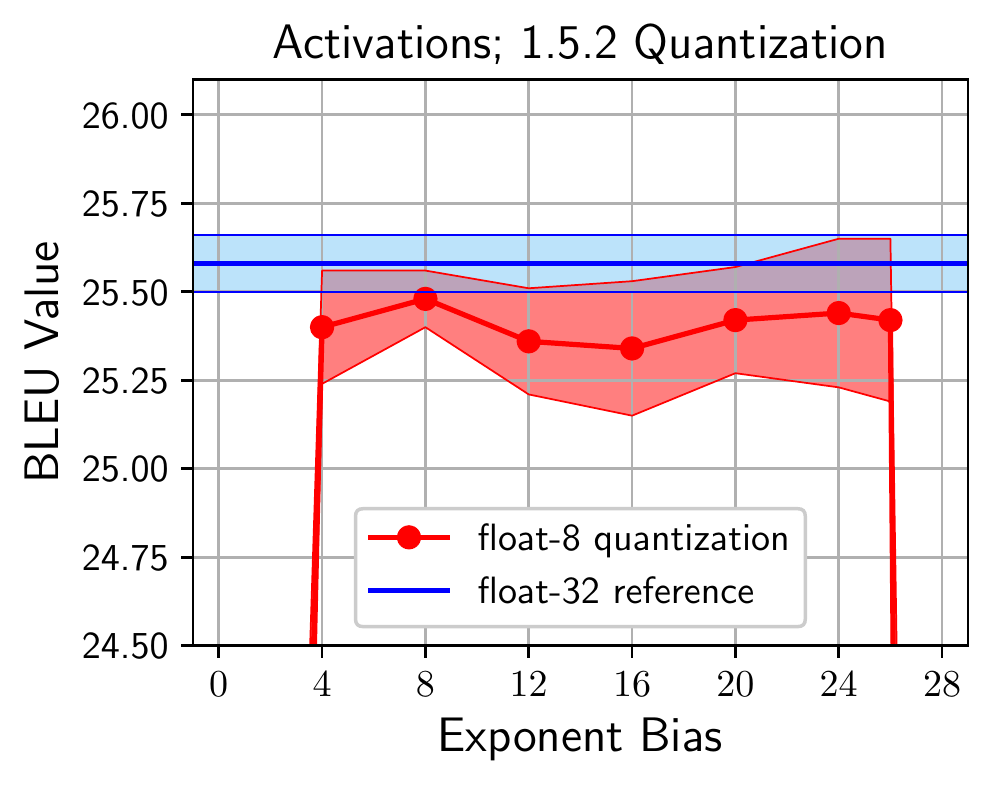}}
\subfloat[]{\includegraphics[width=0.34\columnwidth]{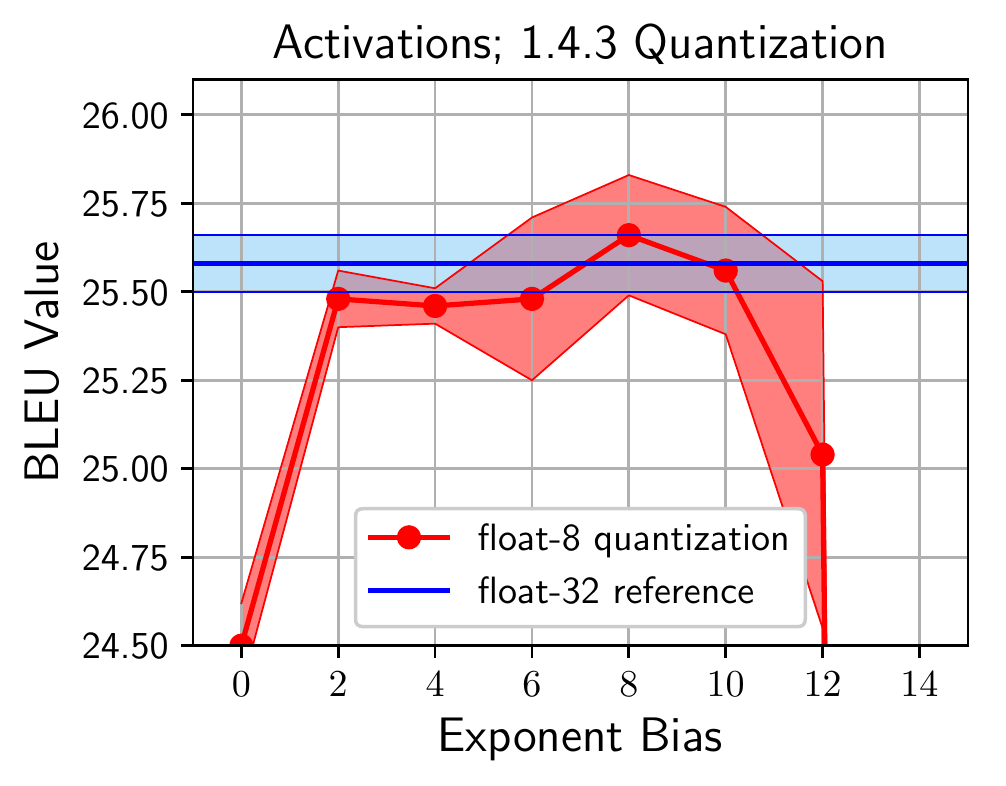}}\\
\subfloat[]{\includegraphics[width=0.34\columnwidth]{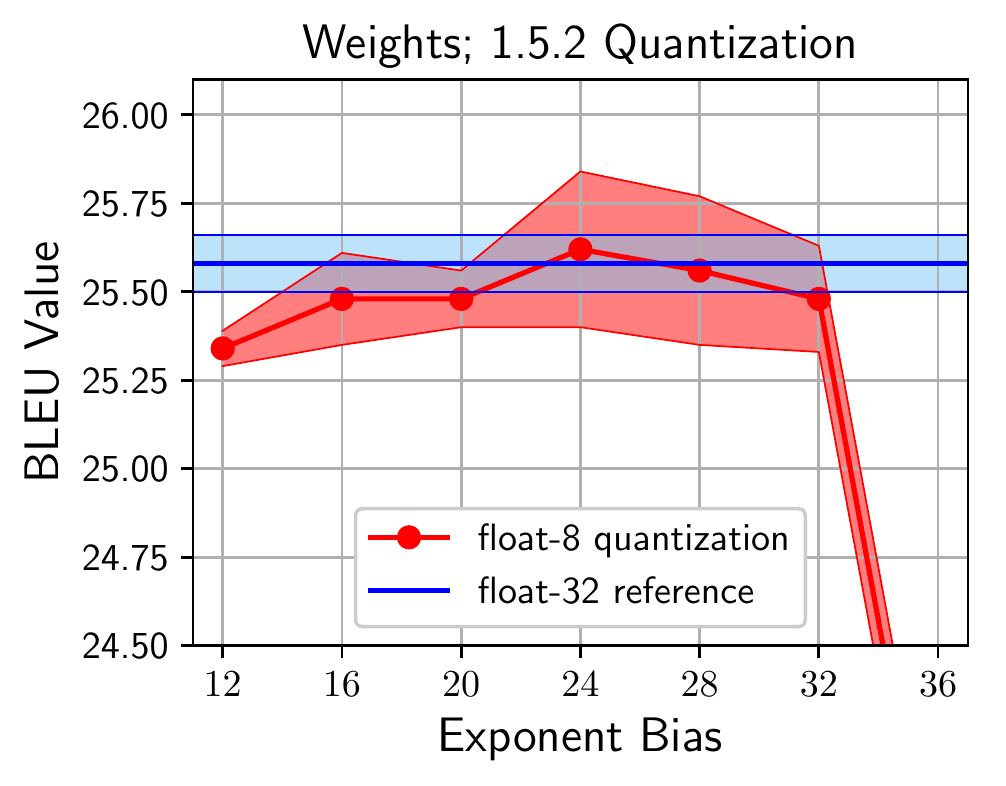}}
\subfloat[]{\includegraphics[width=0.34\columnwidth]{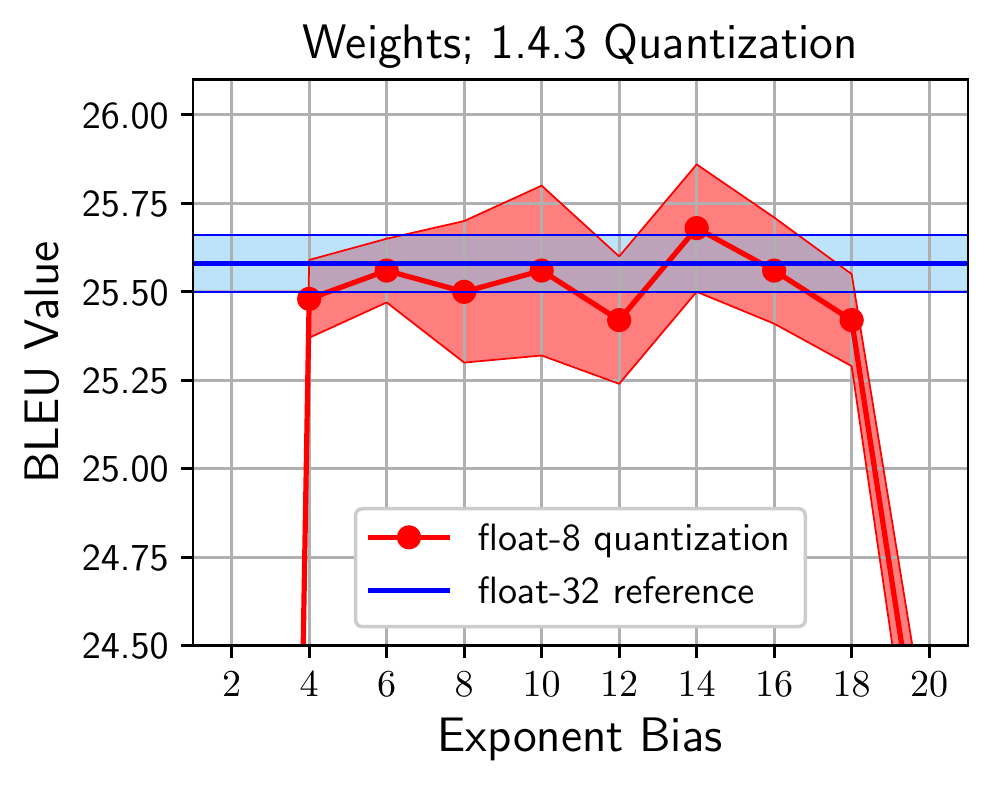}}
\caption{Transformer WMT14 English-German translation performance of different 8\nobreakdashes-bit floating-point formats for representation of the activations (a, b), and the weights (c, d). BLEU mean $\pm$ standard deviation over five independent runs.}
\label{fig:Transformer_WMT14_ActivationsWeightsAccuracy}%
\end{figure}

\begin{figure}[!t]
\centering
\subfloat[]{\includegraphics[width=0.34\columnwidth]{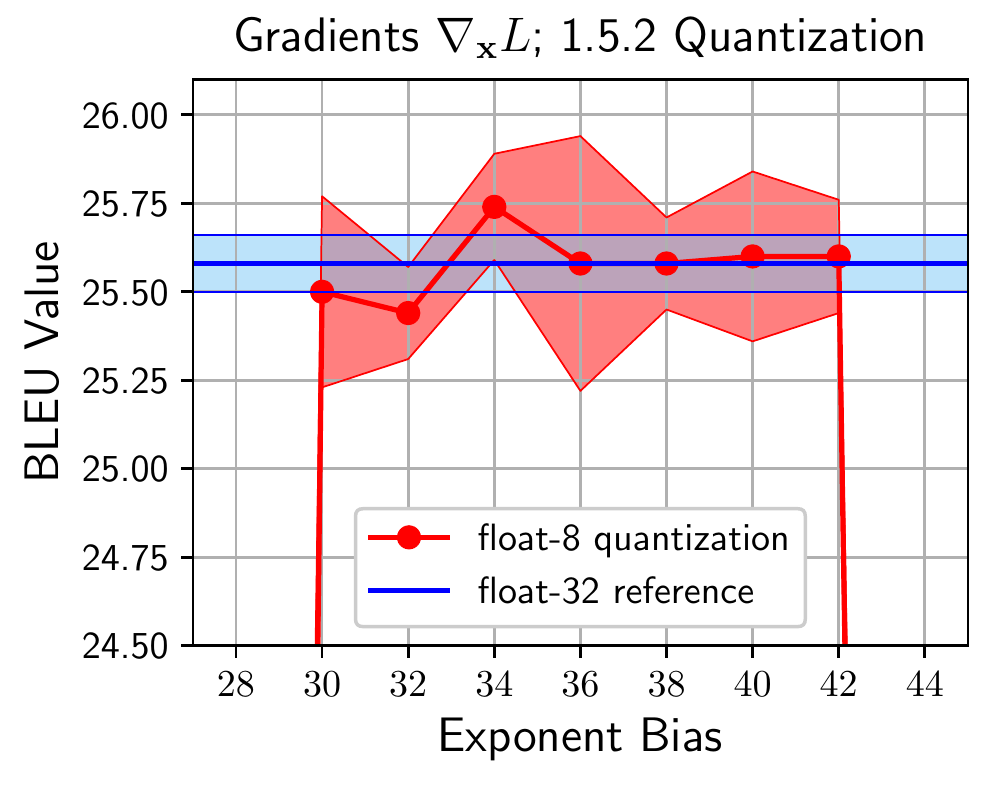}}
\subfloat[]{\includegraphics[width=0.34\columnwidth]{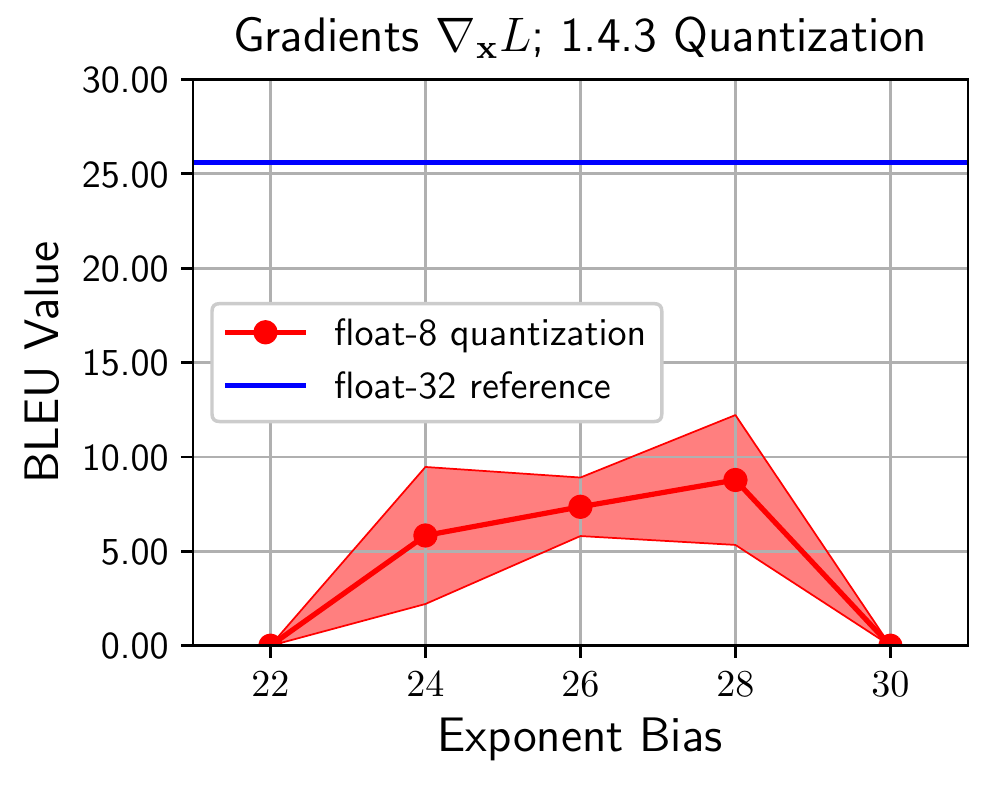}}\\
\subfloat[]{\includegraphics[width=0.34\columnwidth]{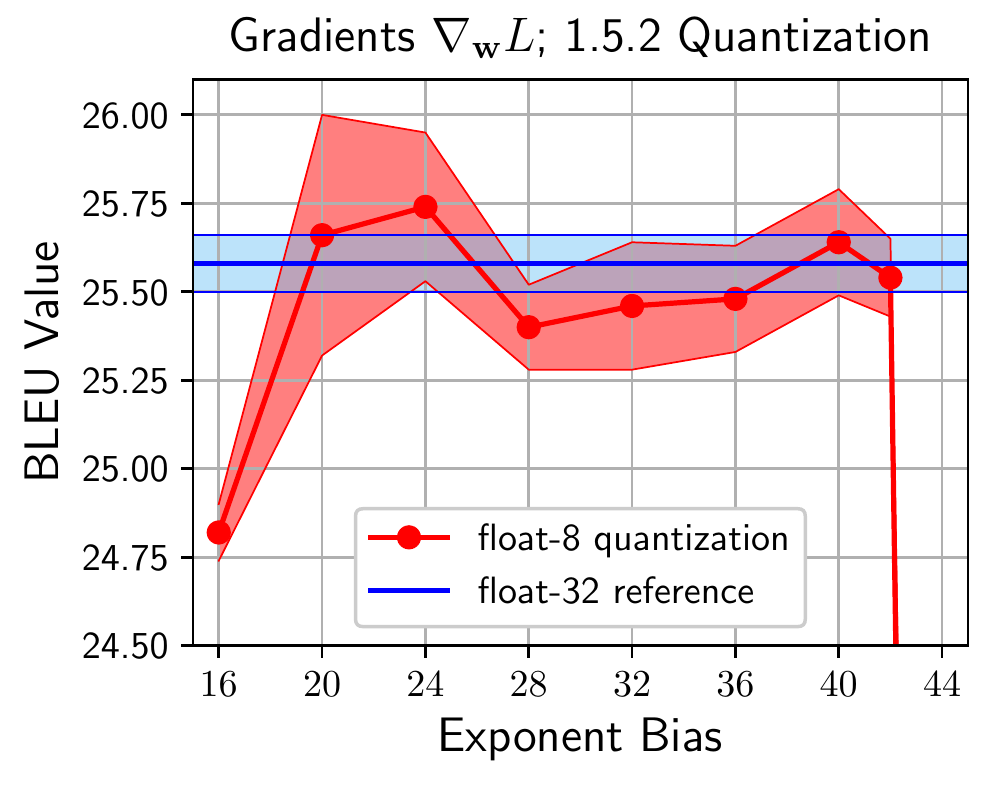}}
\subfloat[]{\includegraphics[width=0.34\columnwidth]{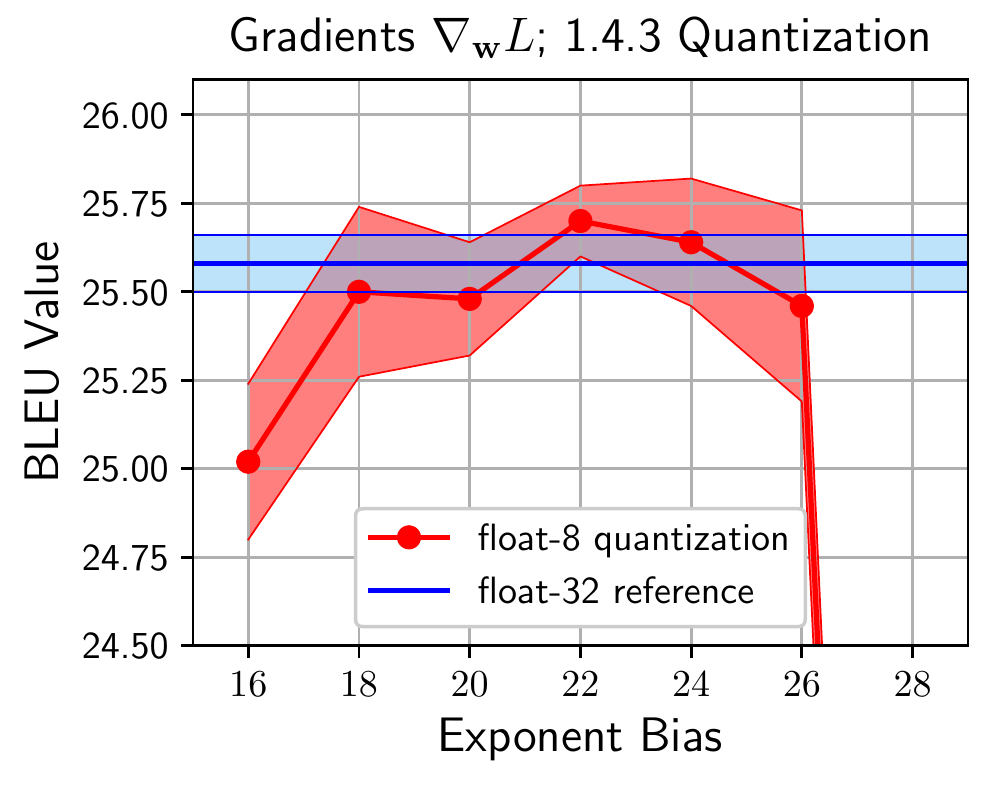}}
\caption{Transformer WMT14 English-German translation performance of different 8\nobreakdashes-bit floating-point formats for representation of the gradients with respect to activations (a, b), and the gradients with respect to weights (c, d). BLEU mean $\pm$ standard deviation over five independent runs.}
\label{fig:Transformer_WMT14_GradientsAccuracy}%
\end{figure}

The final BLEU scores\footnote{All BLEU scores reported in this paper are case insensitive, calculated on the WMT14 official evaluation data \citep{Bojar14} using sacreBLEU~\citep{Post18}. The tokenization followed the Moses mteval-v14 implementation (https://github.com/moses-smt/mosesdecoder/blob/master/scripts/generic/mteval-v14.pl).} are presented for the model with separate quantization of activations (Figure~\ref{fig:Transformer_WMT14_ActivationsWeightsAccuracy}(a) and (b)), weights (Figure~\ref{fig:Transformer_WMT14_ActivationsWeightsAccuracy}(c) and (d)), gradients with respect to activations (Figure~\ref{fig:Transformer_WMT14_GradientsAccuracy}(a) and (b)), and gradients with respect to weights (Figure~\ref{fig:Transformer_WMT14_GradientsAccuracy}(c) and (d)), each using 1.5.2 and 1.4.3 number formats with a range of different exponent biases.

Figure~\ref{fig:Transformer_WMT14_ActivationsWeightsAccuracy}(a) suggests a small degradation in the model accuracy when quantizing the activations of the Transformer using the 1.5.2 format. This is an effect of the limited SNR of the 1.5.2 format, since the 1.4.3 format with its higher SNR yields an accuracy on par with the float-32 baseline for a range of different exponent biases (Figure~\ref{fig:Transformer_WMT14_ActivationsWeightsAccuracy}(b)). The latter format is therefore preferred for quantizing the Transformer activations.

The results shown in Figures~\ref{fig:Transformer_WMT14_ActivationsWeightsAccuracy}(c) and (d) and~\ref{fig:Transformer_WMT14_GradientsAccuracy}(c) and (d) indicate that both 1.5.2 and 1.4.3 number formats are suitable for quantizing weights and gradients with respect to weights of the Transformer base model without any loss in accuracy. These experiments show a wide range of exponent biases which achieve a model accuracy that is statistically indistinguishable from the float\nobreakdash-32 baseline. Figure~\ref{fig:Transformer_WMT14_GradientsAccuracy}(a) and (b) show that, whilst quantizing gradients with respect to activations using the 1.5.2 format yields a good model accuracy, a significant degradation occurs if the 1.4.3 format is used. This demonstrates that the quantization of the gradient with respect to activations requires a dynamic range of at least 5 exponent bits to maintain the float\nobreakdash-32 baseline performance, which is in line with the results reported in~\citet{Sun19}.

For all experiments, especially for the quantization of gradients, we observe that exponent biases larger than the "natural" bias (bias 7 for 1.4.3 and bias 15 for 1.5.2) yield a better performance of the quantized model. A large exponent bias improves the representation of values close to zero. Since the absolute values of activations, weights and gradients tend to be small (see the histograms of Appendix~\ref{sec:AppendixHistograms}, Figures~\ref{fig:Transformer_HistogramActivations}--\ref{fig:Transformer_HistogramGradWeights}) this allows for an optimal use of the available value range (Appendix~\ref{sec:AppendixHistograms}, Figure~\ref{fig:Transformer_QuantizedHistogram}).

Based on the above results we have selected individually well-performing 8\nobreakdash-bit floating-point formats for activations, weights, gradients with respect to activations, and gradients with respect to weights, to be jointly tested in a fully quantized model. As shown in Table~\ref{tab:Transformer_WMT14_WeightsActivationsGradientsAccuracy}, with a suitable choice of the number format, the performance of a fully quantized model is on a par with the float\nobreakdash-32 baseline. Both 1.5.2 and 1.4.3 formats can be used to represent activations and weights, with no clear advantage of one format over the other.

We observe that not only does the accuracy of a quantized model match the performance of the baseline, but also the distribution of the weights of each layer is -- within the range of representable values -- identical to the distribution of the weights of the baseline model (see Appendix~\ref{sec:AppendixHistograms}, Figure~\ref{fig:Transformer_HistogramWeightsFullyQuantized}). This indicates that, even though weights, activations and gradients have been quantized to 8\nobreakdash-bit floating-point values during the entire training process, a model is learned that is consistent with the float\nobreakdash-32 baseline model.

\begin{table*}[!tb]
\caption{Transformer WMT14 English-German translation performance of different 8\nobreakdash-bit floating-point formats (float-8 format, bias) for activations, weights and gradients. BLEU mean $\pm$ standard deviation over five independent runs. The accuracy of all quantized models is statistically indistinguishable from the baseline accuracy based on a one-sided Mann-Whitney U-test~\citep{Mann47} at 5\% level of significance.}
\begin{center}
\renewcommand*{\arraystretch}{1.15}
\begin{tabular}{ l l l l l l } 
\toprule
 & \small \sc Activations & \small \sc Weights & \small \sc $\nabla_{\bf x}L$ & \small \sc $\nabla_{\bf w}L$ & \small \sc BLEU \\
\midrule
\small Baseline & \small float-32 & \small float-32 & \small float-32 & \small float-32 & \small 25.58 $\pm$ 0.08 \\
\hdashline
\small 1.5.2 formats& \small 1.5.2, 24 & \small 1.5.2, 28 & \small 1.5.2, 40 & \small 1.5.2, 40 & \small 25.48 $\pm$ 0.26 \\
\small 1.4.3 / 1.5.2 formats& \small 1.4.3, 8 & \small 1.4.3, 14 & \small 1.5.2, 40 & \small 1.5.2, 40 & \small 25.52 $\pm$ 0.22 \\
\small 1.4.3 / 1.5.2 formats& \small 1.4.3, 10 & \small 1.4.3, 16 & \small 1.5.2, 40 & \small 1.5.2, 40 & \small 25.56 $\pm$ 0.23 \\
\bottomrule
\end{tabular}
\end{center}
\label{tab:Transformer_WMT14_WeightsActivationsGradientsAccuracy}%
\end{table*}

In Tables~\ref{tab:BERT-Large_Phase1_WeightsActivationsGradientsAccuracy}-\ref{tab:BERT-Large_SQuAD_WeightsActivationsGradientsAccuracy} we report the results for quantizing the different matrix multiplication layers in BERT-Large using float-8, for both pre-training and fine-tuning~\citep{BERT18}. Each experiment consists of two pre-training phases and a fine-tuning phase comprising multiple training runs, started from the pre-trained model. All phases use the AdamW optimizer ~\citep{AdamW17}, with $\beta_{1} = 0.9$, $\beta_{2} = 0.999$ and $\epsilon=10^{-6}$. The learning rate follows a linear warm-up decay schedule, whereby the warmup phase lasts for the minimum of either $10^{4}$ steps or a tenth of total number of steps.
Pre-training phase one optimizes the Masked Language Model (MLM) and Next-Sentence Prediction (NSP) loss for corrupted sentence pairs. Masked and padded sequences of length 128 are grouped into batches of approximately 512 sequences. The model is trained for 10 epochs of Wikipedia~\citep{Wiki16} + BookCorpus~\citep{Books15}, corresponding to approximately $8\cdot10^5$ optimizer steps. For all experiments, the learning rate is set to the largest value that maintains stable convergence. Pre-training phase two uses sequence length 384, 5 epochs, and approximately $2\cdot10^{5}$ optimization steps.

\begin{table*}[ht]
\begin{minipage}[h]{\textwidth}
\caption{BERT-Large pre-training phase one performance of different 8-bit floating-point formats (float-8 format, bias) for activations, weights and gradients. MLM+NSP test loss mean $\pm$ standard deviation over five independent runs. Asterisks indicate a difference with respect to baseline accuracy based on a one-sided Mann-Whitney U-test~\citep{Mann47} at 5\% level of significance.}
\begin{center}
\renewcommand*{\arraystretch}{1.15}
\begin{tabular}{ l l l l l } 
\toprule
 & \small \sc Activations & \small \sc Weights & \small \sc $\nabla_{\bf x}L$ & \small \sc Test Loss \\
\midrule
\small Baseline & \small float-16 & \small float-16 & \small float-16 & \small 2.19 $\pm$ 0.02 \\
\hdashline
\small 1.4.3 / 1.5.2 formats & \small 1.4.3, 10 & \small 1.4.3, 14 & \small 1.5.2, 16 & \small 2.22 $\pm$ 0.01 \\
\small 1.4.3 / 1.5.2 formats & \small 1.4.3, 10 & \small 1.4.3, 14 & \small 1.5.2, 18 & \small 2.23 $\pm$ 0.01 \\
\small 1.4.3 / 1.5.2 formats & \small 1.4.3, 10 & \small 1.4.3, 14 & \small 1.5.2, 20 & \small 2.22 $\pm$ 0.02 \\
\small 1.4.3 / 1.5.2 formats & \small 1.4.3, 10 & \small 1.4.3, 14 & \small 1.5.2, 24 & \small 2.22 $\pm$ 0.01 * \\
\bottomrule
\end{tabular}
\end{center}
\label{tab:BERT-Large_Phase1_WeightsActivationsGradientsAccuracy}%
\end{minipage}
\end{table*}

\begin{table*}[ht]
\begin{minipage}[h]{\textwidth}
\caption{BERT-Large pre-training phase two performance of different 8-bit floating-point formats (float-8 format, bias) for activations, weights and gradients. MLM+NSP test loss mean $\pm$ standard deviation over five independent runs. Asterisks indicate a difference with respect to baseline accuracy based on a one-sided Mann-Whitney U-test~\citep{Mann47} at 5\% level of significance.}
\begin{center}
\renewcommand*{\arraystretch}{1.15}
\begin{tabular}{ l l l l l } 
\toprule
 & \small \sc Activations & \small \sc Weights & \small \sc $\nabla_{\bf x}L$ & \small \sc Test Loss \\
\midrule
\small Baseline & \small float-16 & \small float-16 & \small float-16 & \small 1.85 $\pm$ 0.009 \\
\hdashline
\small 1.4.3 / 1.5.2 formats & \small 1.4.3, 10 & \small 1.4.3, 14 & \small 1.5.2, 16 & \small 1.86 $\pm$ 0.023 \\
\small 1.4.3 / 1.5.2 formats & \small 1.4.3, 10 & \small 1.4.3, 14 & \small 1.5.2, 18 & \small 1.90 $\pm$ 0.042 * \\
\small 1.4.3 / 1.5.2 formats & \small 1.4.3, 10 & \small 1.4.3, 14 & \small 1.5.2, 20 & \small 1.87 $\pm$ 0.002 * \\
\small 1.4.3 / 1.5.2 formats & \small 1.4.3, 10 & \small 1.4.3, 14 & \small 1.5.2, 24 & \small 1.92 $\pm$ 0.006 * \\
\bottomrule
\end{tabular}
\end{center}
\label{tab:BERT-Large_Phase2_WeightsActivationsGradientsAccuracy}%
\end{minipage}
\end{table*}

\begin{table*}[ht]
\begin{minipage}[h]{\textwidth}
\caption{BERT-Large fine-tuning SQuAD performance of different 8-bit floating-point formats (float-8 format, bias) for activations, weights and gradients used for both pre-training and fine-tuning. F1 Score and Exact Match mean $\pm$ standard deviation over five independent runs. Asterisks indicate a difference with respect to baseline accuracy based on a one-sided Mann-Whitney U-test~\citep{Mann47} at 5\% level of significance.}
\begin{center}
\renewcommand*{\arraystretch}{1.15}
\begin{tabular}{ l l l l l l} 
\toprule
 & \small \sc Activations & \small \sc Weights & \small \sc $\nabla_{\bf x}L$ & \small \sc F1 Score & \small \sc Exact Match \\
\midrule
\small Baseline & \small float-16 & \small float-16 & \small float-16 & \small 90.51 $\pm$ 0.08 & \small 83.42 $\pm$ 0.12 \\
\hdashline
\small 1.4.3 / 1.5.2 formats & \small 1.4.3, 10 & \small 1.4.3, 14 & \small 1.5.2, 16 & \small 90.69 $\pm$ 0.16 & \small 83.40 $\pm$ 0.23 \\
\bottomrule
\end{tabular}
\end{center}
\label{tab:BERT-Large_SQuAD_WeightsActivationsGradientsAccuracy}%
\end{minipage}
\end{table*}

\newpage

\section{Automatic Loss Scaling} \label{sec:AutomaticLossScaling}
As already mentioned in Section~\ref{sec:Results_ImageClassification}, the same effect as changing the exponent bias for 8-bit representations of the gradients can be obtained by loss scaling. Instead of choosing a large exponent bias to prevent underflow, this method consists in scaling up the loss after the forward pass and scaling down the learning rate by the same factor before updating the weights or otherwise absorbing the scaling factor in the optimizer, as discussed in Section~\ref{sec:LossScaling}. We have tested algorithms that automate the selection of the loss scaling factor, to overcome the necessity to manually tune the bias for a low precision representation of the gradients. An adaptive loss scaling factor also allows one to react to changes in the long term statistics of the magnitudes of gradient components during training. Our experiments with different biases for quantization of the gradients suggest that the optimal range of representable values covers, or slightly clips, the largest occurring values and extends as far as possible to small values (see Appendix~\ref{sec:AppendixHistograms}, Figure \ref{fig:Transformer_QuantizedHistogram}). Therefore, we have considered two algorithms that aim at using a factor that scales the maximal gradient values to the maximum of the representable range: {\it a)} {\em Backoff} loss scaling and {\it b)} {\em LogMax} loss scaling \citep{Kuchaiev18}.

For the set of experiments with Backoff loss scaling, we disable the clipping of gradients upon overflow and return \texttt{NaN} instead. When this happens, Backoff skips the weight update and reduces the loss scaling coefficient by a factor of 2. Then, whenever no overflow occurs for 2,000 consecutive iterations, the algorithm increases the loss scaling coefficient by a factor of 2.

The LogMax algorithm estimates the mean $\mu$ and standard deviation $\sigma$ of the quantity
$ \,\log_2\left[\, \max\left(| \nabla_{\bf w}L |\right) \, \right] \,$ for every mini-batch, and scales the loss such that $\mu+c\sigma$ equals $\log_2$ of the maximum representable value for some constant $c$. We have tested the performance of LogMax loss scaling with different multiples of the estimated standard deviation and with clipping of the gradients upon overflow.

For WMT14 English-German translation, better results are obtained with large loss scaling factors that make it possible to represent small gradients, and in the case of LogMax can lead to a moderate clipping of gradients (see Table~\ref{tab:Transformer_WMT14_AutomaticLossScale_WeightsActivationsGradientsAccuracy}). We find that a model with 1.5.2 activations, weights and gradients reaches the same performance as our float\nobreakdash-32 baseline when Backoff is used to scale the loss. When using 1.4.3 quantized activations and weights with the natural bias of 7, we see a degradation in the final BLEU score that could be recovered by hand tuning the respective biases. A quantized model using the 1.4.3 format for representing the gradients diverged in all cases (Table~\ref{tab:Transformer_WMT14_AutomaticLossScale_WeightsActivationsGradientsAccuracy}). This is in agreement with our finding that at least 5 exponent bits are required to represent the gradient with respect to the activations $\nabla_{\bf x}L$ (Figure~\ref{fig:Transformer_WMT14_GradientsAccuracy}(a) and (b)).

\begin{table*}[!tb]
\caption{Effect of different automatic loss scaling algorithms on the Transformer WMT14 English-German translation performance with 8-bit floating-point formats (float-8 format, bias) for activations, weights and gradients. BLEU mean $\pm$ standard deviation over five independent runs. Asterisks indicate a difference to baseline accuracy based on a one-sided Mann-Whitney U-test~\citep{Mann47} at 5\% level of significance.}
\begin{center}
\renewcommand*{\arraystretch}{1.15}
\begin{tabular}{ l l l l l l } 
\toprule
 & \small \sc Activations & \small \sc Weights & \small \sc $\nabla_{\bf x}L$ & \small \sc $\nabla_{\bf w}L$ & \small \sc BLEU \\
\midrule
\small Baseline & \small float-32 & \small float-32 & \small float-32 & \small float-32 & \small 25.58 $\pm$ 0.08 \\
\hdashline
\small Backoff 1.5.2 & \small 1.5.2, 15 & \small 1.5.2, 15 & \small 1.5.2, 15 & \small 1.5.2, 15 & \small 25.50 $\pm$ 0.24 \\
\small Backoff 1.5.2& \small 1.5.2, 24 & \small 1.5.2, 28 & \small 1.5.2, 15 & \small 1.5.2, 15 & \small 25.48 $\pm$ 0.18 \\
\small Backoff 1.4.3 / 1.5.2& \small 1.4.3, 7 & \small 1.4.3, 7 & \small 1.5.2, 15 & \small 1.5.2, 15 & \small 25.42 $\pm$ 0.13 * \\
\small Backoff 1.4.3 / 1.5.2& \small 1.4.3, 10 & \small 1.4.3, 16 & \small 1.5.2, 15 & \small 1.5.2, 15 & \small 25.50 $\pm$ 0.19 \\
\small Backoff 1.4.3& \small 1.4.3, 10 & \small 1.4.3, 16 & \small 1.4.3, 7 & \small 1.4.3, 7 & \small 0.00 $\pm$ 0.00\footref{footnote:divergence} * \\
\small Backoff 1.4.3 / 1.5.2& \small 1.5.2, 24 & \small 1.5.2, 28 & \small 1.4.3, 7 & \small 1.4.3, 7 & \small 0.00 $\pm$ 0.00\footref{footnote:divergence} * \\
\hdashline
\small LogMax 1.5.2, $3\sigma$ & \small 1.5.2, 24 & \small 1.5.2, 28 & \small 1.5.2, 15 & \small 1.5.2, 15 & \small 25.34 $\pm$ 0.26\\
\small LogMax 1.5.2, $2\sigma$ & \small 1.5.2, 24 & \small 1.5.2, 28 & \small 1.5.2, 15 & \small 1.5.2, 15 & \small 24.94 $\pm$ 0.42 *\\
\small LogMax 1.5.2, $1\sigma$ & \small 1.5.2, 24 & \small 1.5.2, 28 & \small 1.5.2, 15 & \small 1.5.2, 15 & \small 25.12 $\pm$ 0.23 *\\
\small LogMax 1.5.2, $0\sigma$ & \small 1.5.2, 24 & \small 1.5.2, 28 & \small 1.5.2, 15 & \small 1.5.2, 15 & \small 25.30 $\pm$ 0.29 \\
\bottomrule
\end{tabular}
\end{center}
\label{tab:Transformer_WMT14_AutomaticLossScale_WeightsActivationsGradientsAccuracy}%
\end{table*}

\stepcounter{footnote}\footnotetext{\label{footnote:divergence}~Model did not converge.}

From Table~\ref{tab:Transformer_WMT14_AutomaticLossScale_WeightsActivationsGradientsAccuracy} and Appendix~\ref{AppendixLossScalingFactor}, Figure~\ref{fig:LossScalingFactor}, the LogMax algorithm yields the highest BLEU scores when using a factor of $c=0$ (LogMax, $\! 0\sigma$ in Table~\ref{tab:Transformer_WMT14_AutomaticLossScale_WeightsActivationsGradientsAccuracy}) and thereby employing a large loss scaling factor. In this case its performance is similar to that of our baseline model.

The automatic loss scaling algorithms with the best performance, Backoff and LogMax with constant $c=0$, result in a loss scaling factor of about $2^{22}$, together with a gradient representation based on the 8-bit floating-point format 1.5.2 with bias 15 (see Appendix~\ref{AppendixLossScalingFactor}, Figure \ref{fig:LossScalingFactor}). This is equivalent to using a bias of 37 for gradients~$\nabla_{\mathbf{x}}L$ and gradients~$\nabla_{\mathbf{w}}L$ and no loss scaling, putting these results in line with our experiments using different biases in Figure~\ref{fig:Transformer_WMT14_GradientsAccuracy}.

\begin{table*}[tb]
\caption{Transformer WMT14 English-German translation performance of different 8-bit floating-point formats (float-8 format, bias) for activations, weights and gradients after training for 100,000 iterations with batch size 1024. BLEU mean $\pm$ standard deviation over five independent runs. Asterisks indicate a difference to baseline accuracy based on a one-sided Mann-Whitney U-test~\citep{Mann47} at 5\% level of significance.}
\begin{center}
\renewcommand*{\arraystretch}{1.15}
\begin{tabular}{ l l l l l l } 
\toprule
 & \small \sc Activations & \small \sc Weights & \small \sc $\nabla_{\bf x}L$ & \small \sc $\nabla_{\bf w}L$ & \small \sc BLEU \\
\midrule
\small Baseline & \small float-32 & \small float-32 & \small float-32 & \small float-32 & \small 27.04 $\pm$ 0.11 \\
\hdashline
\small 1.5.2 formats& \small 1.5.2, 24 & \small 1.5.2, 28 & \small 1.5.2, 40 & \small 1.5.2, 40 & \small 26.94 $\pm$ 0.11 \\
\small 1.4.3 / 1.5.2 formats& \small 1.4.3, 10 & \small 1.4.3, 16 & \small 1.5.2, 40 & \small 1.5.2, 40 & \small 26.88 $\pm$ 0.18 \\
\small Backoff 1.5.2& \small 1.5.2, 24 & \small 1.5.2, 28 & \small 1.5.2, 15 & \small 1.5.2, 15 & \small 26.84 $\pm$ 0.05 * \\
\small Backoff 1.4.3 / 1.5.2& \small 1.4.3, 10 & \small 1.4.3, 16 & \small 1.5.2, 15 & \small 1.5.2, 15 & \small 26.96 $\pm$ 0.09 \\
\bottomrule
\end{tabular}
\end{center}
\label{tab:Transformer_WMT14_8GPUs_WeightsActivationsGradientsAccuracy}%
\end{table*}

To compare the performance of our quantized Transformer base model to results presented in the literature \citep{Vaswani17,Kuchaiev18,Mellempudi19,Prato19,Sun19} we trained the float-32 baseline model and selected fully quantized models with a batch size of 1024 sentence pairs for 100,000 iterations. To account for the larger batch size, we reduced the number of warm-up steps to 4,000 for these experiments, in line with \citet{Vaswani17}. Furthermore, we obtained the final weights from averaging the last five checkpoints which were saved at intervals of 1,000 iterations \citep{Vaswani17}.

The results reported in Table~\ref{tab:Transformer_WMT14_8GPUs_WeightsActivationsGradientsAccuracy} show that the BLEU scores achieved by the quantized models with the 1.5.2 format for gradients and the 1.4.3 or 1.5.2 format for activations and weights still match those of the float-32 baseline model and are similar to previously published results. This confirms our results of Table~\ref{tab:Transformer_WMT14_WeightsActivationsGradientsAccuracy} and Table~\ref{tab:Transformer_WMT14_AutomaticLossScale_WeightsActivationsGradientsAccuracy} on the applicability of 8-bit floating point formats for training the Transformer model.

\section{Conclusions}\label{sec:Conclusions}
We have reported a range of experiments that confirms the robust training and generalization performance of mixed-precision implementations based on 8-bit floating-point formats 1.4.3, with 4 bits of exponent and 3 bits of significand, for activations and weights, and 1.5.2, with 5 bits of exponent and 2 bits of significand, for gradients with respect to activations and weights.

The study provides evidence for the trade-off between the required number of bits used to represent the exponent and the significand for float-8 training. In particular, the results confirm the superior performance of 8-bit floating-point formats compared to 8-bit scaled integers, and indicate that using 3 bits of significand can be advantageous for quantization of the activations, while 5 bits of exponent is often necessary for quantization of the gradients with respect to activations.
We demonstrate that, with the choice of an exponent bias that allows for representing the majority of occurring values, models which use these 8-bit number formats in matrix multiplications and convolutions show no degradation of test performance compared to the corresponding float-32 baseline, with the most robust choice being to use the float-8 format 1.4.3 for activations and weights, and the float-8 format 1.5.2 for gradients with respect to activations and weights. These findings confirm that 8-bit floating-point formats are a useful alternative to higher precision number formats to accelerate the training of deep learning models and make efficient use of the available power.

\section{Broader impact}
The ongoing trend to use larger deep learning model architectures not only has consequences on the time and cost to train a model, but it also has a significant environmental impact \citep{Strubell2019}. It is critical for the research community to study the implementation of more efficient training schemes to counter the increase in energy consumption with growing model size.

The results of this study confirm that low precision numerical formats can be a key component of large machine learning models that provide state of the art accuracy while reducing their environmental impact. In particular, by using 8-bit floating point arithmetic the energy efficiency can be increased by up to $4\times$ with respect to float-16 arithmetic and up to $16\times$ with respect to float-32 arithmetic.

\bibliographystyle{plainnat}
\bibliography{numerical_format,additional_literature}

\clearpage
\appendix
\section{Dynamic range and signal-to-noise ratio of fixed-point quantization}\label{sec:AppendixSNR_FixedPoint}
For a scaled fixed-point representation with 1 sign bit, $n$ integer bits and quantization step $q$, the dynamic range is
\begin{equation} \label{eq:1}
D = 20\log_{10}[\, 2^{n} \,] \approx 6.02 \, n \; \mathrm{dB}.
\end{equation}

The roundoff error of a uniform quantizer with quantization step $q$ can be modelled by additive noise with uniform distribution over the interval ( ${-q/2}$, ${q/2}$ ). If the input signal $x$ satisfies Widrow's quantization theorems (i.e., if the input signal characteristic function $\Phi_x(u) \approx 0$ for $|u| > 2\pi/q$) \citep{Widrow96}, the statistical model of the quantization error $\nu$ of uniform quantization corresponds to an independent noise source with zero mean and variance $\mathbb{E}\{ \nu^{2} \}=q^2/12$ \citep{Widrow96,Widrow08}.
When quantizing a standard normal distributed random variable using a fixed-point quantizer with $n$ bits of precision and a quantization interval $q$, $\mathbb{E}\{\nu^2\}$ can be broken down into a {\em clipping noise} part that occurs when values that exceed the dynamic range of the quantizer get clipped and the {\em rounding noise} part that occurs within the dynamic range:
\begin{equation}
    \begin{split}
        \mathbb{E}\{\nu^2\} = &\int_{-\infty}^{\infty} \nu^2(x) \, f_x(x) \dx \\
        = &\underbrace{\int_{-(m+\frac{1}{2})q}^{(m+\frac{1}{2})q} \nu^2(x) \, f_x(x) \dx}_{\text{rounding noise}}
        \: + \: \underbrace{2\int_{-\infty\vphantom{\frac{1}{2}}}^{-(m+\frac{1}{2})q} \nu^2(x) \, f_x(x) \dx}_{\text{clipping noise}}
    \end{split}
\end{equation}
where $f_x(x)$ is the standard normal PDF and $m=2^n-1$.

With the approximation ${f_x(x)\approx f_x(iq)}$ in the interval $[(i-1/2)q, (i+1/2)q]$, the rounding noise component can be written as
\begin{equation*}
    \begin{split}
        \int_{-(m+\frac{1}{2})q}^{(m+\frac{1}{2})q} \nu^2(x) \, f_x(x) \dx
        &= \sum_{i=-m}^m \int_{(i-\frac{1}{2})q}^{(i+\frac{1}{2})q} \nu^2(x) \, f_x(x) \dx \\
        &\approx \sum_{i=-m}^m f_x(iq) \int_{-\frac{1}{2}q}^{\frac{1}{2}q} x^2 \dx \\
        &\approx \int_{-m}^{m} f_x(xq) \dx \cdot \frac{q^3}{12} \\
        &= \frac{q^2}{12} \;\mathrm{erf}\!\left(\frac{mq}{\sqrt{2}}\right) \,.
    \end{split}
\end{equation*}
This expression goes to $q^2/12$ for sufficiently large $mq$, i.e. if almost all values fall within the dynamic range.

For the clipping noise component we have
\vspace{2mm}
\begin{equation*}
    \begin{split}
        \; 2\int_{-\infty}^{-(m+\frac{1}{2})q} \nu^2(x) \, f_x(x) \dx 
        = & \; 2\int_{-\infty}^{-(m+\frac{1}{2})q} (x+mq)^2 f_x(x) \dx \\
        = & \; 2\int_{-\infty}^{-(m+\frac{1}{2})q} x^2 f_x(x) \dx
            + 4mq\int_{-\infty}^{-(m+\frac{1}{2})q} x \, f_x(x) \dx \\
            &+ 2m^2q^2\int_{-\infty}^{-(m+\frac{1}{2})q} f_x(x) \dx \\
        \approx & \; (2+2m^2q^2) \, F_x(-mq) - 2mq\,f_x(mq)  \\
        = & \; (1+m^2q^2)\left(1-\mathrm{erf}\! \left(\frac{mq}{\sqrt{2}}\right)\right) 
        -  \sqrt{\frac{2}{\pi}} \, mq \; e^{-\frac{m^2q^2}{2}}
    \end{split}
\end{equation*}
\vspace{2mm}

where $F_x(x)$ denotes the standard normal CDF. This gives rise to an overall signal-to-noise ratio of
\vspace{5mm}
\begin{equation}
    \begin{split}
        \mathrm{SNR} = &-10\log_{10}[\, \mathbb{E}\{ \nu^{2} \} \,] \\
        \approx &-10\log_{10}\!\biggl[\;\frac{q^2}{12} \;\mathrm{erf}\!\left(\frac{mq}{\sqrt{2}}\right)
            + \left(1+m^2q^2\right)\left(1-\mathrm{erf}\left(\frac{mq}{\sqrt{2}}\right)\right)
            - \sqrt{\frac{2}{\pi}} \,mq \; e^{-\frac{m^2q^2}{2}} \;\biggr] \, .
    \end{split}
    \label{eq:SNR}
\end{equation}
\vspace{2mm}

\begin{figure}[htb]
\centering
\subfloat[]{\includegraphics[width=0.36\columnwidth]{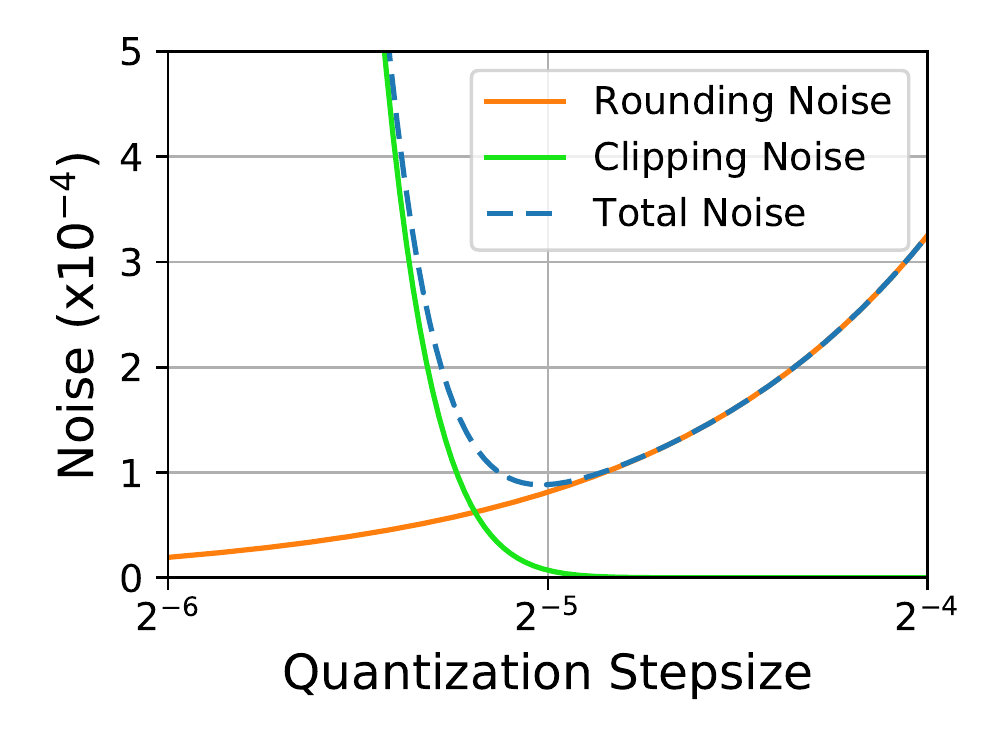}}
\quad
\subfloat[]{\includegraphics[width=0.36\columnwidth]{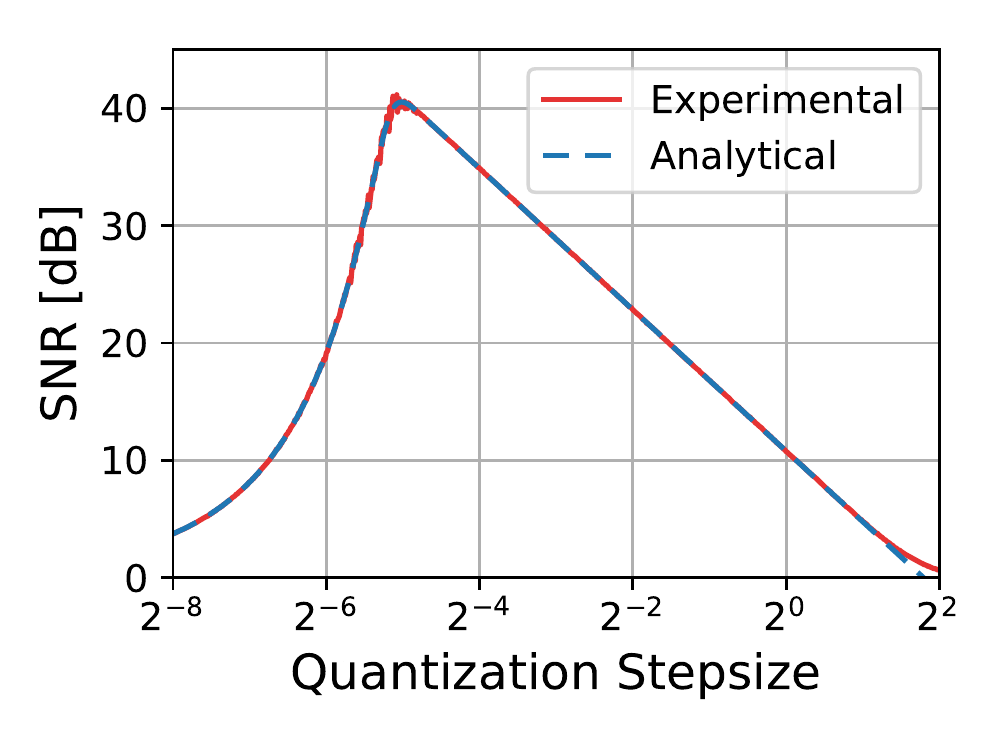}}%
\caption{(a) Breakdown of the noise components for 8-bit fixed-point quantization of a standard normal signal. (b) SNR for 8-bit fixed-point quantization of a standard normal signal, derived from (\ref{eq:SNR}) (blue) and experimentally obtained from a Gaussian distribution (red).}
\label{fig:FixedPointSNR}%
\end{figure}

\FloatBarrier\clearpage
\section{Statistical model of floating-point quantization} \label{sec:AppendixFloatingPointQuantizationModel}
A floating-point representation with $p$ bits of significand\footnote{Here $p$ denotes the precision of a floating point number with sub-normal values including the implicitly stored hidden leading bit.} and exponent range $E_{max}-E_{min}$, has a dynamic range
\begin{equation} \label{eq:3}
    \begin{split}
        D &= (2^p-1) \, 2^{E_{max}-E_{min}} \\
        &\approx 6.02 \, (E_{max}-E_{min}+p) \; \mathrm{dB}.
    \end{split}
\end{equation}

\begin{figure}[h]
\centering
\includegraphics[width=0.56\columnwidth]{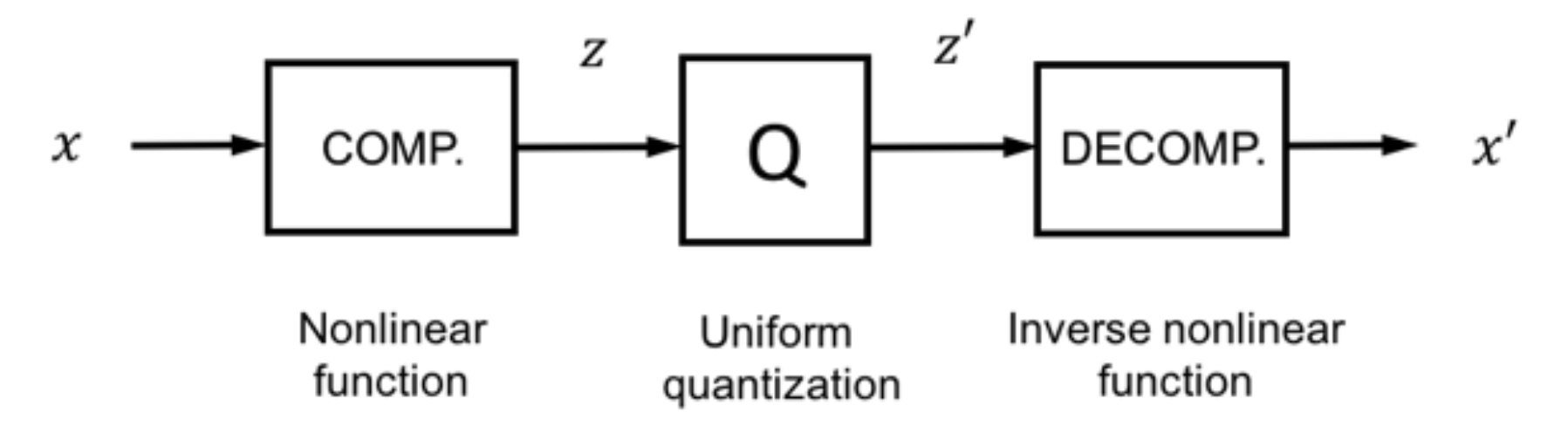}
\caption{Model of floating-point quantization.}
\label{fig:FloatingPointQuantizationModel}
\end{figure}

The quantization error $\nu_{FL}$ associated with a floating-point representation is approximately proportional to the amplitude of the input signal~\citep{Widrow08}.
The input-output characteristic of a floating-point quantizer with $p$ bits of significand and quantization step $q$ is only piecewise uniform over intervals of width $\Delta = 2^p q$.
The results of uniform quantization derived in Appendix~\ref{sec:AppendixSNR_FixedPoint} are therefore generally not applicable to floating-point quantization. However, a statistical model of the floating-point quantization error can be obtained by representing the floating-point quantizer by a cascade of a piecewise linear compressors, followed by a uniform quantizer and a piecewise linear expansion, as illustrated in Figure~\ref{fig:FloatingPointQuantizationModel}~\citep{Widrow96,Widrow08}.
\vspace{5mm}

\vspace{5mm}

The floating-point quantization error $\nu_{FL}=x'-x$ can then be expressed in terms of the uniform quantization error $\nu=z'-z$, that in practice can be represented by the classical uniform quantization model \citep{Widrow08}.
The decompression function of Figure~\ref{fig:FloatingPointQuantizationModel} is approximately an exponential function, and its output contains an additional quantization noise term $\nu_{EXP}$ corresponding to the quantized exponent~\citep{Widrow08}
\[ 
\nu_{FL}=\sqrt{2} \; \nu \cdot \frac{|x|}{\Delta} \, 2^{\, \nu_{EXP}} \, .
\]
If the uniform quantization model applies to both the inner quantizer and the quantized exponent noise $\nu_{EXP}$, the variance of the floating-point quantization noise $\nu_{FL}$ is given by \citep{Widrow08}
\[
\begin{split}
    &\mathbb{E}\{ {\nu_{FL}}^2 \} = 2.16 \cdot \mathbb{E}\{ \nu^2 \} \, \mathbb{E}\{ \frac{x^2}{\Delta^2} \} \\
    = \; &2.16 \cdot \frac{\, q^2}{12} \, \frac{ \mathbb{E}\{ x^2 \} }{\Delta^2} 
    = 0.18 \cdot 2^{-2p} \, \mathbb{E}\{ x^2 \}
\end{split}
\]
which corresponds to
\begin{equation}
\mathrm{SNR} = 5.55 \cdot 2^{2p} \; \Rightarrow \; \mathrm{SNR}_{\mathrm{dB}} = 7.44 + 6.02 p \, .
\end{equation}

\FloatBarrier\clearpage
\section{Results for other float-8 formats} \label{sec:AppendixOtherNumberFormats}
\begin{figure}[!htb]
\centering
\subfloat[]{\includegraphics[width=0.32\columnwidth]{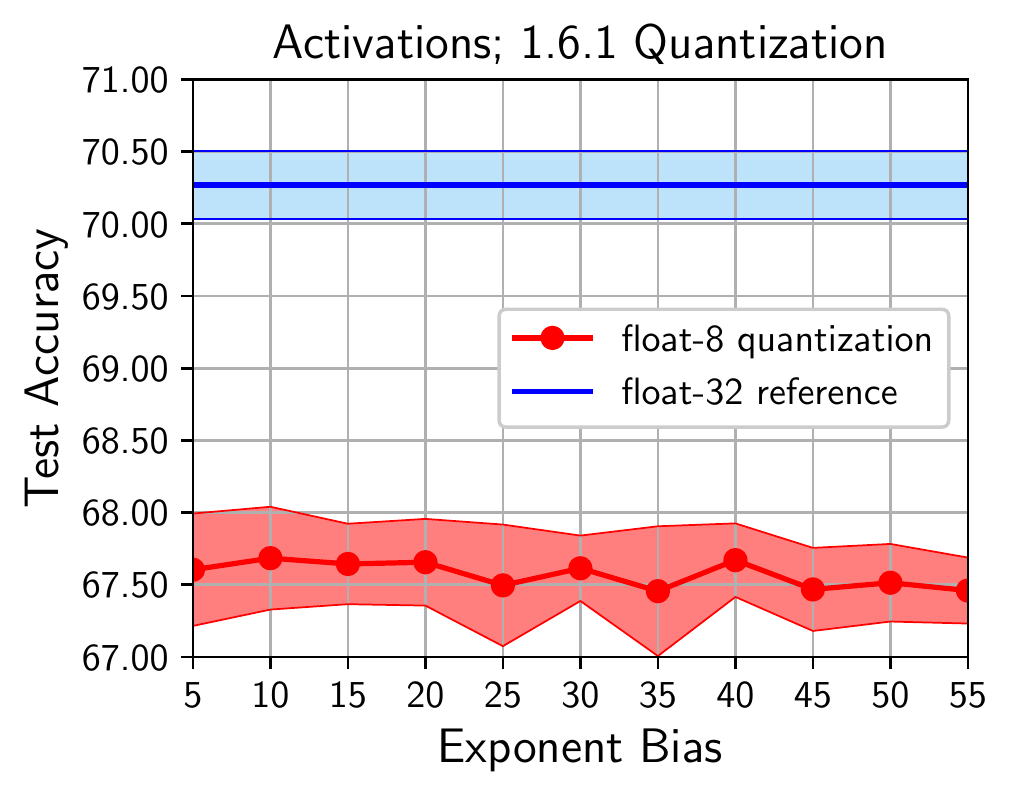}}%
\subfloat[]{\includegraphics[width=0.32\columnwidth]{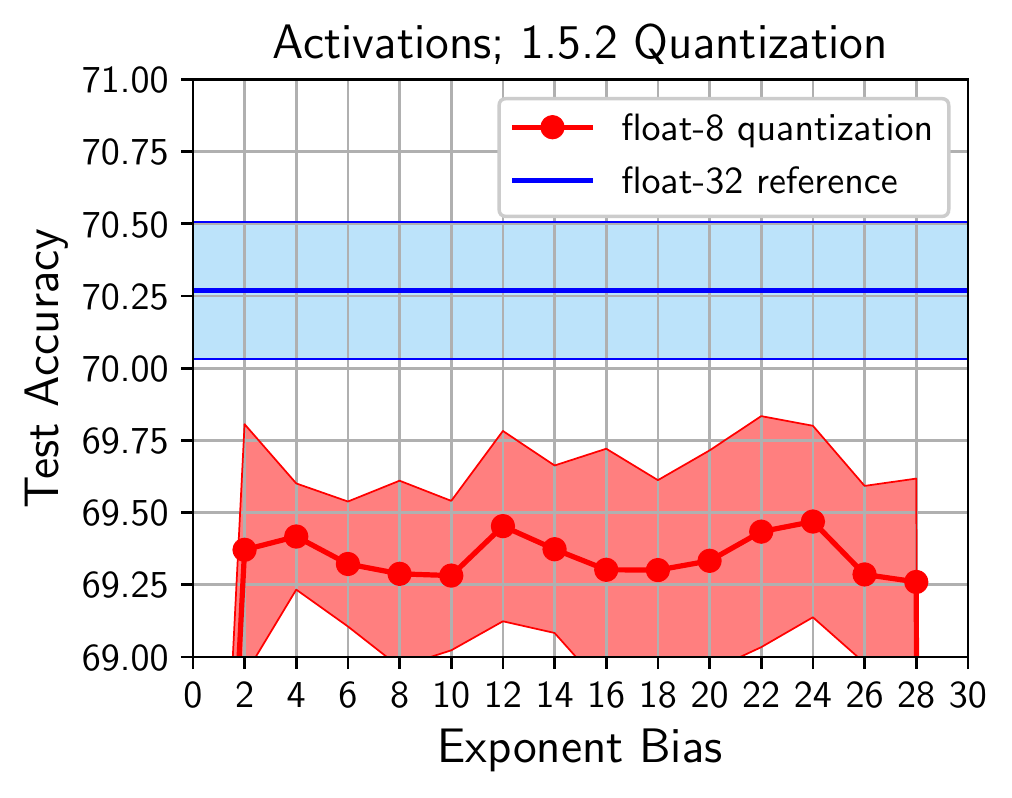}}
\subfloat[]{\includegraphics[width=0.32\columnwidth]{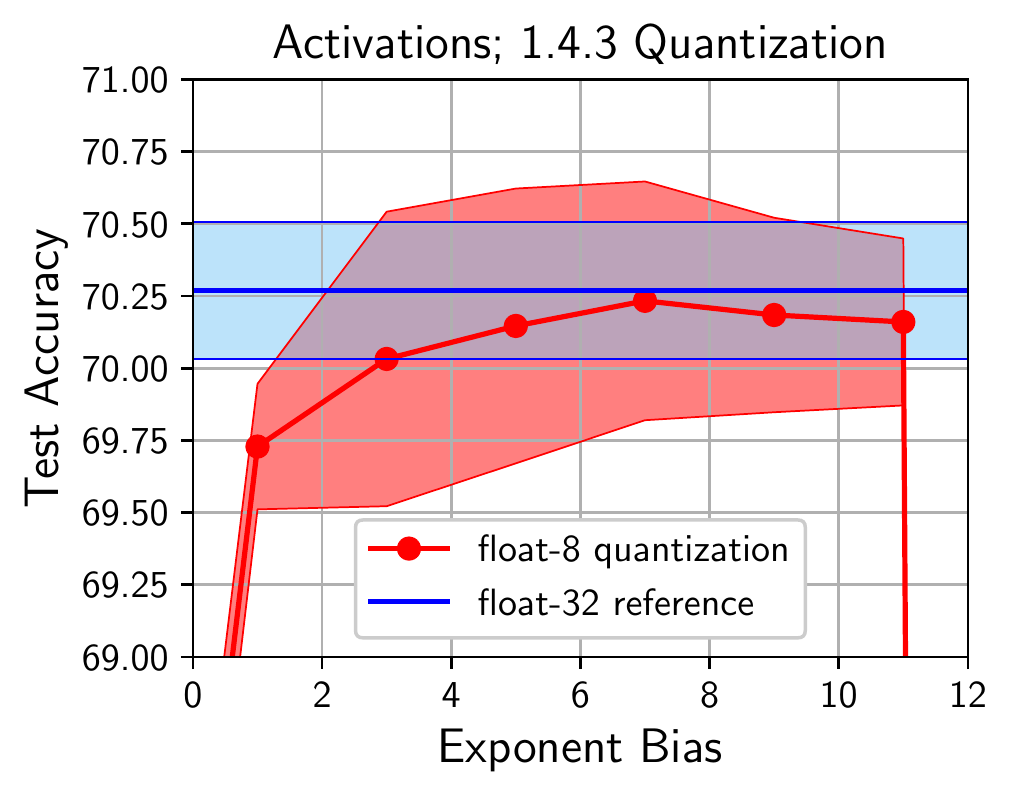}}\\
\subfloat[]{\includegraphics[width=0.32\columnwidth]{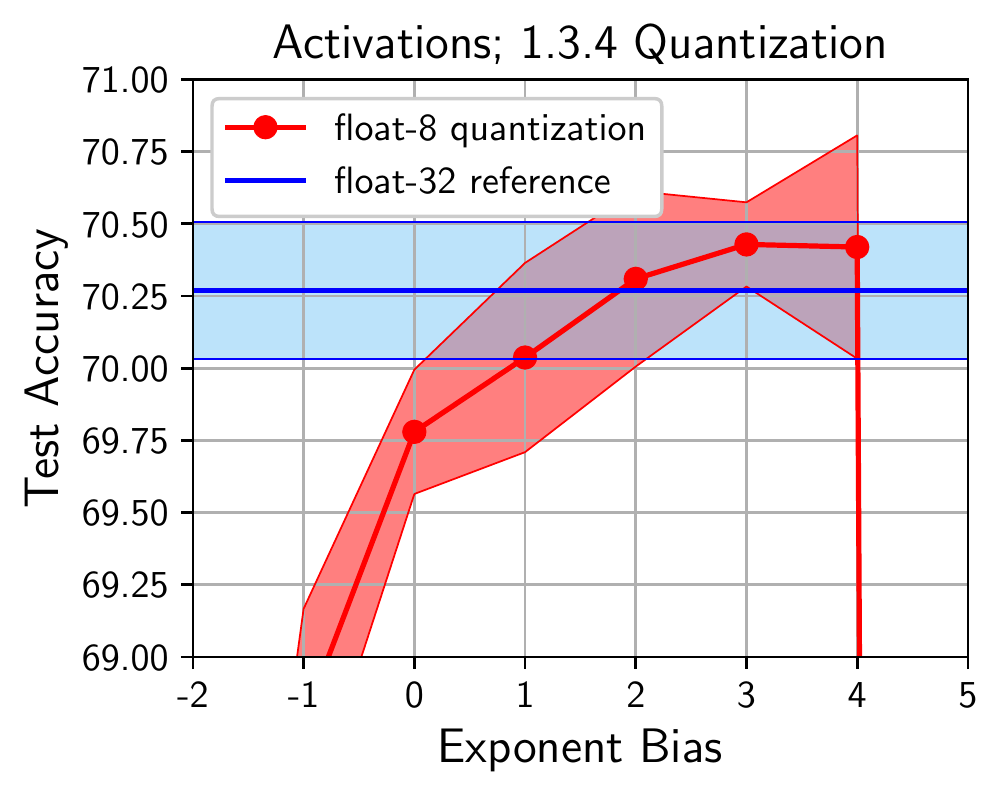}}
\subfloat[]{\includegraphics[width=0.32\columnwidth]{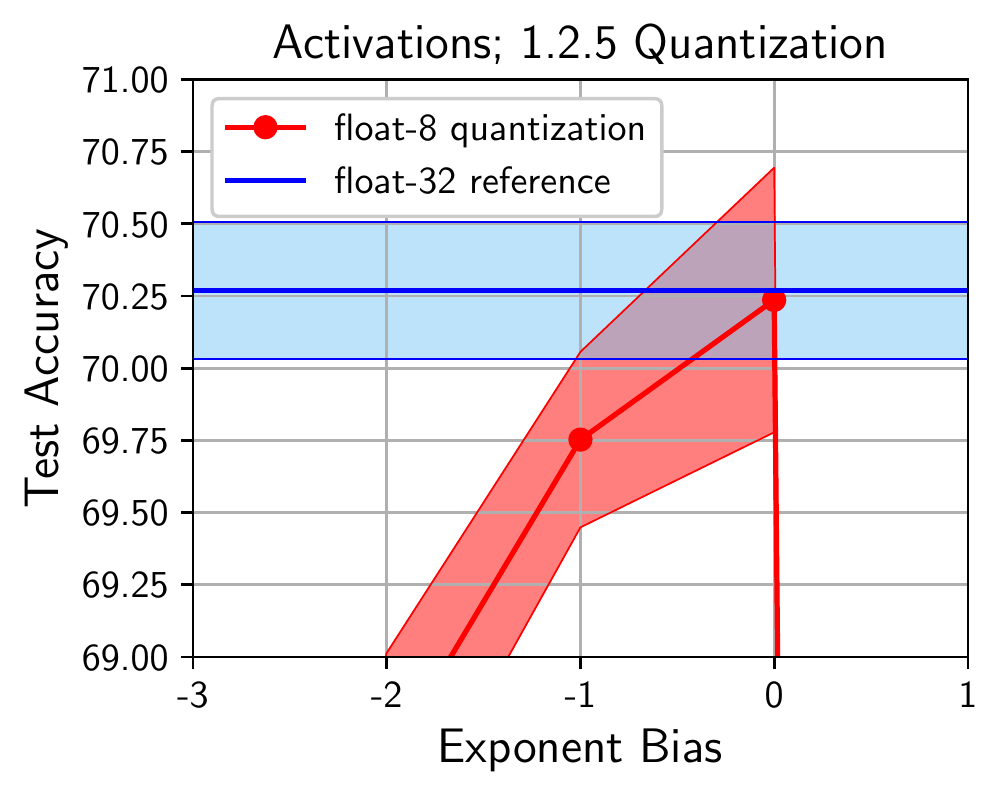}}%
\subfloat[]{\includegraphics[width=0.32\columnwidth]{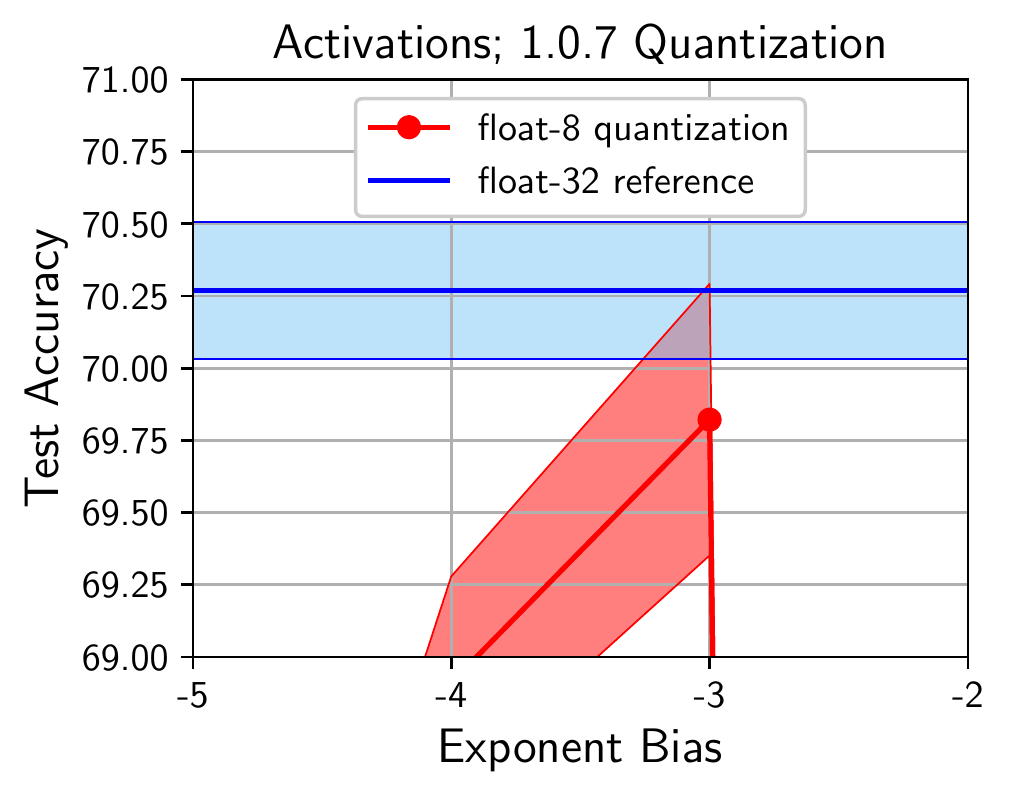}}%
\caption{ResNet-32 CIFAR-100 test performance of different 8-bit floating-point formats for representation of the activations with quantization of the input to the first layer. Test accuracy mean $\pm$ standard deviation over ten independent runs.}%
\label{fig:ResNet-32_CIFAR-100_ActivationsAccuracy_Input}%
\end{figure}

\begin{figure*}[!htb]
\centering
\subfloat[]{\includegraphics[width=0.32\columnwidth]{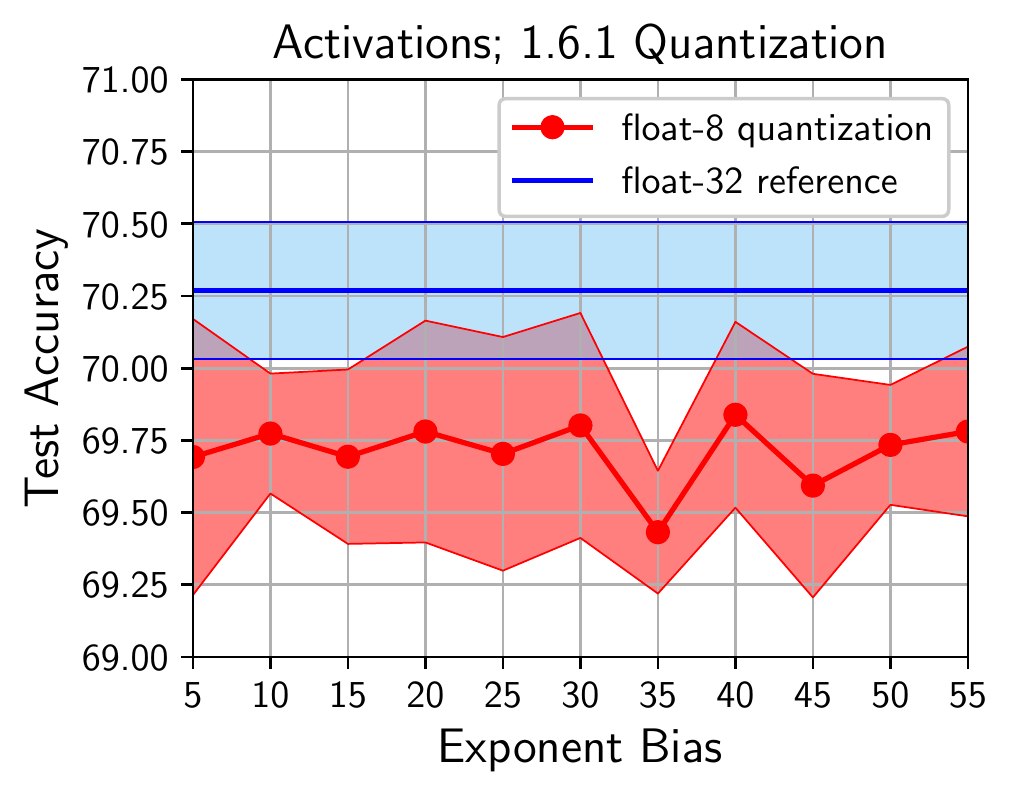}}%
\subfloat[]{\includegraphics[width=0.32\columnwidth]{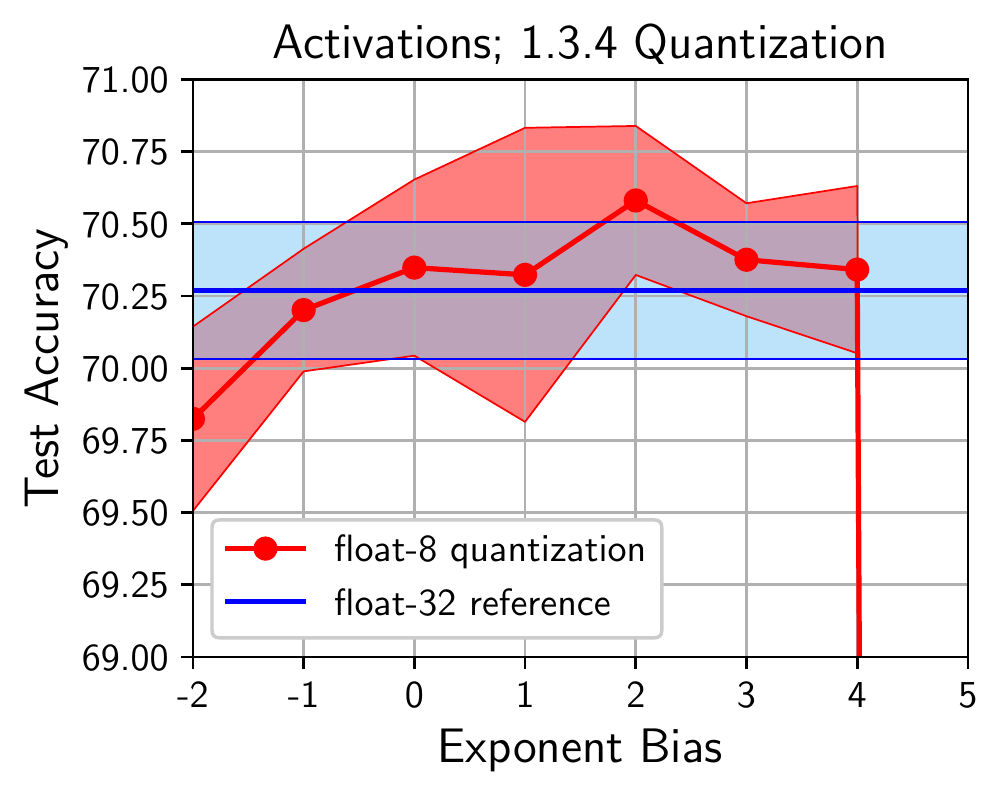}}
\subfloat[]{\includegraphics[width=0.32\columnwidth]{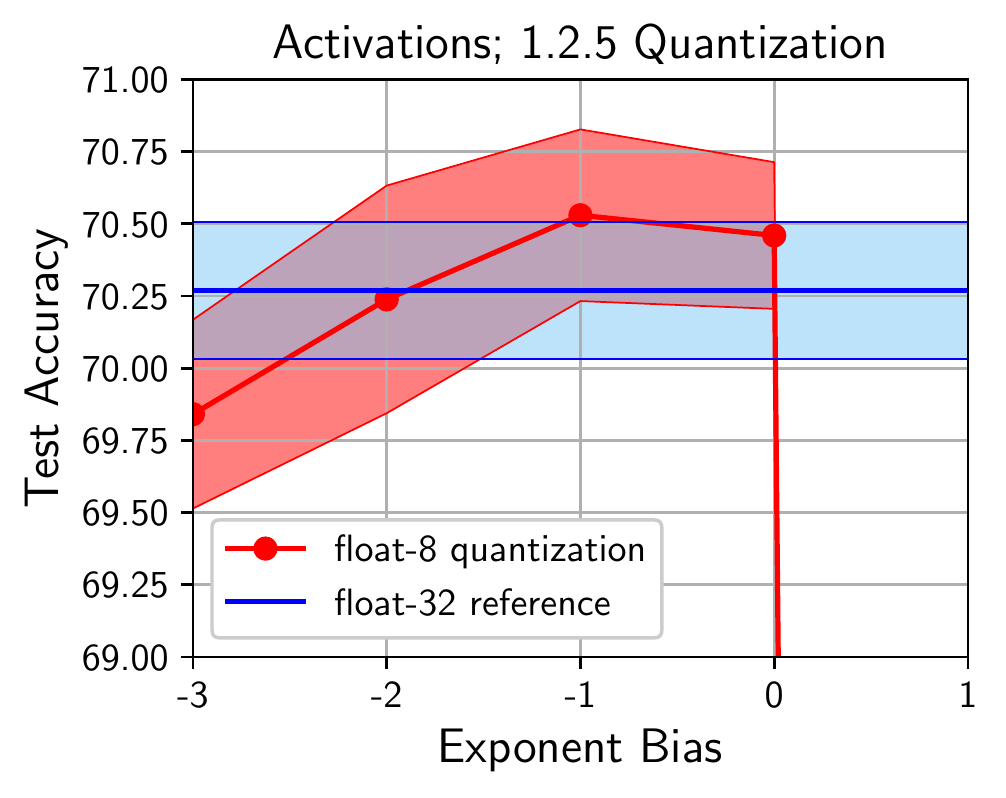}}
\caption{ResNet-32 CIFAR-100 test performance of different 8-bit floating-point formats for representation of the activations without quantization of the input to the first layer. Test accuracy mean $\pm$ standard deviation over ten independent runs.}%
\label{fig:ResNet-32_CIFAR-100_ActivationsAccuracy_NoInput_Other}%
\end{figure*}

\begin{figure*}[!htb]
\centering
\subfloat[]{\includegraphics[width=0.32\columnwidth]{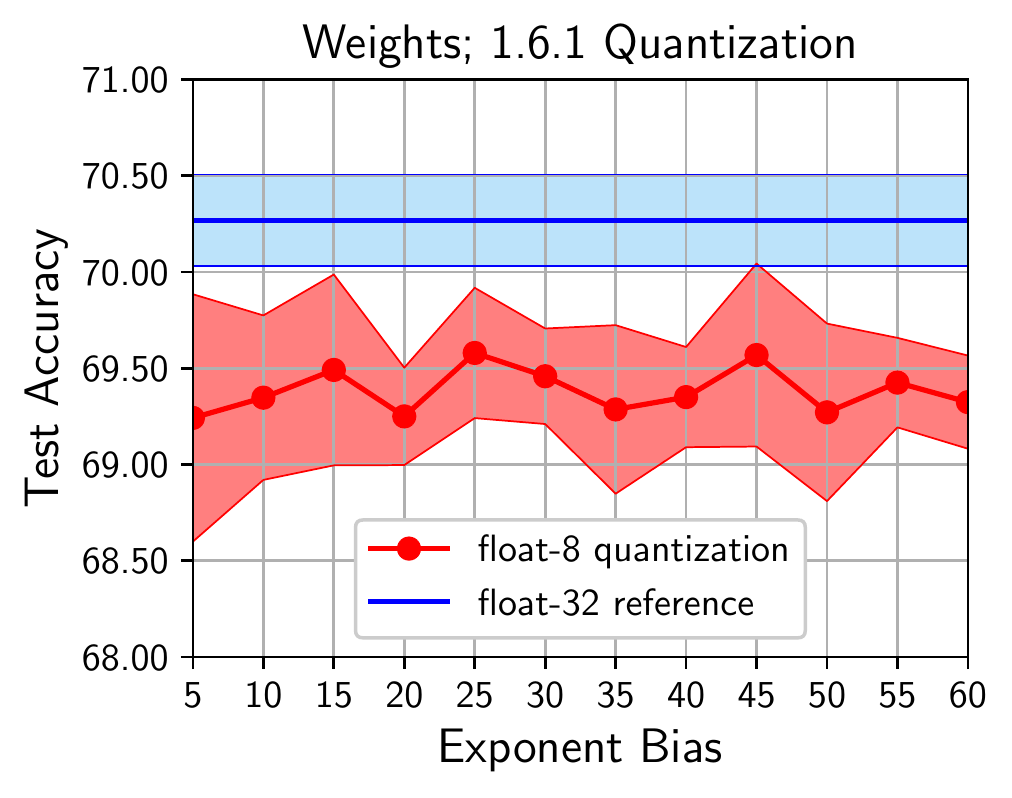}}%
\subfloat[]{\includegraphics[width=0.32\columnwidth]{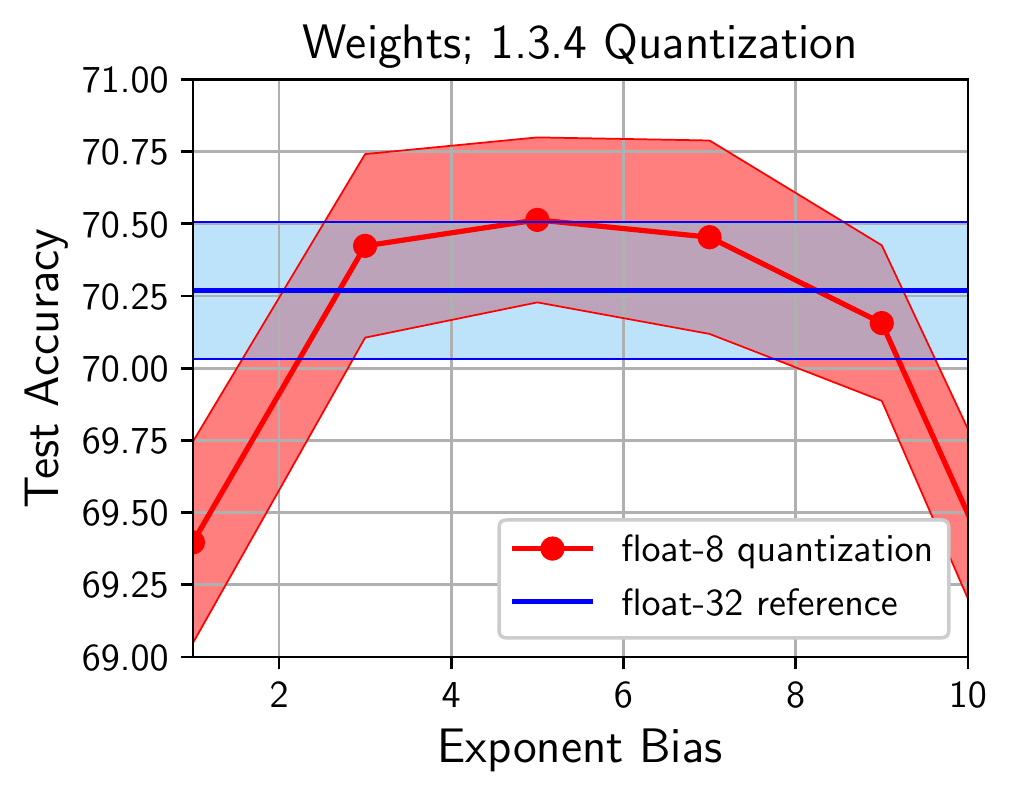}}
\subfloat[]{\includegraphics[width=0.32\columnwidth]{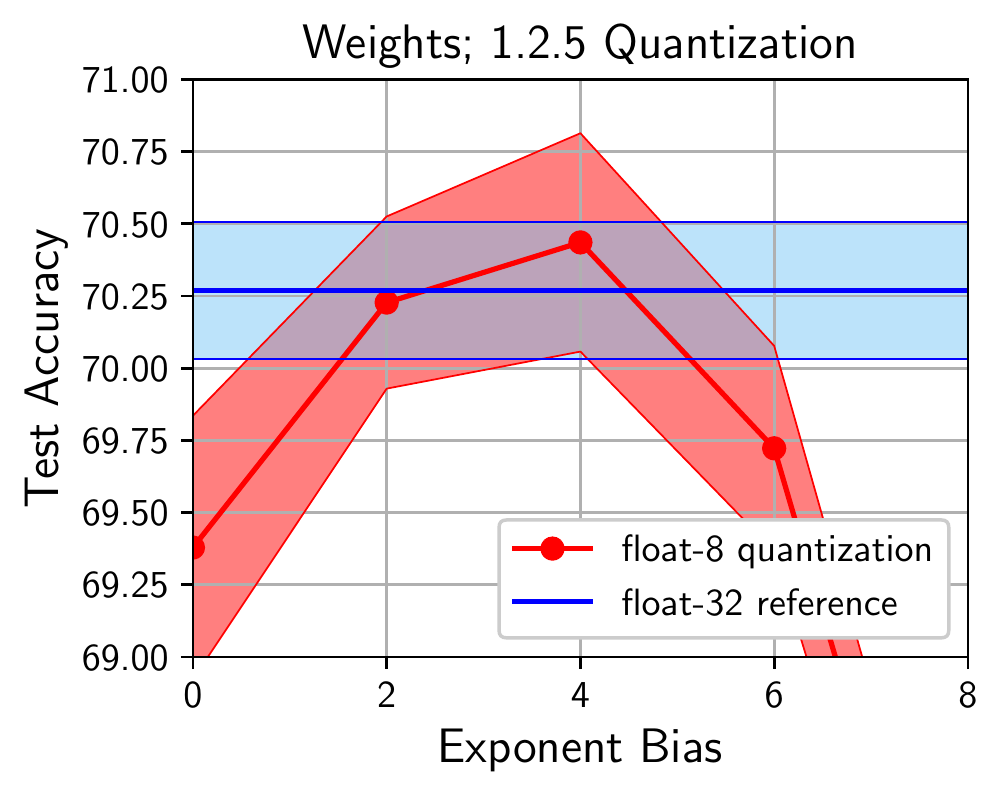}}
\caption{ResNet-32 CIFAR-100 test performance of different 8-bit floating-point formats for representation of the weights. Test accuracy mean $\pm$ standard deviation over ten independent runs.}%
\label{fig:ResNet-32_CIFAR-100_WeightsAccuracy_Other}%
\end{figure*}

\begin{figure*}[!htb]
\centering
\subfloat[]{\includegraphics[width=0.32\columnwidth]{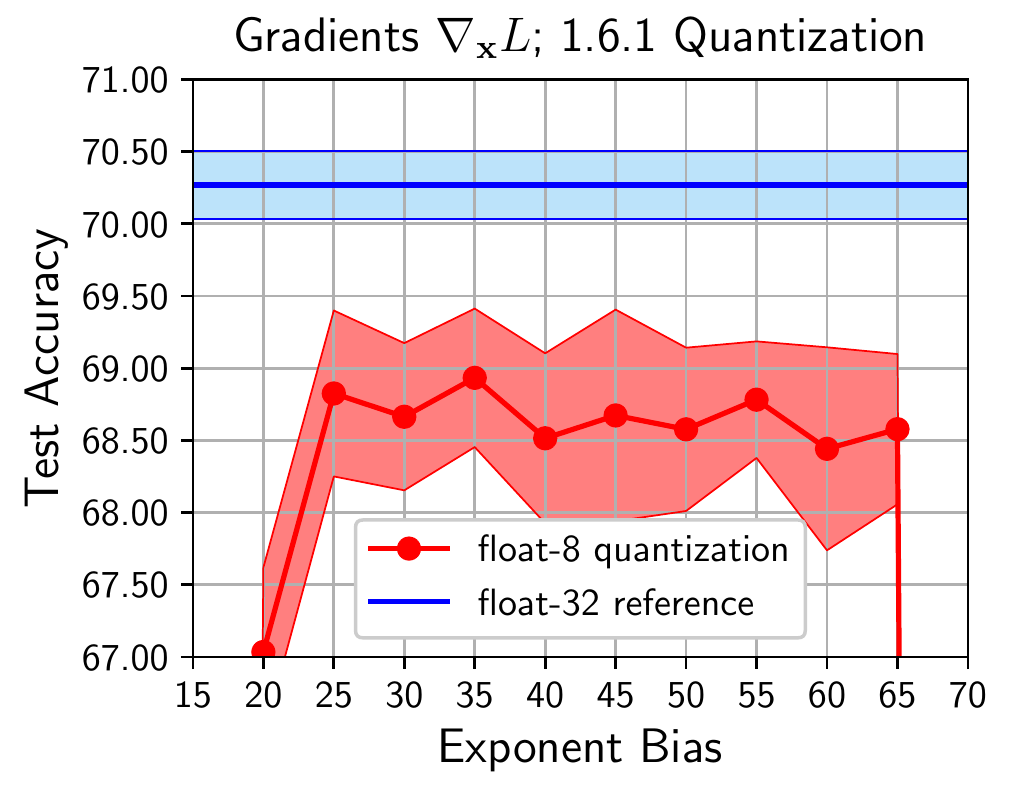}}%
\subfloat[]{\includegraphics[width=0.32\columnwidth]{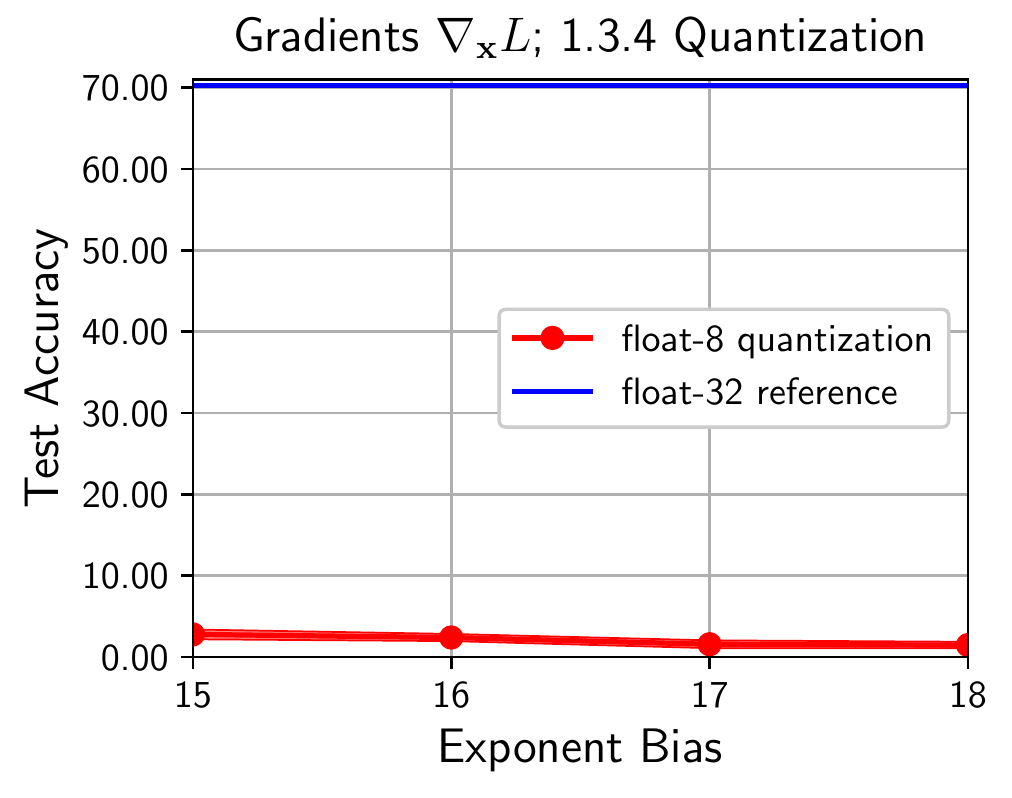}}
\subfloat[]{\includegraphics[width=0.32\columnwidth]{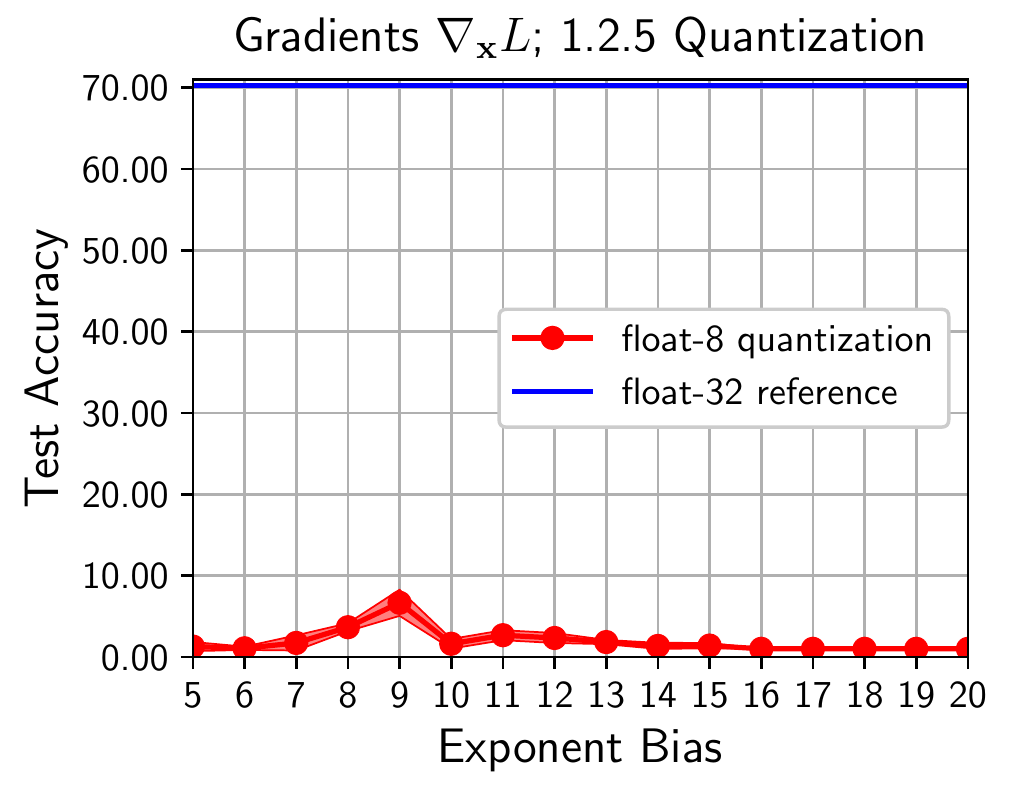}}
\caption{ResNet-32 CIFAR-100 test performance of different 8-bit floating-point formats for representation of the gradients with respect to activations. Test accuracy mean $\pm$ standard deviation over ten independent runs.}%
\label{fig:ResNet-32_CIFAR-100_GradientsActivationsAccuracy_Other}%
\end{figure*}

\begin{figure*}[!htb]
\centering
\subfloat[]{\includegraphics[width=0.32\columnwidth]{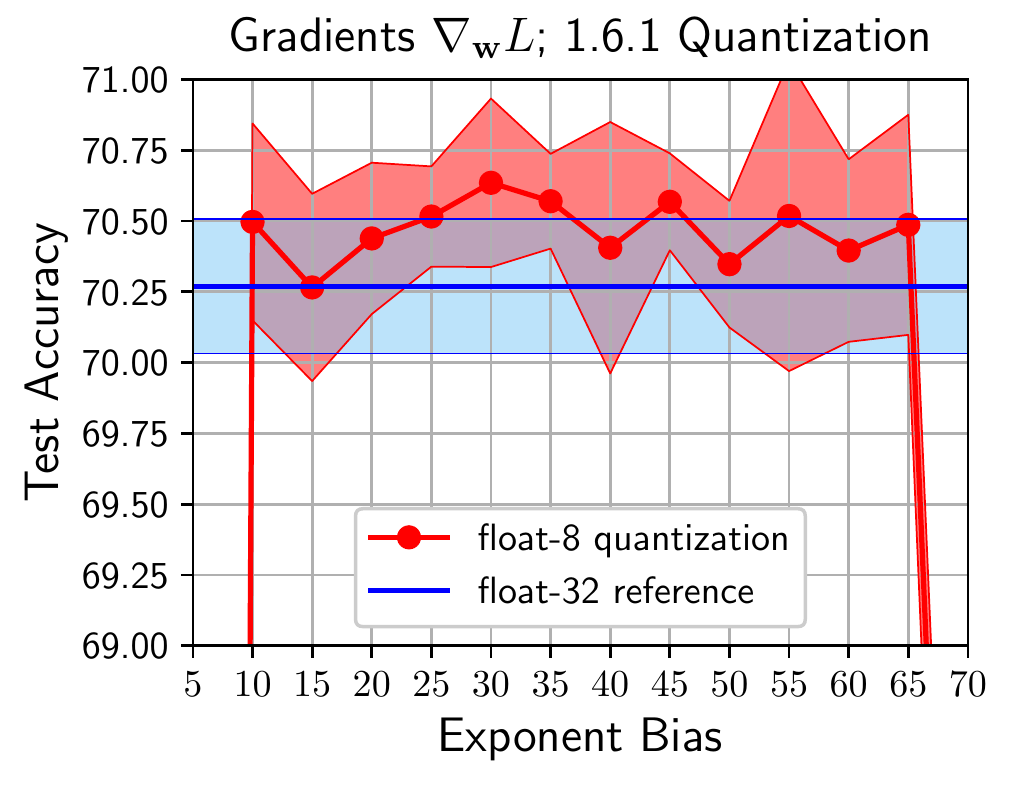}}%
\subfloat[]{\includegraphics[width=0.32\columnwidth]{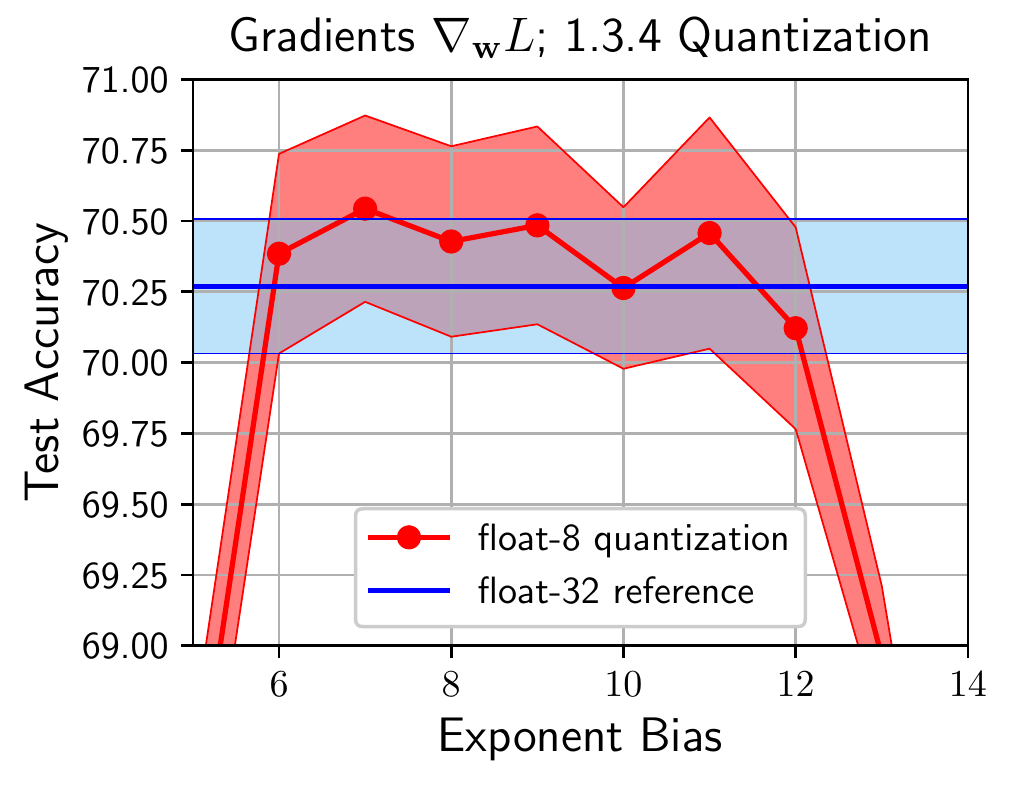}}
\subfloat[]{\includegraphics[width=0.32\columnwidth]{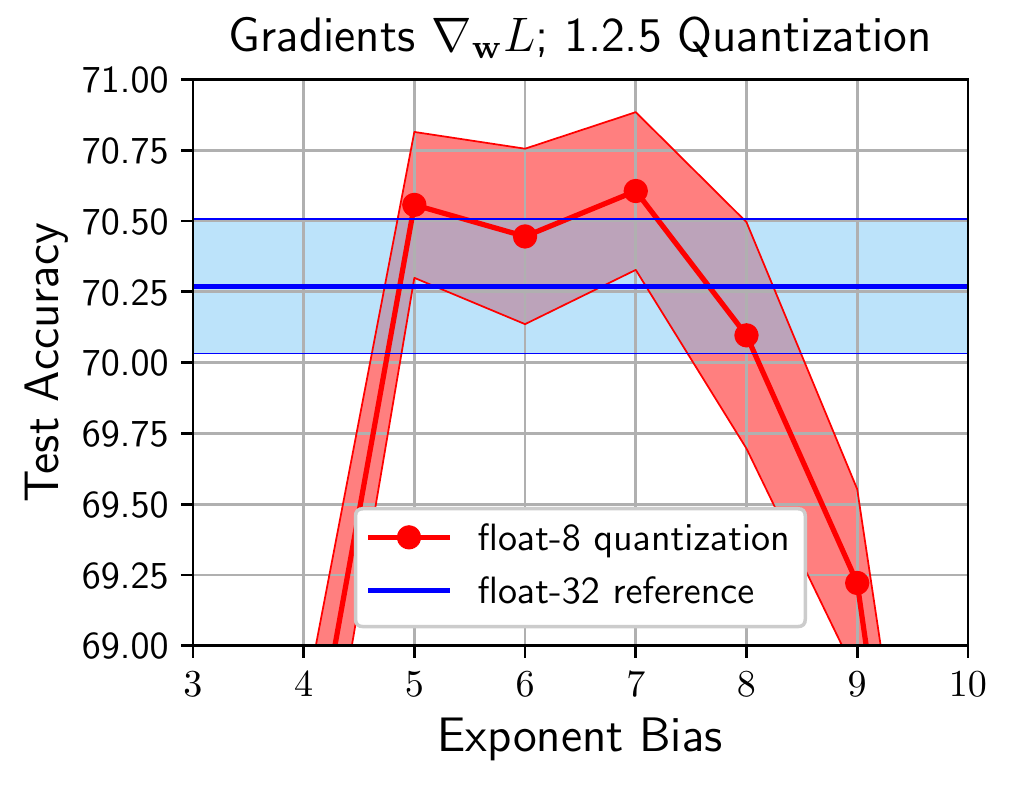}}
\caption{ResNet-32 CIFAR-100 test performance of different 8-bit floating-point formats for representation of the gradients with respect to weights. Test accuracy mean $\pm$ standard deviation over ten independent runs.}%
\label{fig:ResNet-32_CIFAR-100_GradientsWeightsAccuracy_Other}%
\end{figure*}

In this appendix we present results on the test accuracy of a ResNet-32 model trained on the CIFAR-100 dataset, when we additionally quantize the inputs to the first layer using an 8-bit floating point format (Figure~\ref{fig:ResNet-32_CIFAR-100_ActivationsAccuracy_Input}). 
In Figures~\ref{fig:ResNet-32_CIFAR-100_ActivationsAccuracy_NoInput_Other}--\ref{fig:ResNet-32_CIFAR-100_GradientsWeightsAccuracy_Other} we also report the CIFAR-100 performance for ResNet-32 training with separate quantization of activations, weights, and gradients with the 1.6.1, 1.3.4, and 1.2.5 number formats.
\vspace{10mm}

\vspace{10mm}

\FloatBarrier\clearpage
\section{Histograms}
\label{sec:AppendixHistograms}
A critical consideration in the selection of a reduced precision numerical format is the number {\em range} to be represented. To aid this selection, histograms of the appropriate quantities observed during a training session in which all quantities are represented with full precision have been generated. This section presents these histograms, together with a visual indication of how they relate to the limits of the candidate numerical formats.

\begin{figure}[!htb]
\centering
\includegraphics[width=0.9\textwidth]{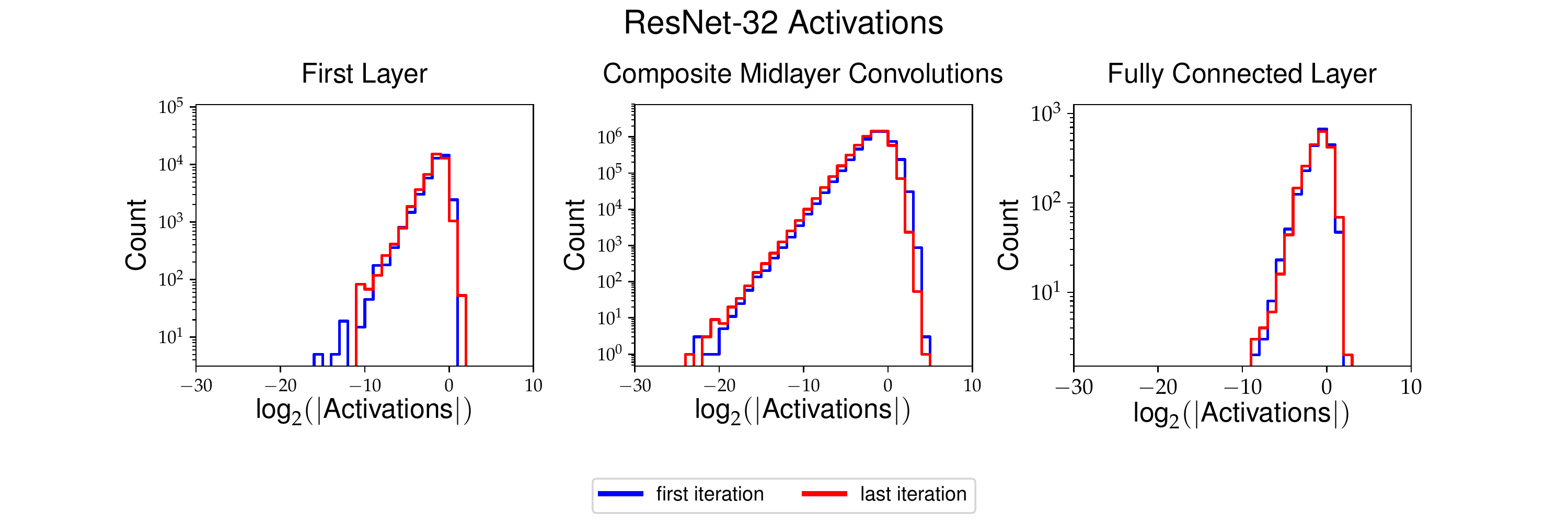}
\caption{Histograms of the activations of different ResNet-32 layers for CIFAR-100 training.}
\label{fig:ResNet-32_CIFAR-100_ActivationsHistogram}
\end{figure}

\begin{figure}[!htb]
\centering
\includegraphics[width=0.9\textwidth]{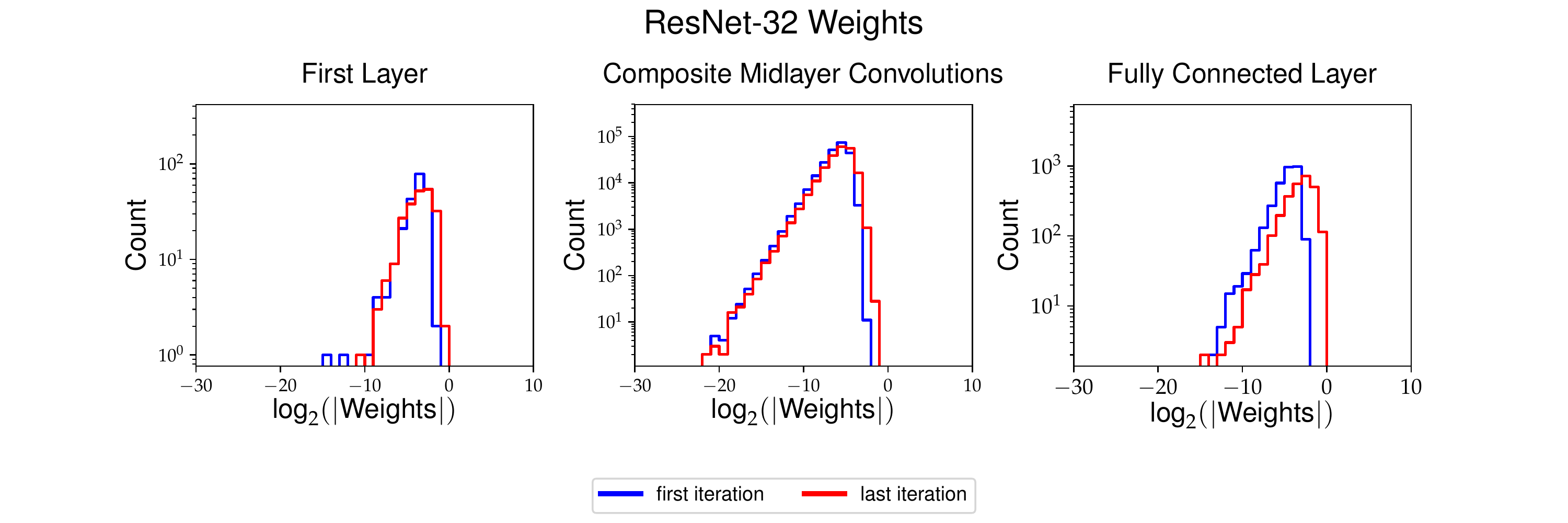}
\caption{Histograms of the weights of different ResNet-32 layers for CIFAR-100 training.}
\label{fig:ResNet-32_CIFAR-100_WeightsHistogram}
\end{figure}

\begin{figure}[!htb]
\centering
\includegraphics[width=0.9\textwidth]{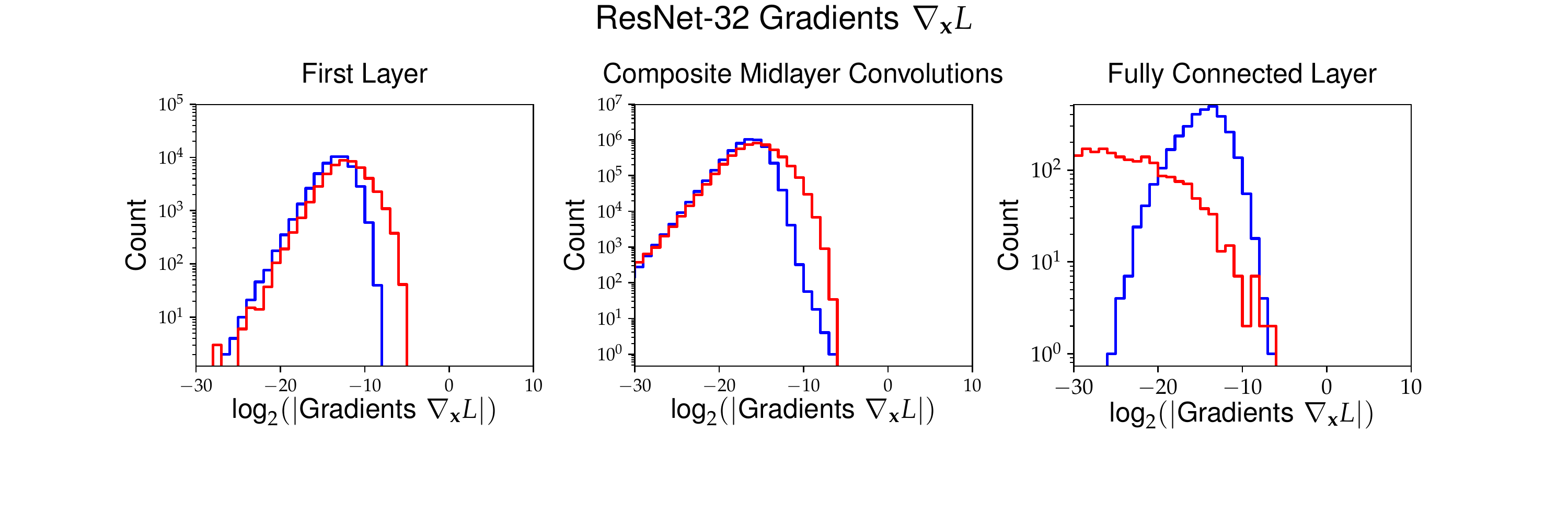} \\
\includegraphics[width=0.9\textwidth]{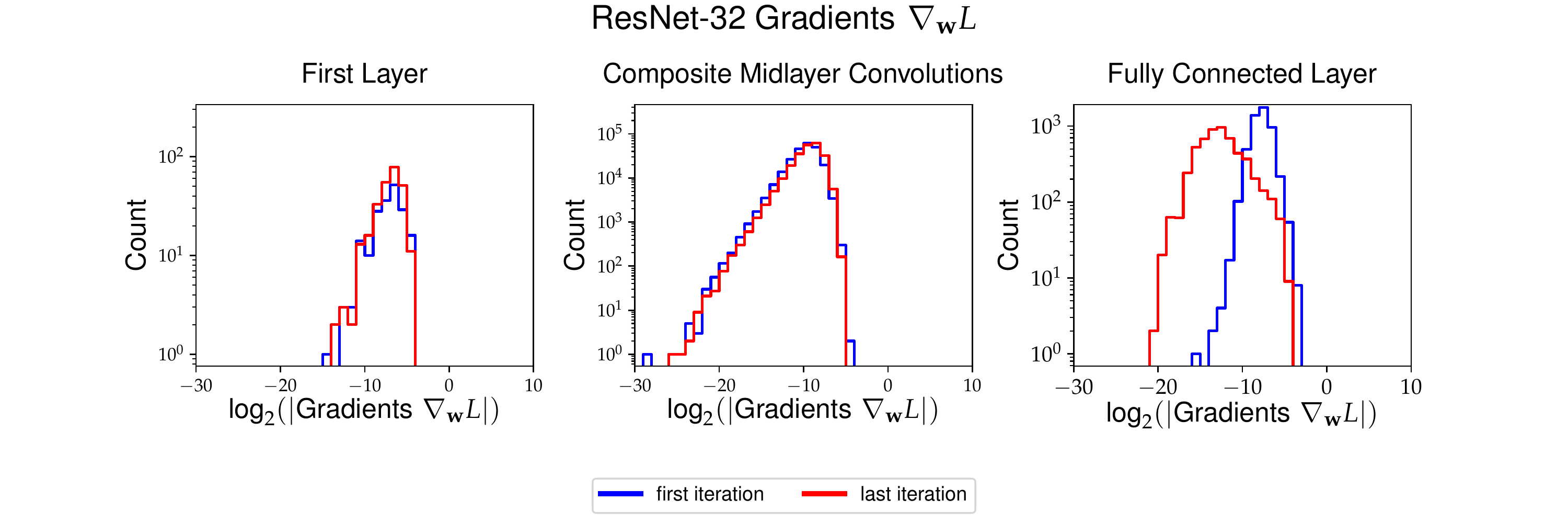}
\caption{Histograms of the gradients of different ResNet-32 layers for CIFAR-100 training.}
\label{fig:ResNet-32_CIFAR-100_GradientsHistogram}
\end{figure}

Figures~\ref{fig:ResNet-32_CIFAR-100_ActivationsHistogram}, \ref{fig:ResNet-32_CIFAR-100_WeightsHistogram} and \ref{fig:ResNet-32_CIFAR-100_GradientsHistogram} respectively show the histograms of the activations, weights and gradients observed at the beginning (blue curves) and end (red curves) of a 200 epoch training session using ResNet-32 on the CIFAR-100 dataset. The histograms display the counts of exponent values, computed as $\log_2$ of the absolute values of the tensor components. Separate histograms are given for the first layer, the last (fully connected) layer, and the composite counts from all other layers (all of which are very similar, hence the relevant information is included in the counts obtained by summing over these layers).

\begin{figure}[!htb]
\hspace{-0.9cm}
\includegraphics[width=1.1\textwidth]{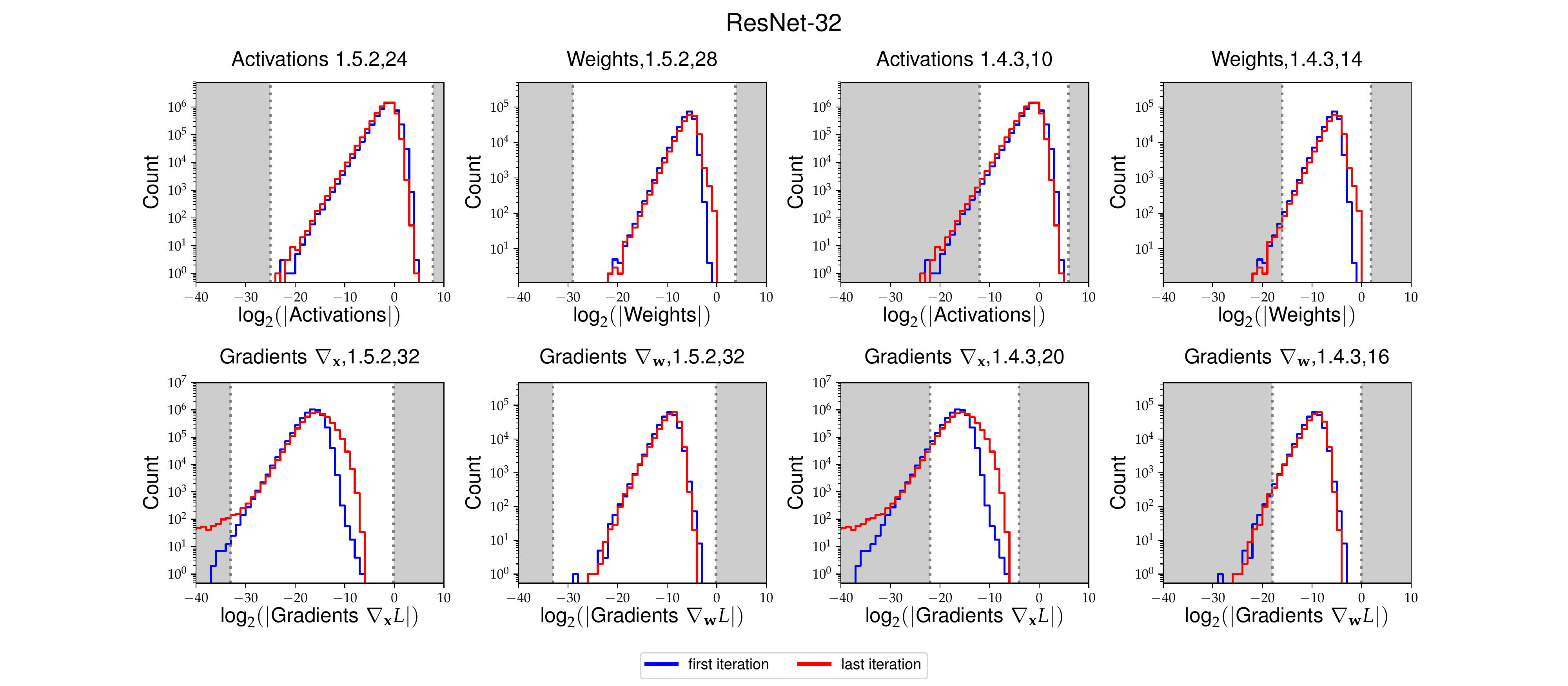}
\caption{Histograms of the weights, activations and gradients of all layers for ResNet-32 CIFAR-100 training. The white areas correspond to the range of values representable by the respective 8-bit floating-point number format.}
\label{fig:ResNet-32_CIFAR-100_AllLayer_Histogram}
\end{figure}

\begin{figure}[!htb]
\hspace{-0.9cm}
\includegraphics[width=1.1\textwidth]{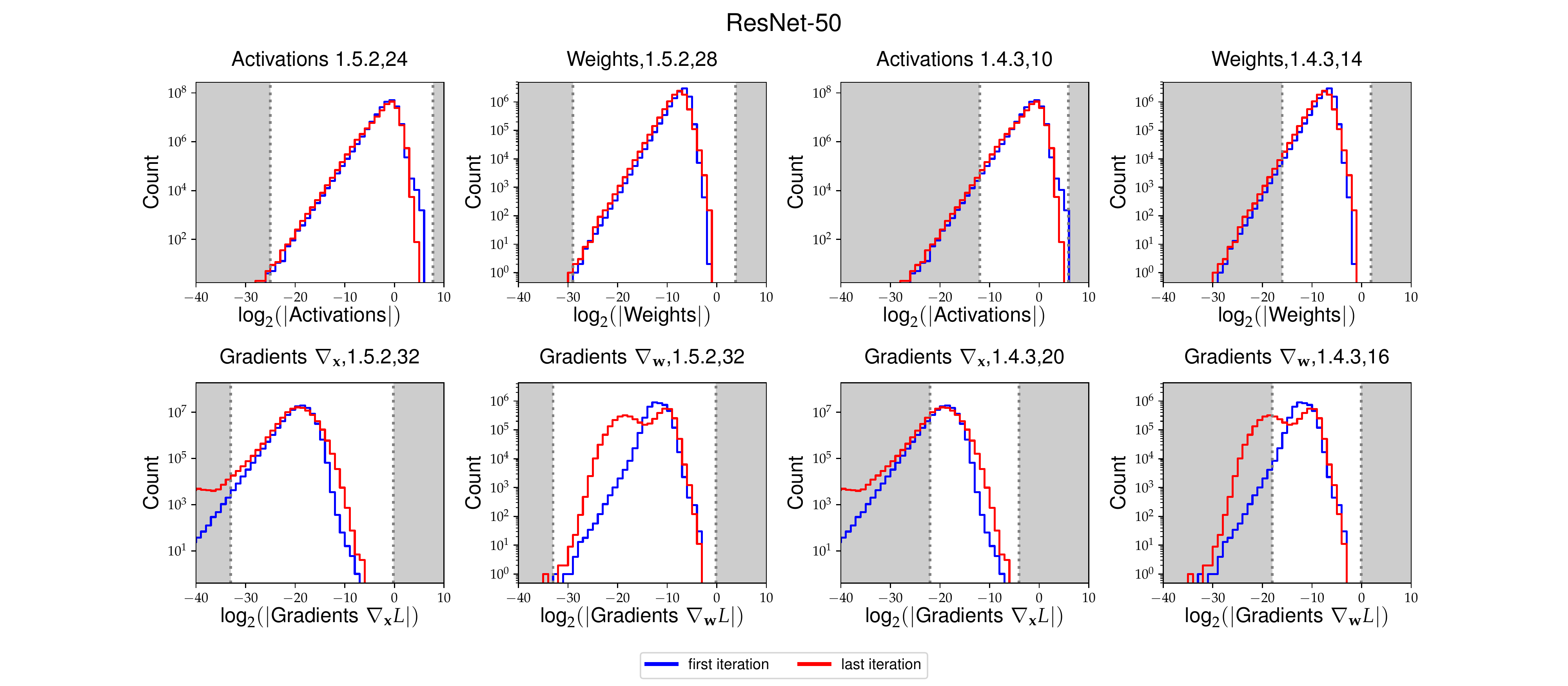}
\caption{Histograms of the weights, activations and gradients of different layers for ResNet-50 ImageNet training. The white areas correspond to the range of values representable by the respective 8-bit floating-point number format.}
\label{fig:ResNet-50_ImageNet_Histogram}
\end{figure}

Figure~\ref{fig:ResNet-32_CIFAR-100_AllLayer_Histogram} shows the relationship between the chosen reduced precision formats and the histograms for the ResNet-32/CIFAR-100 combination. The range covered by the formats (including denorm range) is shown in the white area on the plots. The histogram data in this figure gives composite counts of the exponent values for the tensor components to which the reduced precision format in the subplot title has been applied (so they do not include first layer activations and gradients with respect to activations, both of which are left unquantized). Figure~\ref{fig:ResNet-50_ImageNet_Histogram} shows the analogous data for ResNet-50 trained on the ImageNet dataset. Histograms of activations, weights and gradients of ResNet-18 trained on ImageNet are virtually identical to the ResNet-50 histograms of Figure~\ref{fig:ResNet-50_ImageNet_Histogram}, and have not been included.

\begin{figure}[!htb]
\hspace{-0.9cm}
\includegraphics[width=1.1\textwidth]{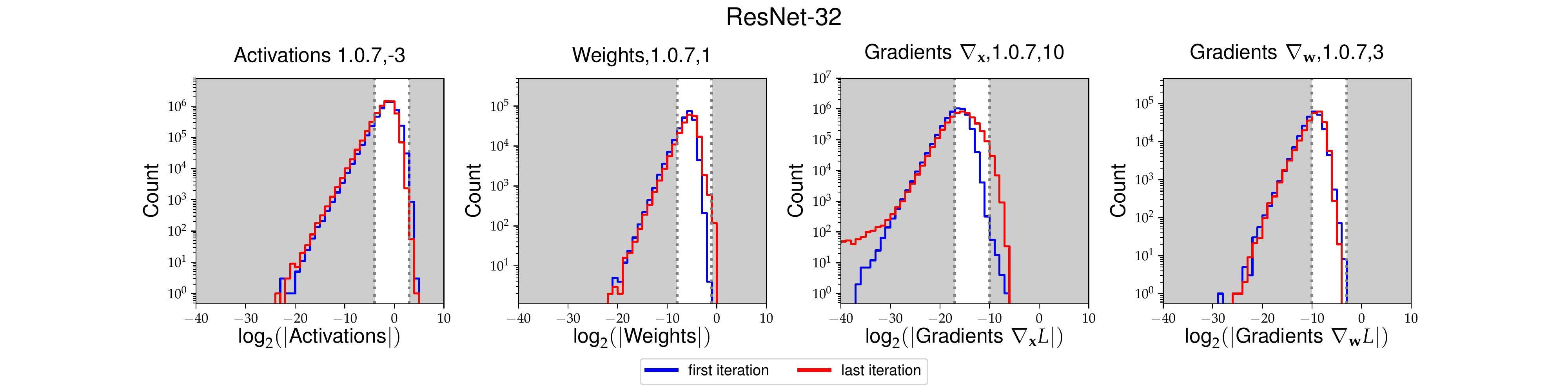}
\caption{Histograms of the weights, activations and gradients of all layers for ResNet-32 CIFAR-100 training. The white areas correspond to the range of values representable by the 1.0.7 (scaled integer) number format.}
\label{fig:ResNet-32_CIFAR-100_Format107_AllLayer_Histogram}
\end{figure}

Figure~\ref{fig:ResNet-32_CIFAR-100_Format107_AllLayer_Histogram} gives the histograms for ResNet-32 CIFAR-100 training, together with the dynamic range covered by the scaled integer format 1.0.7.

\begin{figure}[!htb]
\hspace{-0.9cm}
\includegraphics[width=1.1\textwidth]{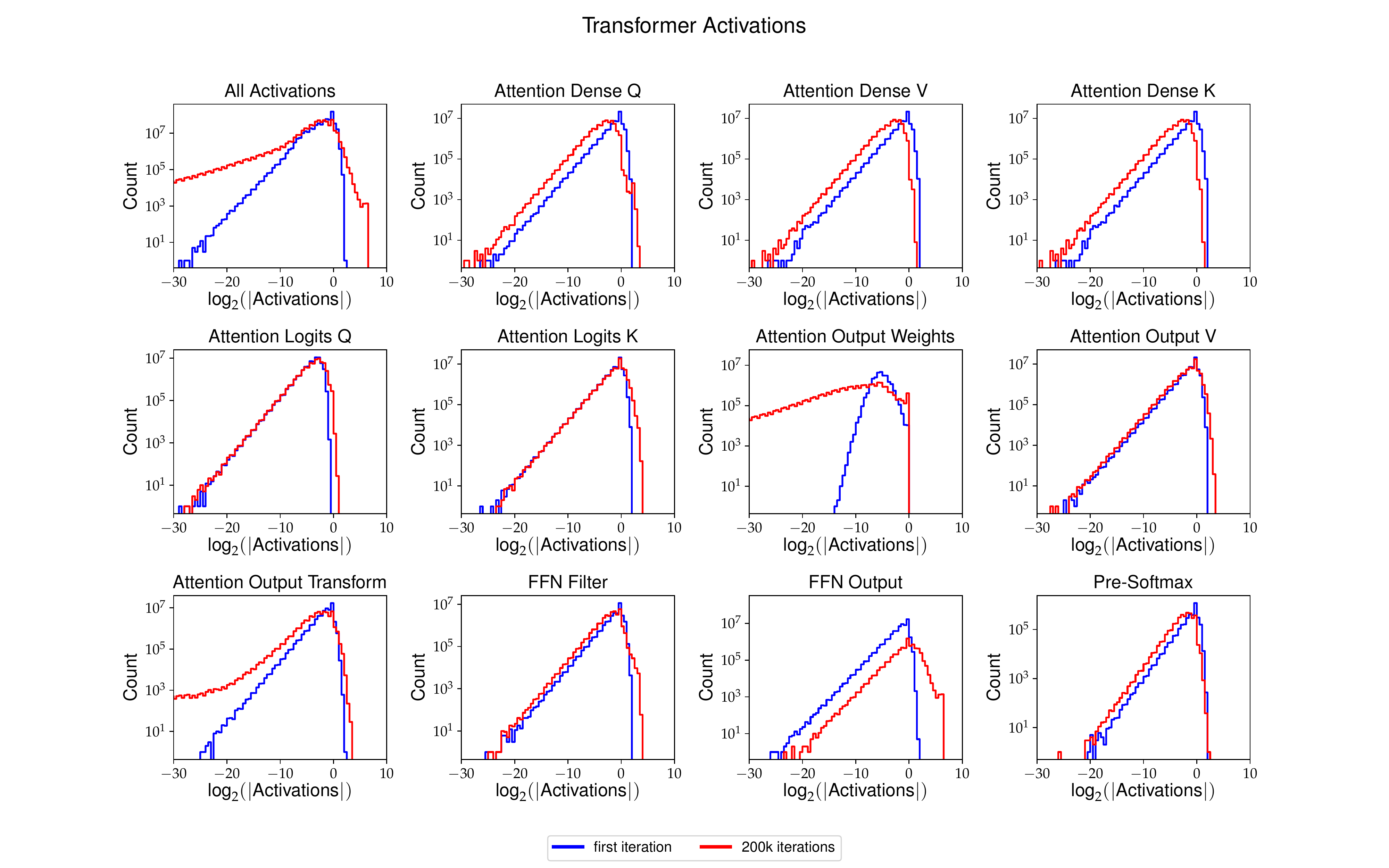}
\caption{Histograms of the activation of different layers of the Transformer for WMT14 English-German training.}
\label{fig:Transformer_HistogramActivations}
\end{figure}

\begin{figure}[!htb]
\hspace{-0.9cm}
\includegraphics[width=1.1\textwidth]{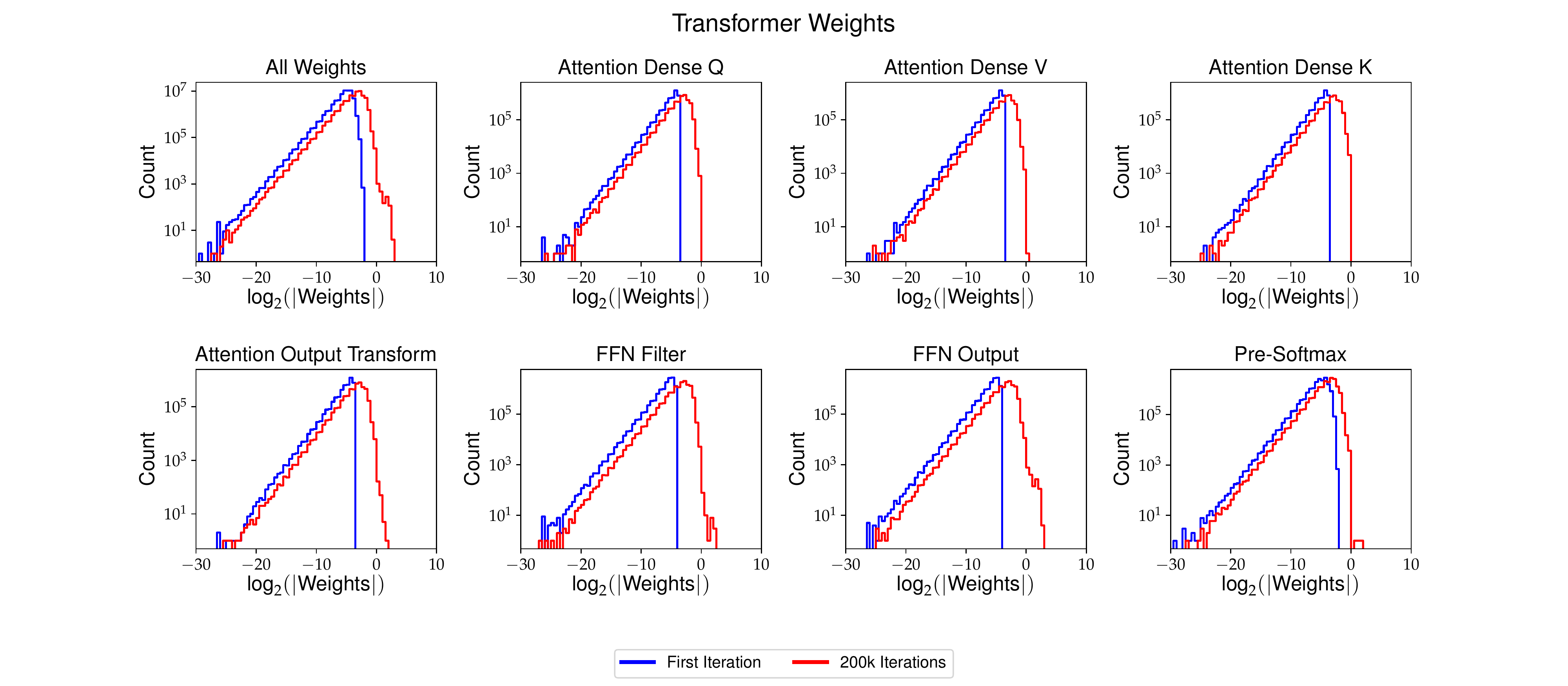}
\caption{Histograms of the weights of different layers of the Transformer for WMT14 English-German training.}
\label{fig:Transformer_HistogramWeights}
\end{figure}

\begin{figure}[!htb]
\hspace{-0.9cm}
\includegraphics[width=1.1\textwidth]{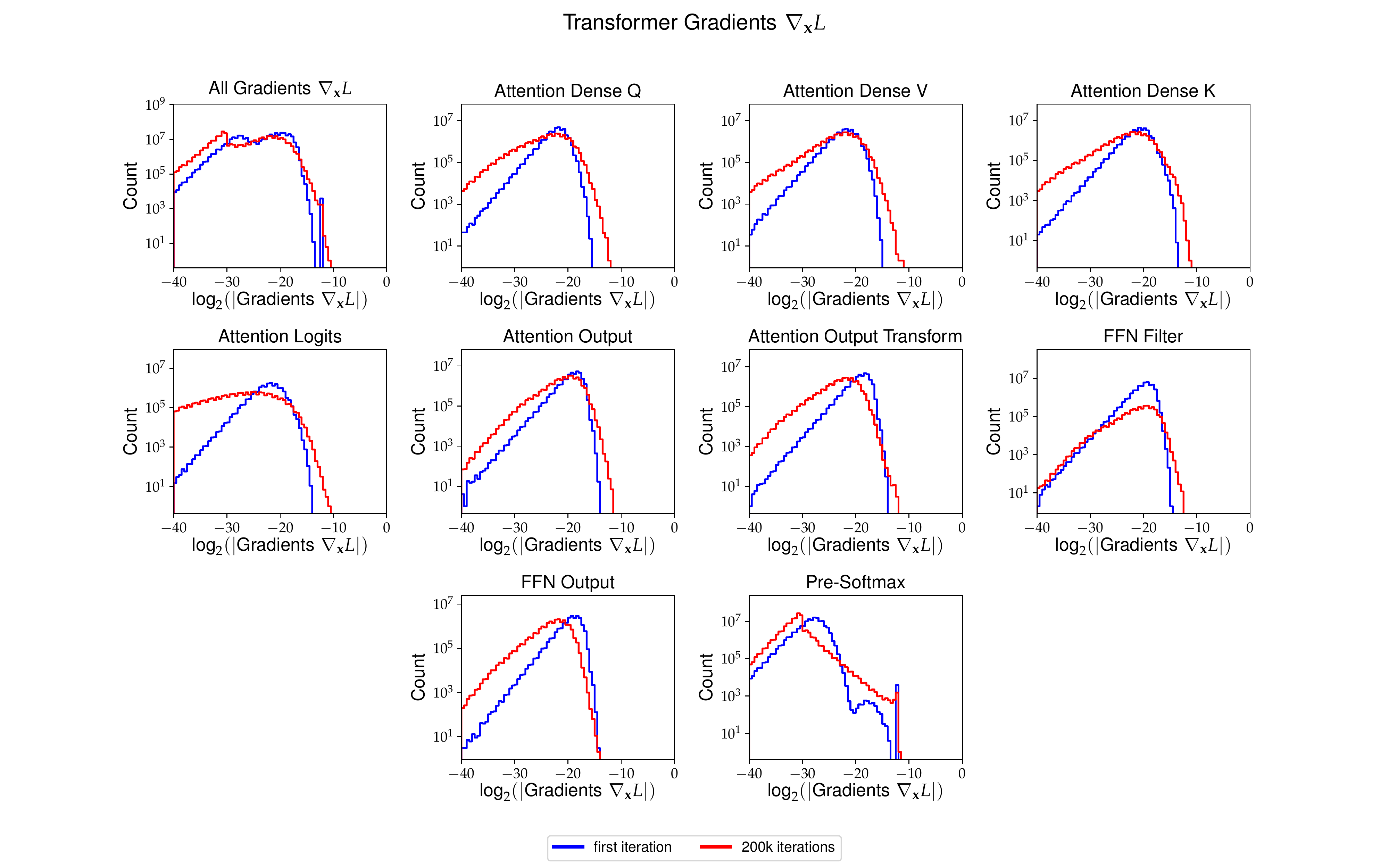}
\caption{Histograms of the gradients with respect to activations of different layers of the Transformer for WMT14 English-German training.}
\label{fig:Transformer_HistogramGradActivations}
\end{figure}

\begin{figure}[!htb]
\hspace{-0.9cm}
\includegraphics[width=1.1\textwidth]{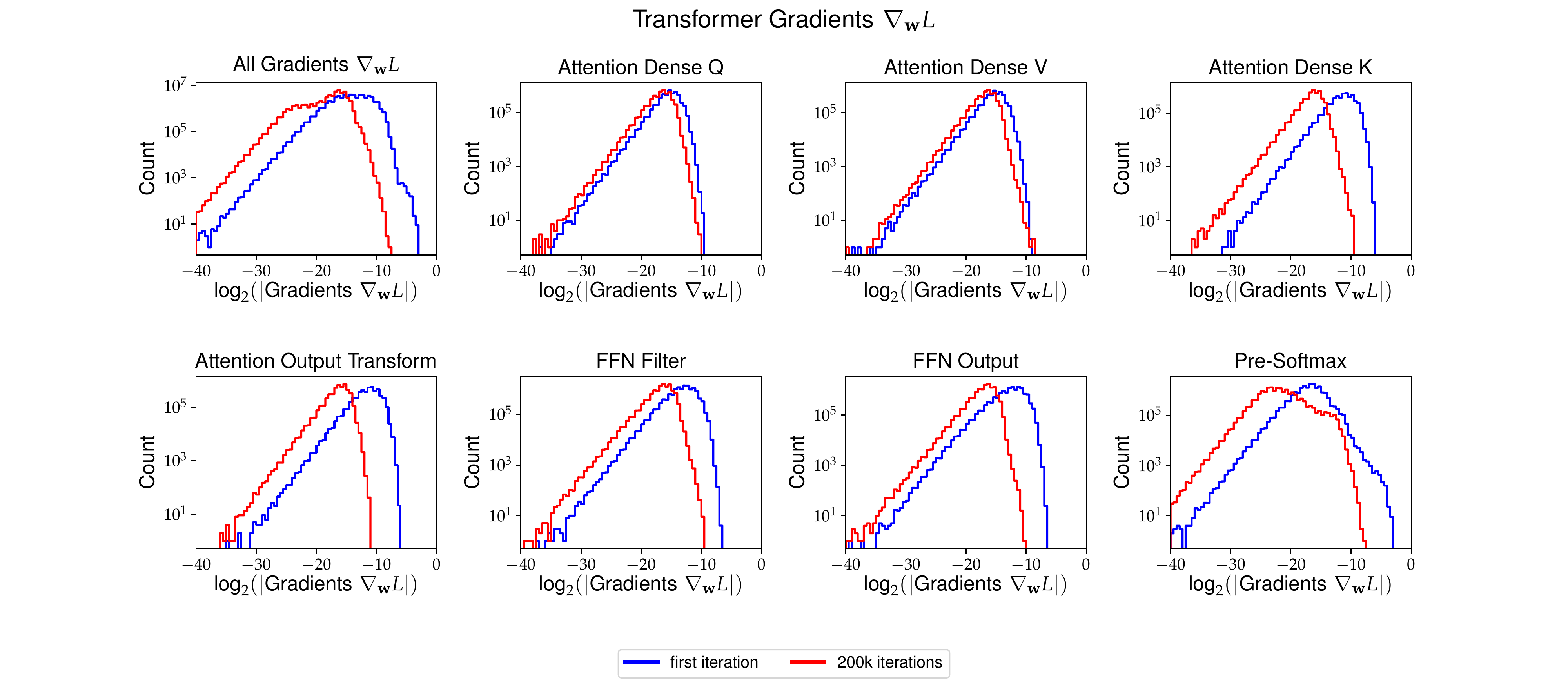}
\caption{Histograms of the gradient with respect to weights of different layers of the Transformer for WMT14 English-German training.}
\label{fig:Transformer_HistogramGradWeights}
\end{figure}

\begin{figure}[!htb]
\hspace{-0.9cm}
\includegraphics[width=1.1\textwidth]{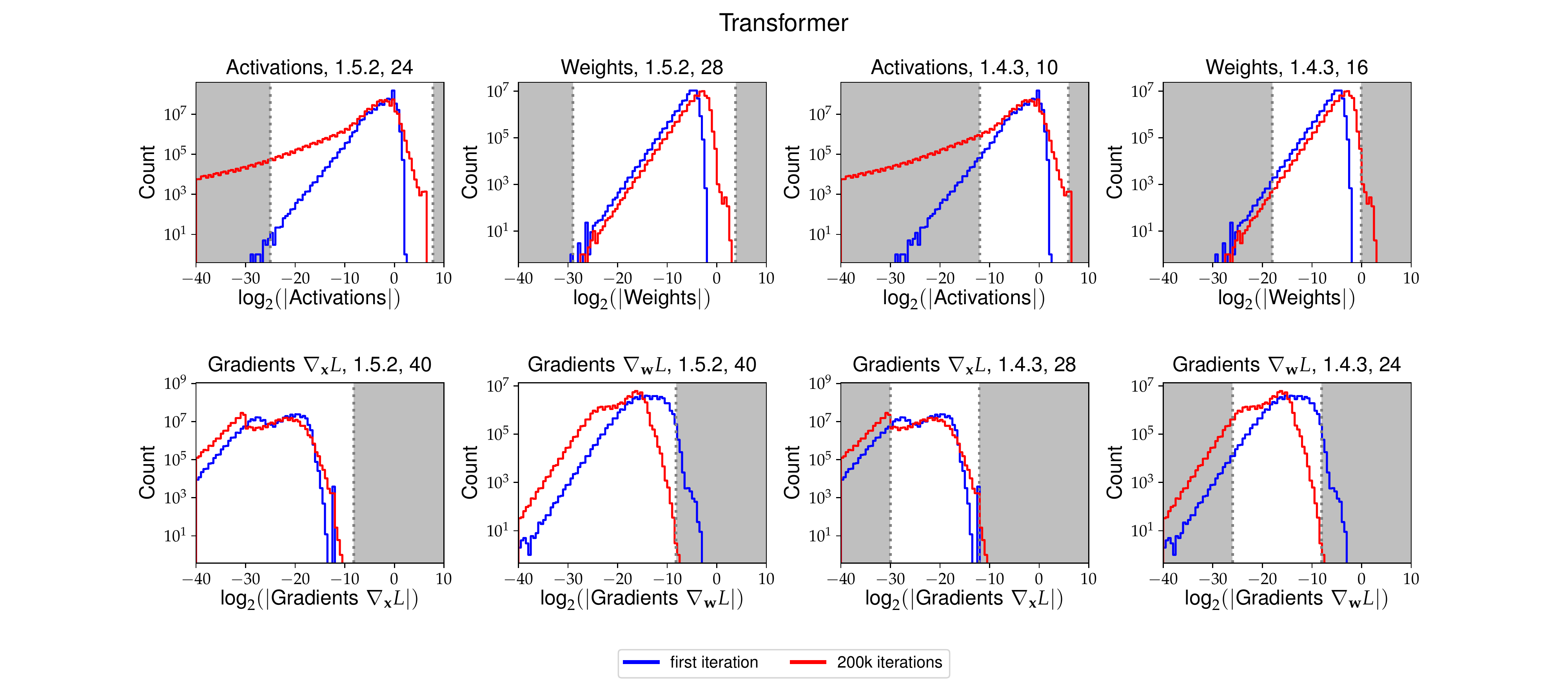}
\caption{Histogram of the weights, activations and gradients of the Transformer model for WMT14 English-German training. The white areas correspond to the range of values representable by the respective 8-bit floating-point number format.}
\label{fig:Transformer_QuantizedHistogram}
\end{figure}

\begin{figure}[!htb]
\hspace{-0.9cm}
\includegraphics[width=1.1\textwidth]{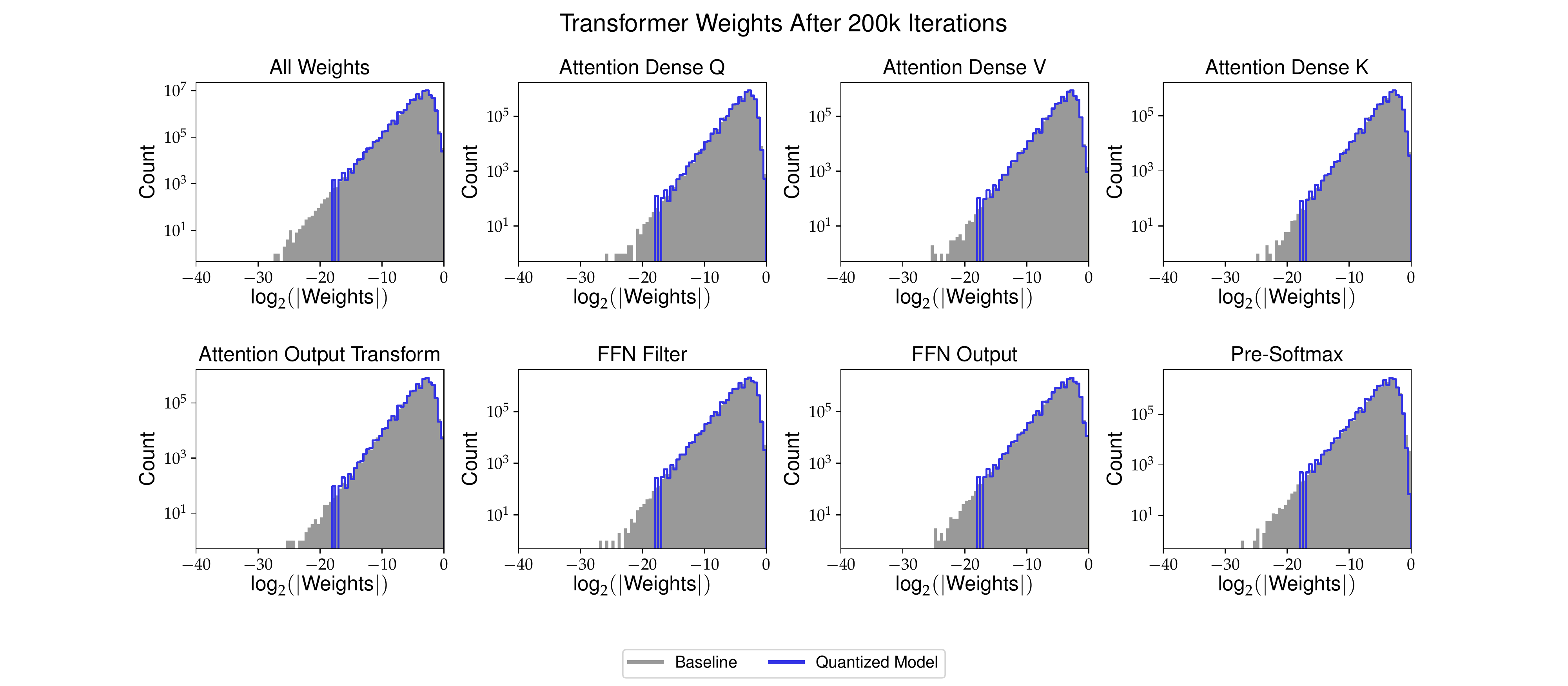}
\caption{Histogram of the weights of the Transformer model for WMT14 English-German training after 200,000 iterations. The quantized model uses 1.4.3, bias 16 for weights, 1.4.3, bias 10 for activations and 1.5.2, bias 40 for gradients with respect to weights and activations.}
\label{fig:Transformer_HistogramWeightsFullyQuantized}
\end{figure}

Figures~\ref{fig:Transformer_HistogramActivations}--\ref{fig:Transformer_HistogramGradWeights} show the histograms of activations, weights, gradients with respect to activations, and gradients with respect to weights of the different layer types of the Transformer, trained on the WMT14 English-German dataset. Figure~\ref{fig:Transformer_QuantizedHistogram} presents the composite histograms of the Transformer model together with the dynamic range of different 8-bit floating-point number formats that have been used throughout this paper. Finally, Figure~\ref{fig:Transformer_HistogramWeightsFullyQuantized} provides evidence that a model learned through quantized training has a similar distribution of weights as the corresponding float-32 baseline model.
\vspace{10mm}

\FloatBarrier
\section{Loss scaling factor}\label{AppendixLossScalingFactor}
\begin{figure}[!htb]
\centering
\subfloat[]{\includegraphics[width=0.39\columnwidth]{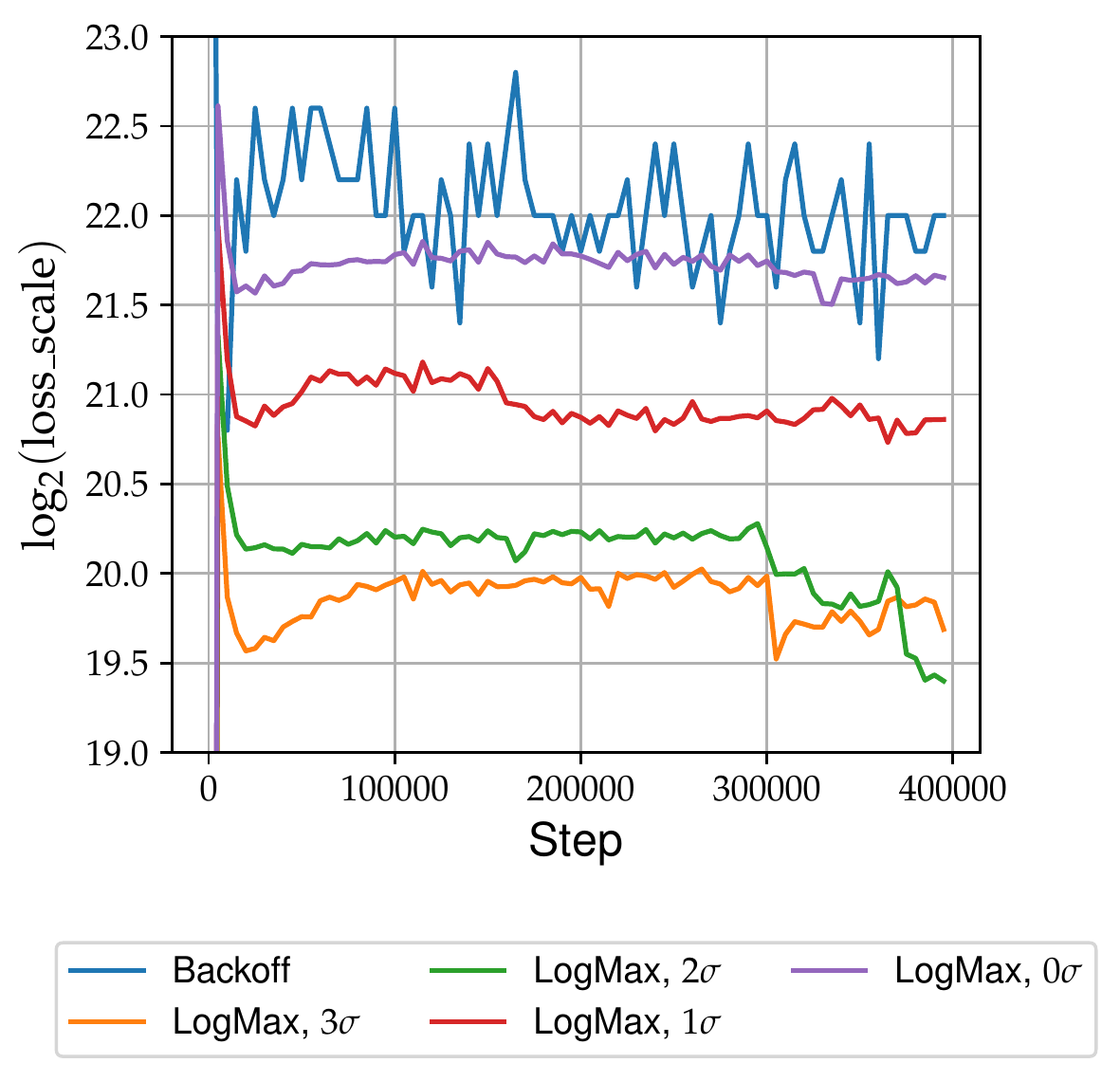}}
\subfloat[]{\includegraphics[width=0.39\columnwidth]{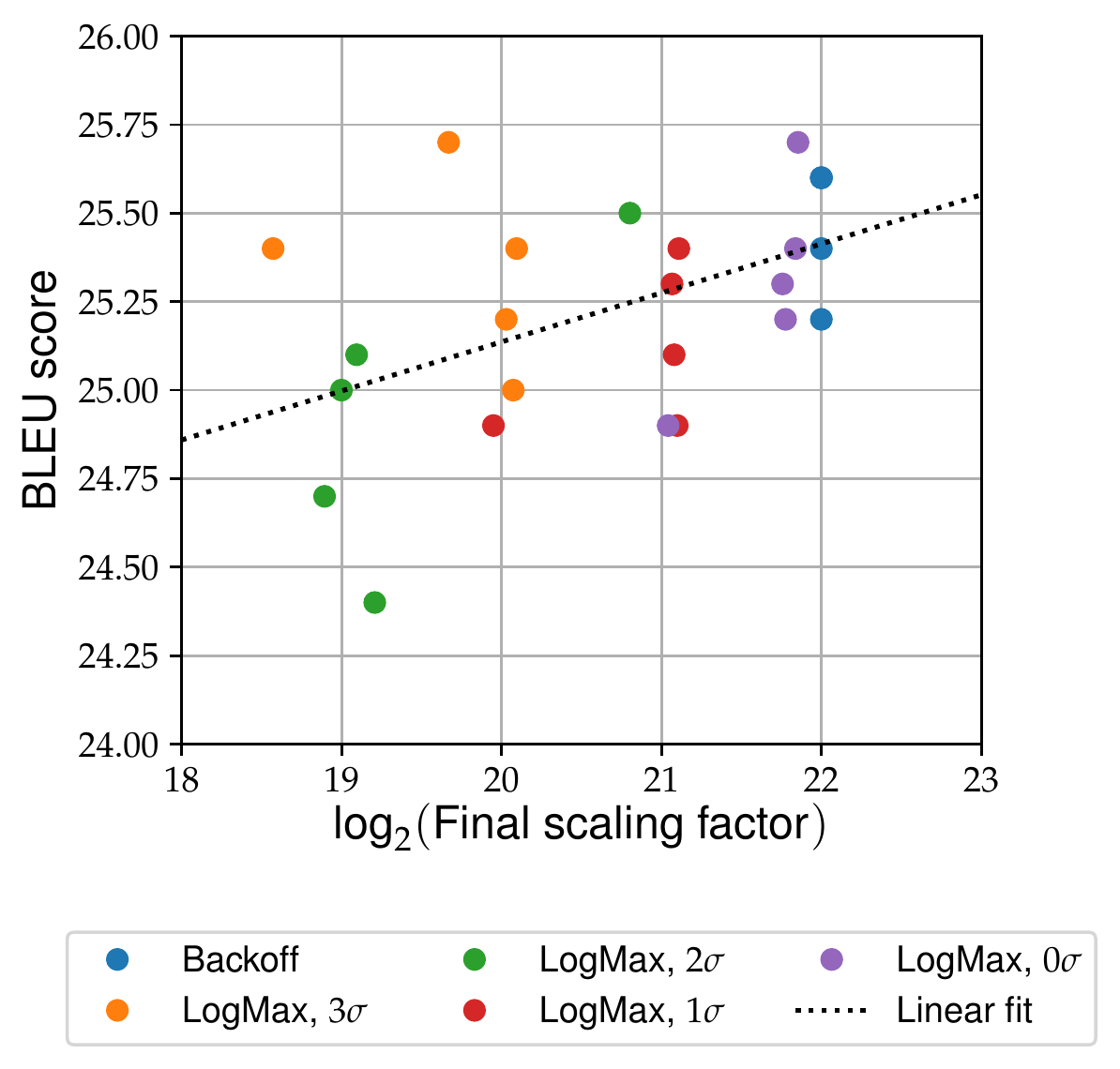}}%
\caption{(a) Loss scaling factor during training of the Transformer with different automatic loss scaling algorithms. The 1.5.2 format is used for activations (bias 24), weights (bias 28) and gradients $\nabla_{\bf x}L$ and $\nabla_{\bf w}L$ (bias 15). Each line corresponds to the average of five independent training runs.
(b) The BLEU score plotted against the final loss scaling factor for the experiments presented in (a).}
\label{fig:LossScalingFactor}
\end{figure}

Figure~\ref{fig:LossScalingFactor}(a) reports the evolution of the values of the loss scaling factor that was used by the different automatic loss scaling algorithms, throughout the training of the Transformer model on the WMT14 English-German dataset. 
Figure~\ref{fig:LossScalingFactor}(b) shows the dependence between the BLEU score and the loss scaling factor at the end of the training for the experiments presented in Figure~\ref{fig:LossScalingFactor}(a).
\vspace{5mm}

\end{document}